\documentclass[sigconf, nonacm]{acmart}
\usepackage{amsmath}
\usepackage{bbold}
\usepackage{mathtools}
\usepackage{amsthm}
\usepackage[capitalize,noabbrev]{cleveref}
\usepackage{url}
\usepackage{amsfonts}
\usepackage{nicefrac}
\usepackage{xcolor}
\usepackage{wrapfig}
\usepackage{relsize}
\usepackage{multirow}
\begin{document}
\title{Reckoning with the Disagreement Problem:\\ Explanation Consensus as a Training Objective}

\author{Avi Schwarzschild}
\authornote{Work completed while working at Arthur.}
\affiliation{
\institution{University of Maryland}
\city{College Park}
\state{Maryland}
\country{USA}
}

\author{Max Cembalest}
\affiliation{
\institution{Arthur}
\city{New York City}
\state{New York}
\country{USA}
}
\author{Karthik Rao}
\affiliation{
\institution{Arthur}
\city{New York City}
\state{New York}
\country{USA}
}
\author{Keegan Hines}
\affiliation{
\institution{Arthur}
\city{New York City}
\state{New York}
\country{USA}
}
\author{John Dickerson}
\authornote{Correspondence to: John Dickerson at <\texttt{john@arthur.ai}>, Avi Schwarzschild at <\texttt{avi1@umd.edu}>.}
\affiliation{
\institution{Arthur}
\city{New York City}
\state{New York}
\country{USA}
}

\begin{abstract}
As neural networks increasingly make critical decisions in high-stakes settings, monitoring and explaining their behavior in an understandable and trustworthy manner is a necessity. 
One commonly used type of explainer is post hoc feature attribution, a family of methods for giving each feature in an input a score corresponding to its influence on a model's output. 
A major limitation of this family of explainers in practice is that they can disagree on which features are more important than others.
Our contribution in this paper is a method of training models with this \emph{disagreement problem} in mind.
We do this by introducing a Post hoc Explainer Agreement Regularization (PEAR) loss term alongside the standard term corresponding to accuracy, an additional term that measures the difference in feature attribution between a pair of explainers. 
We observe on three datasets that we can train a model with this loss term to improve explanation consensus on unseen data, and see improved consensus between explainers other than those used in the loss term.
We examine the trade-off between improved consensus and model performance. 
And finally, we study the influence our method has on feature attribution explanations.
\end{abstract}


\maketitle

\section{Introduction}
\label{sec:intro}

As machine learning becomes inseparable from important societal sectors like healthcare and finance, increased transparency of how complex models arrive at their decisions is becoming critical. In this work, we examine a common task in support of model transparency that arises with the deployment of complex black-box models in production settings: explaining which features in the input are most influential in the model's output. This practice allows data scientists and machine learning practitioners to rank features by importance -- the features with high impact on model output are considered more important, and those with little impact on model output are considered less important. These measurements inform how model users debug and quality check their models, as well as how they explain model behavior to stakeholders.

\begin{figure}[ht!]
    \centering
    \includegraphics[width=0.95\columnwidth]{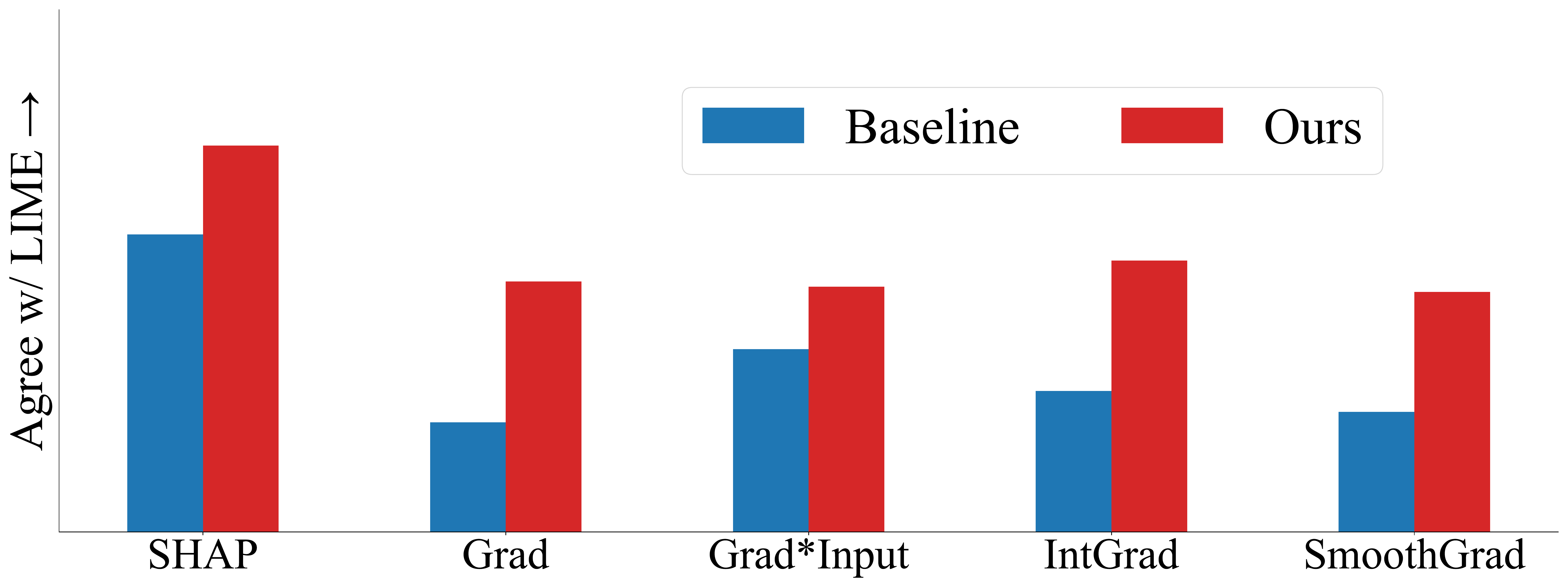}
    \caption{Our loss that encourages explainer consensus boosts the correlation between LIME and other common post hoc explainers. This comes with a cost of less than two percentage points of accuracy compared with our baseline model on the Electricity dataset. Our method improves consensus on six agreement metrics and all pairs of explainers we evaluated. Note that this plot measures the \emph{rank correlation} agreement metric and the specific bar heights depend on this choice of metric.}
    \label{fig:banner}
\end{figure}

\subsection{Post Hoc Explanation}
\label{sec:post-hoc-xai}

The methods of model explanation considered in this paper are post hoc local feature attribution scores. The field of explainable artificial intelligence (XAI) is rapidly producing different methods of this type to make sense of model behavior \citep[e.g.,][]{ribeiro2016should, lundberg2017unified, simonyan2013deep, shrikumar2017learning, sundararajan2017axiomatic}. Each of these methods has a slightly different formula and interpretation of its raw output, but in general they all perform the same task of attributing a model's behavior to its input features. When tasked to explain a model's output with a corresponding input (and possible access to the model weights), these methods answer the question, ``How influential is each individual feature of the input in the model's computation of the output?''

Data scientists are using post hoc explainers at increasing rates -- popular methods like LIME and SHAP have had over 350 thousand and 6 million downloads of their Python packages in the last 30 days, respectively~\citep{pepy}.


\subsection{The Disagreement Problem}
\label{sec:disagreement}

The explosion of different explanation methods leads
\citet{krishna2022disagreement} to observe that when neural networks are trained naturally, i.e. for accuracy alone, often post hoc explainers disagree on how much different features influenced a model's outputs. They coin the term \emph{the disagreement problem} and argue that when explainers disagree about which features of the input are important, practitioners have little concrete evidence as to which of the explanations, if any, to trust. 

There is an important discussion around local explainers and their true value in reaching the communal goal of model transparency and interpretability~\citep[see, e.g.,][]{Doshi17:Towards,Lipton18:Mythos,Rudin19:Stop}; indeed, there are ongoing discussions about the efficacy of present-day explanation methods in specific domains~\citep[for healthcare see, e.g.,][]{Ghassemi21:False}. 
Feature importance estimates may fail at making a model more transparent when the model being explained is too complex to allow for easily attributing the output to the contribution of each individual feature.

In this paper, we make no normative judgments with respect to this debate, but rather view ``explanations'' as signals to be used alongside other debugging, validation, and verification approaches in the machine learning operations (MLOps) pipeline.  Specifically, we take the following practical approach: make the amount of explanation disagreement a controllable model parameter instead of a point of frustration that catches stakeholders off-guard. 

\subsection{Encouraging Explanation Consensus}
\label{sec:encouraging-consensus}

Consensus between two explainers does not require that the explainers output the same exact scores for each feature. 
Rather, consensus between explainers means that whatever disagreement they exhibit can be reconciled. 
Data scientists and machine learning practitioners say in a survey that explanations are in basic agreement if they satisfy agreement metrics that align with human intuition, which provides a quantitative way to evaluate the extent to which consensus is being achieved \cite{krishna2022disagreement}. 
When faced with disagreement between explainers, a choice has to be made about what to do next -- if such an arbitrary crossroads moment is avoidable via specialized model training, we believe it would be a valuable addition to a data scientist's toolkit.

We propose, as our main contribution, a training routine to help alleviate the challenge posed by post hoc explanation disagreement.
Achieving better consensus between explanations does not provide more interpretability to a model inherently. But, it may lend more trust to the explanations if different approaches to attribution agree more often on which features are important. This gives consensus the practical benefit of acting as a sanity check -- if consensus is observed, the choice of which explainer a practitioner uses is less consequential with respect to downstream stakeholder impact, making their interpretation less subjective.

\section{Related Work}
\label{sec:related-work}

Our work focuses on post hoc explanation tools. Some post hoc explainers, like LIME \citep{ribeiro2016should} and SHAP \cite{lundberg2017unified}, are proxy models trained atop a base machine learning model with the sole intention of ``explaining'' that base model. 
These explainers rely only on the model's inputs and outputs to identify salient features. 
Other explainers, such as Vanilla Gradients (Grad) \citep{simonyan2013deep}, Gradient Times Input (Grad*Input) \citep{shrikumar2017learning}, Integrated Gradients (IntGrad) \citep{sundararajan2017axiomatic} and SmoothGrad \cite{smilkov2017smoothgrad}, do not use a proxy model but instead compute the gradients of a model with respect to input features to identify important features.\footnote{%
In many settings, there may be a strong case to consider \emph{interpretable-by-design} models---that is, models that need no proxy model or gradient computation to be explained, and are instead interpretable in their base form.  \citet{Rudin19:Stop} provides an overview of this space, and we specifically call out directions such as falling rule lists~\citep{Wang15:Falling}, generalized additive models~\citep{Lou12:Intelligible}, and concept/prototype-based models~\citep{Ghorbani19:Towards,Koh20:Concept}.  We acknowledge this direction of research as well as subsequent push-back claiming that performance drops from prioritizing interpretability may be prohibitively high~\citep[e.g., when compared to so-called foundation models, see][]{Bommasani21:Opportunities}.  Given industry uptake of post hoc explanations, our paper focuses on that approach alone.%
}
Each of these explainers has its quirks and there are reasons to use, or not use, them all---based on input type, model type, downstream task, and so on. But there is an underlying pattern unifying all these explanation tools.
\citet{han2022explanation} provide a framework that characterizes all the post hoc explainers used in this paper as different types of local-function approximation.
For more details about the individual post hoc explainers used in this paper, we refer the reader to the individual papers and to other works about when and why to use each one \citep[see, e.g.,][]{buhrmester2021analysis, jesus2021can}.

We build directly on prior work that defines and explores the disagreement problem \citep{krishna2022disagreement}. Disagreement here refers to the difference in feature importance scores between two feature attribution methods, but can be quantified several different ways as are described by the metrics \citet{krishna2022disagreement} define and use. We describe these metrics in Section \ref{sec:experiments}.

The method we propose in this paper relates to previous work that trains models with constraints on explanations via penalties on the disagreement between feature attribution scores and hand-crafted ground-truth scores \citep{ross2017right, align_prior_knowledge, feature_attribution_priors}. Additionally, work has been done to leverage the disagreement between different post-hoc explanations to construct new feature attribution scores that improve metrics like stability and pairwise rank agreement \citep{partial_order, eval_aggr_explanations, aggregation_stability}. 


\section{PEAR: Post Hoc Explainer Agreement Regularizer}
\label{sec:problem}

Our contribution is the first effort to train models to be both accurate and to be explicitly regularized via consensus between local explainers. 
When neural networks are trained \emph{naturally} (i.e. with a single task-specific loss term like cross-entropy), disagreement between post hoc explainers often arises.
Therefore, we include an additional loss term to measure the amount of explainer disagreement during training to encourage consensus between explanations. Since human-aligned notions of explanation consensus can be captured by more than one agreement metric (listed in \ref{sec:app-metrics}), we aim to improve several agreement metrics with one loss function.\footnote{The PEAR package will be publicly for download on the Package Installer for Python (pip), and it is also available upon request from the authors.}

\begin{figure}[ht!]
    \centering
    \includegraphics[width=\columnwidth]{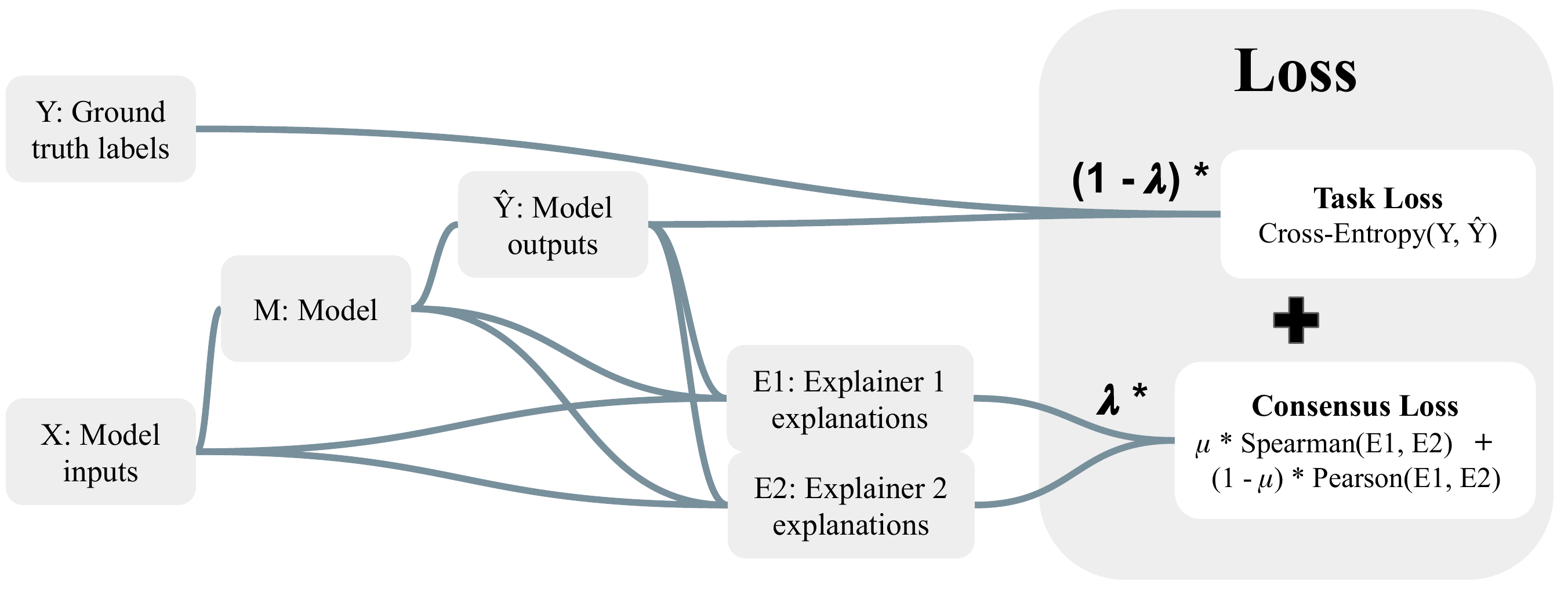}
    \caption{Our loss function measures the task loss between the model outputs and ground truth (task loss), as well as the disagreement between explainers (consensus loss). The weight given to the consensus loss term is controlled by a hyperparameter $\lambda$. The consensus loss term term is a convex combination of the Spearman and Pearson correlation measurements between feature importance scores, since increasing both rank correlation (Spearman) and raw-score correlation (Pearson) are useful for improving explainer consensus on our many agreement metrics.}
    \label{fig:loss}
\end{figure}

Our consensus loss term is a convex combination of the Pearson and Spearman correlation measurements between the vectors of attribution scores (Spearman correlation is just the Pearson correlation on the ranks of a vector).

\begin{figure}[ht!]
    \centering
    \includegraphics[width=\columnwidth,trim=0 4em 0 4em,clip]{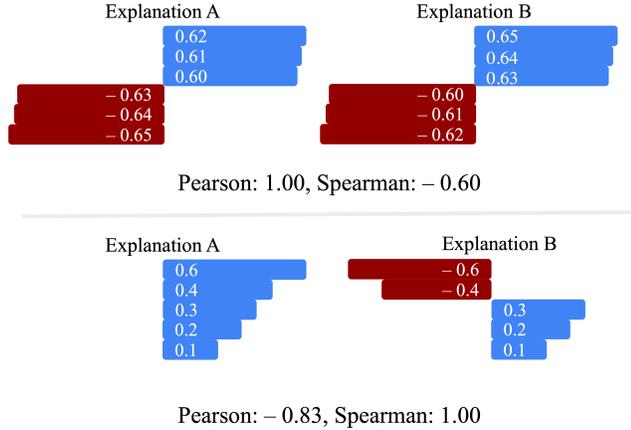}
    \caption{Example feature attribution vectors where Pearson and Spearman show starkly different scores. Recall, both Pearson and Spearman correlation range from $-1$ to $+1$.  Both of these pairs of vectors satisfy some human-aligned notions of consensus. But in each circumstance, one of the correlation metrics gives a low similarity score. Thus, in order to successfully encourage explainer consensus (by all of our metrics), we use both types of correlation in our consensus loss term.}
    \label{fig:disagreement-example}
\end{figure}

To paint a clearer picture of the need for two terms in the loss, consider the examples shown in Figure \ref{fig:disagreement-example}. In the upper example, the raw feature scores are very similar and the Pearson correlation coefficient is in fact 1 (to machine precision). However, when we rank these scores by magnitude, there is a big difference in their ranks as indicated by the Spearman value. Likewise, in the lower portion of Figure \ref{fig:disagreement-example} we show that two explanations with identical magnitudes will show a low Pearson correlation coefficient. Since some of the metrics we use to measure disagreement involve ranking and others do not, we conclude that a mixture of these two terms in the loss is appropriate.

While the example in Figure \ref{fig:disagreement-example} shows two explanation vectors with similar scale, different explanation methods do not always align. Some explainers have the sums of their attribution scores constrained by various rules, whereas other explainers have no such constraints. The correlation measurements we use in our loss provide more latitude when comparing explainers than a direct difference measurement like mean absolute error or mean squared error, allowing our correlation measurement.

More formally, our full loss function is defined as follows. Let $f$ denote a model. Let $E_1$ and $E_2$ be any two post-hoc explainers, each of which take a data point $x$ and its predicted label $\hat y$ as input and output a vector, which is the same size as $x$ and has corresponding feature attribution scores. We define $R$ to be the ranking function, so it replaces each entry in a vector with the rank of its magnitude among all entries in the vector.\footnote{When more than one of the entries have the same magnitude, they get a common ranking value equal to the average rank if they were ordered arbitrarily.}

Let the functions $p(a, b)$ and $s(a, b)$ be Pearson and Spearman correlation measurements, respectively. We denote the average value of all entries in a vector with the $\bar \cdot$ notation.
\begin{equation}
    p(a, b) = \sum_i \frac{(a_i - \bar a)(b_i - \bar b)}{\Vert a \Vert \Vert b \Vert}
    \label{eq:pearson}
\end{equation}
\begin{equation}
    s(a, b) = \sum_i \frac{(R(a)_i - \overline{R(a)})(R(b)_i - \overline{R(b)})}{\Vert R(a) \Vert \Vert R(b) \Vert}
    \label{eq:spearman}
\end{equation}

We refer to the first term in the loss function as the \emph{task loss}, or $\ell_\text{task}$, and for our classification tasks we use cross-entropy loss. A graphical depiction of the flow from data to loss value is shown in Figure \ref{fig:loss}. Formally, our complete loss function can be expressed as follows with two hyperparameters $\lambda, \mu ~\in~[0, 1]$. We weight the influence of our consensus term with  $\lambda$, so lower values give more priority to task loss. We weight the influence between the two explanation correlation terms with $\mu$, so lower values give more weight to Pearson correlation and higher values give more weight to Spearman correlation.

\begin{equation}
    \begin{aligned}
        L(x, y, f, E_1,& E_2) = \\
        (1 - \lambda) & \ell_\text{task} \\
        + \lambda \bigg( &\mu ~s\big(E_1(x, y), E_2(x, y)\big) \\
         &+ (1 - \mu) ~p\big(E_1(x, y), E_2(x, y)\big)\bigg)
    \label{eq:loss}
    \end{aligned}
\end{equation}

\subsection{Choosing a Pair of Explainers}

The consensus loss term is defined for any two explainers in general, but since we train with standard backpropagation we need these explainers to be differentiable. 
With this constraint in mind, and with some intuition about the objective of improving agreement metrics, we choose to train for consensus between Grad and IntGrad. 
If Grad and IntGrad align, then the function should become more locally linear in logit space. IntGrad computes the average gradient along a path in input space toward each point being explained. 
So, if we train the model to have a local gradient at each point (Grad) closer to the average gradient along a path to the point (IntGrad), then perhaps an easy way for the model to accomplish that training objective would be for the gradient along the whole path to equal the local gradient from Grad. 
This may push the model to be more similar to a linear model.
This is something we investigate with qualitative and quantitative analysis in Section~\ref{sec:linearity}.

\subsection{Differentiability}

On the note of differentiability, the ranking function $R$ is not differentiable. We substitute a soft ranking function from the \texttt{torchsort} package \citep{blondel2020fast}. This provides a floating point approximation of the ordering of a vector rather than an exact integer computation of the ordering of a vector, which allows for differentiation. 

\section{The Efficacy of Consensus Training}
\label{sec:experiments}

\begin{figure*}[ht!]
    \centering
    \includegraphics[width=0.49\textwidth,trim={0.3cm 0.8cm 0.8cm 0.5cm},clip]{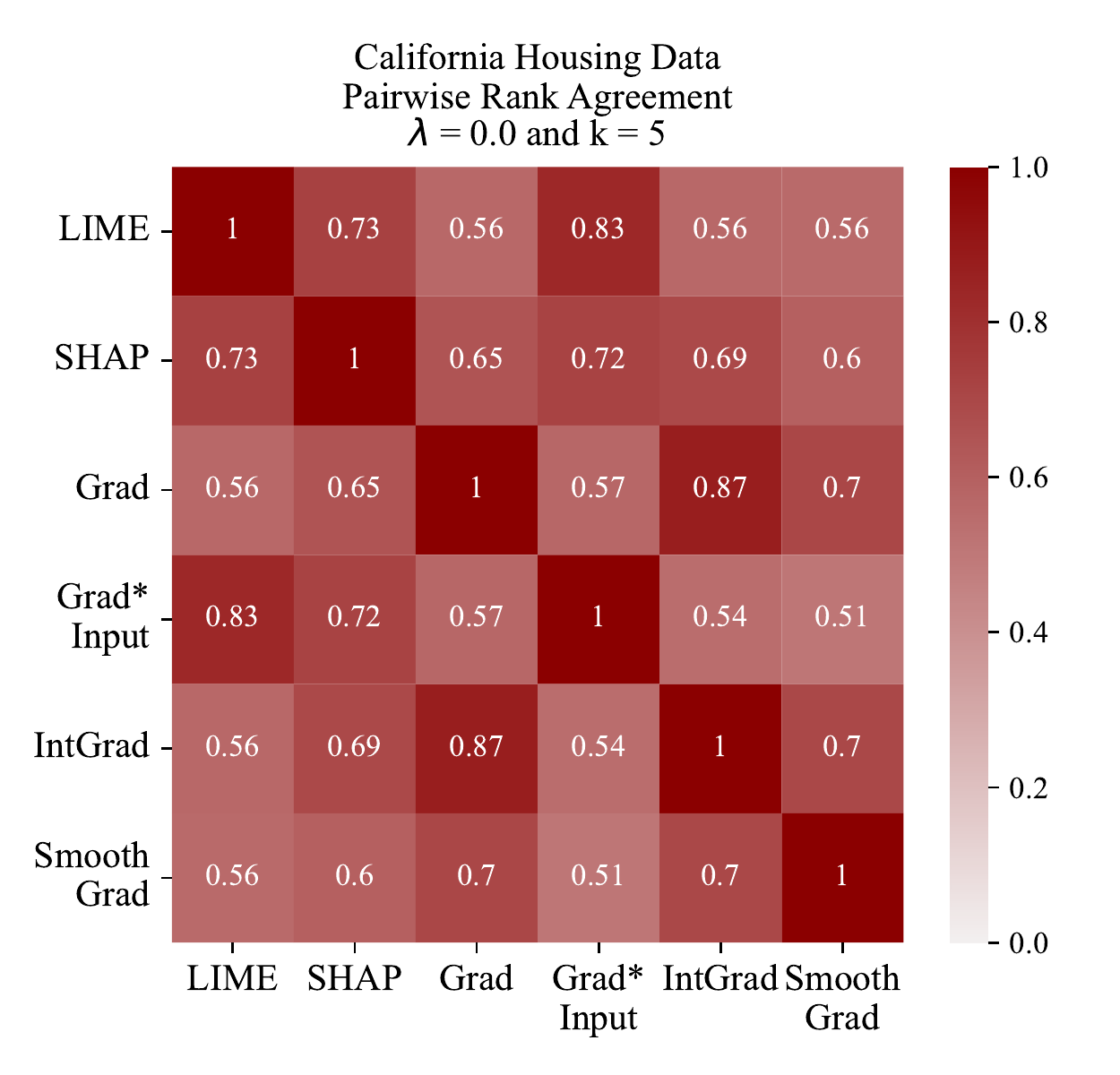}
    \includegraphics[width=0.49\textwidth,trim={0.3cm 0.8cm 0.8cm 0.5cm},clip]{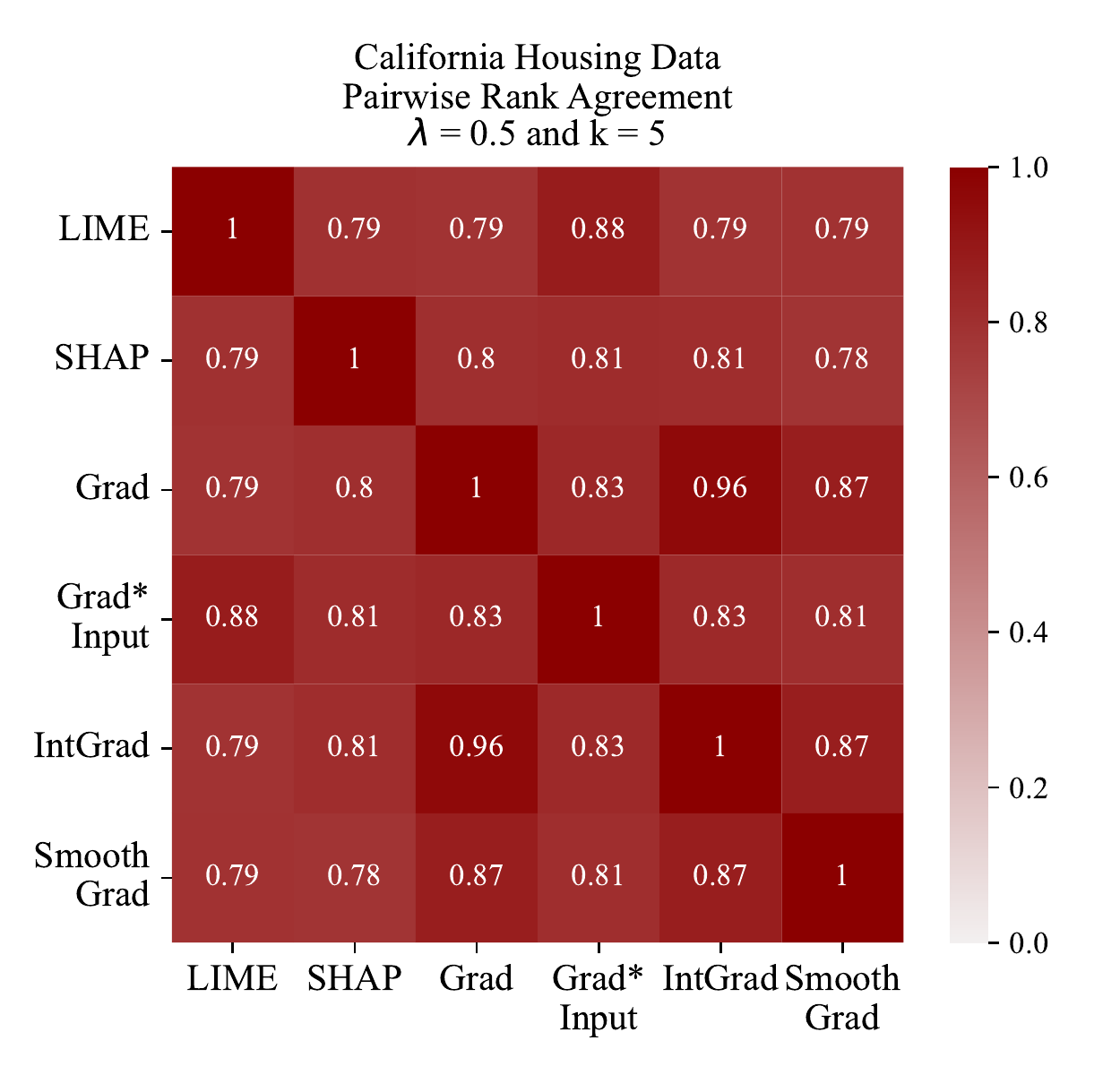}
    \caption{When models are trained naturally, we see disagreement among post hoc explainers (left). However, when trained with our loss function, we see a boost in agreement with only a small cost in accuracy (right). This can be observed visually by the increase in saturation or in more detail by comparing the numbers in corresponding squares.}
    \label{fig:redgrids}
\end{figure*}

In this section we present each experiment with the hypothesis it is designed to test.
The datasets we use for our experiments are Bank Marketing, California Housing, and Electricity, three binary classification datasets available on the OpenML database \citep{OpenML2013}. For each dataset, we use a linear model's performance (logistic regression) as a lower bound of realistic performance because linear models are considered inherently explainable.

The models we train to study the impact of our consensus loss term are multilayer perceptrons (MLPs). While the field of tabular deep learning is still growing, and MLPs may be an unlikely choice for most data scientists on tabular data, deep networks provide the flexibility to adapt training loops for multiple objectives \citep{arik2021tabnet, somepalli2021saint, gorishniy2021revisiting, rubachev2022revisiting, levin2022transfer, shwartz2022tabular}. We also verify that our MLPs outperform linear models on each dataset, because if deep models trained to reach consensus are less accurate than a linear model, we would be better off using the linear model.

We include XGBoost \citep{chen2016xgboost} as a point of comparison for our approach, as it has become a widely popular method with high performance and strong consensus metrics on many tabular datasets (figures in Appendix \ref{sec:app-three-point}). 
There are cases where we achieve more explainer consensus than XGBoost, but this point is tangential as we are invested in exploring a loss for training neural networks. 

For further details on our datasets and model training hyperparameters, see Appendices \ref{sec:app-datasets} and \ref{sec:app-hyperparameters}. 

\subsection{Agreement Metrics}
\label{sec:metrics}

In their work on the disagreement problem, \citet{krishna2022disagreement} introduce six metrics to measure the amount of agreement between post hoc feature attributions. 
Let $[E_1(x)]_i$, $[E_2(x)]_i$ be the attribution scores from explainers for the $i$-th feature of an input $x$.
A feature's \textit{rank} is its index when features are ordered by the absolute value of their attribution scores.
A feature is considered in the \textit{top-$k$ most important} features if its rank is in the top-$k$.
For example, if the importance scores for a point $ x = [x_1, x_2, x_3, x_4]$, output by one explainer are $E_1(x) = [0.1, -0.9, 0.3, -0.2]$, then the most important feature is $x_2$ and its rank is 1 (for this explainer).

\textbf{Feature Agreement} counts the number of features $x_i$ such that $[E_1(x)]_i$ and $[E_2(x)]_i$ are both in the top-$k$. \textbf{Rank Agreement} counts the number of features in the top-$k$ with the same rank in $E_1(x)$ and $E_2(x)$. \textbf{Sign Agreement} counts the number of features in the top-$k$ such that $[E_1(x)]_i$ and $[E_2(x)]_i$ have the same sign. \textbf{Signed Rank Agreement} counts the number of features in the top-$k$ such that $[E_1(x)]_i$ and $[E_2(x)]_i$ agree on both sign and rank. \textbf{Rank Correlation} is the correlation between $E_1(x)$ and $E_2(x)$ (on all features, not just in the top-$k$), and is often referred to as the Spearman correlation coefficient. Lastly, \textbf{Pairwise Rank Agreement} counts the number of pairs of features $(x_i, x_j)$ such that $E_1$ and $E_2$ agree on whether $x_i$ or $x_j$ is more important. All of these metrics are formalized as fractions and thus range from $0$ to $1$, except Rank Correlation, which is a correlation measurement and ranges from $-1$ to $+1$. Their formal definitions are provided in Appendix \ref{sec:app-metrics}.

In the results that follow, we use all of the metrics defined above and reference which one is used where appropriate. When we evaluate a metric to measure the agreement between each pair of explainers, we average the metric over the test data to measure agreement. Both agreement and accuracy measurements are averaged over several trials (see Appendices \ref{sec:app-std-error} and \ref{sec:app-redgrids} for error bars).

\begin{figure*}[ht!]
    \centering
    \includegraphics[width=0.33\textwidth]{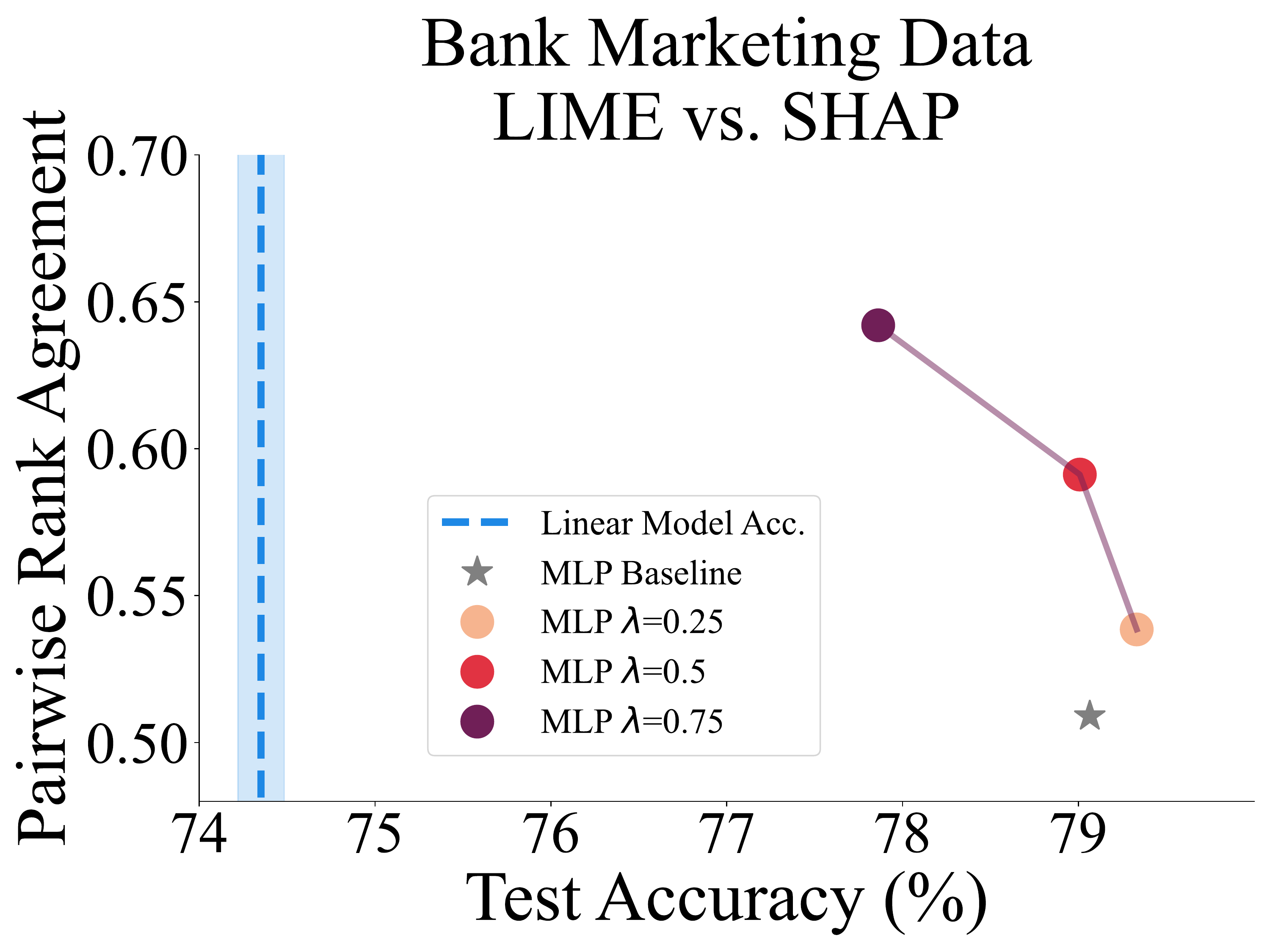}
    \includegraphics[width=0.33\textwidth]{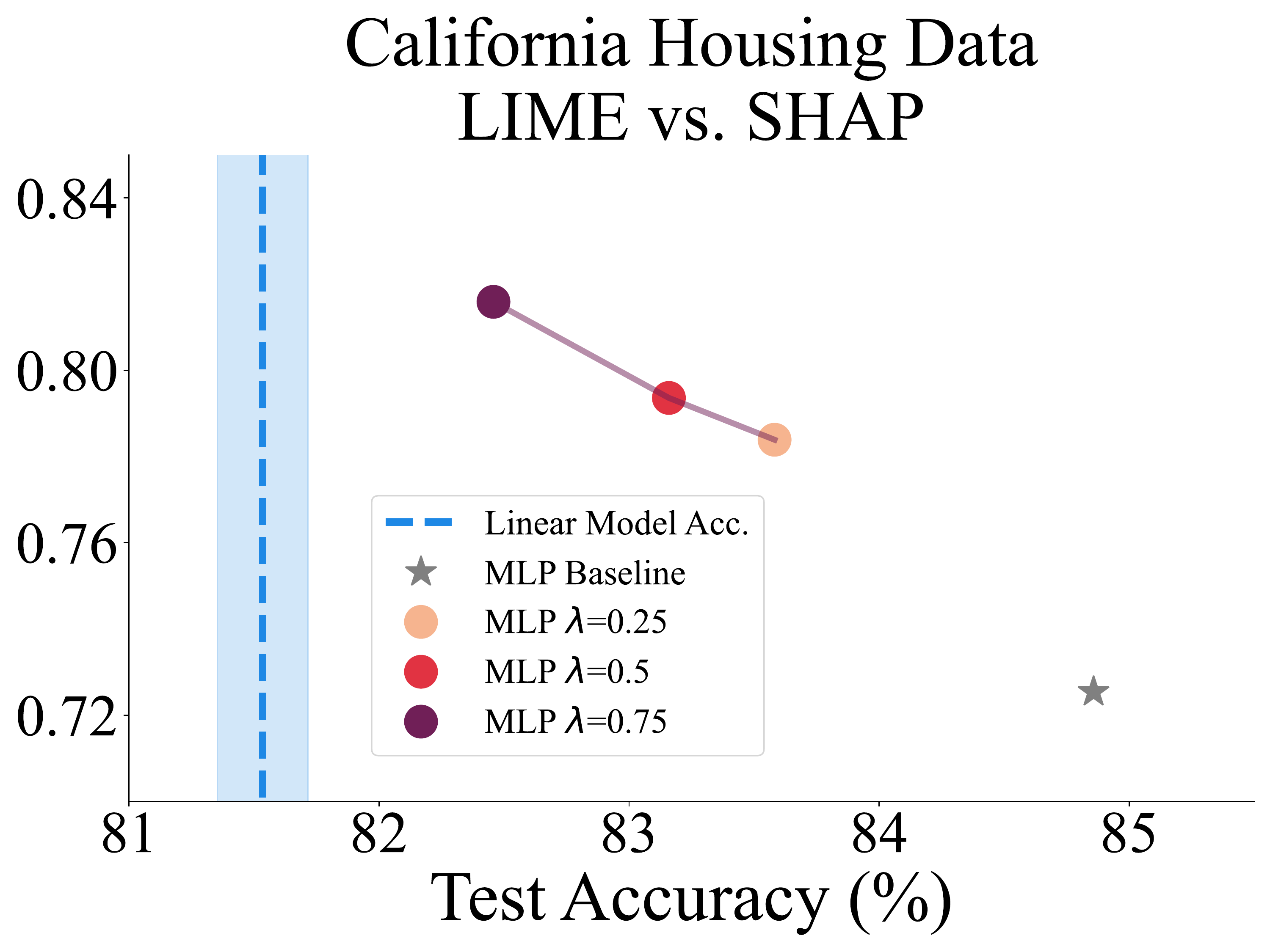}
        \includegraphics[width=0.33\textwidth]{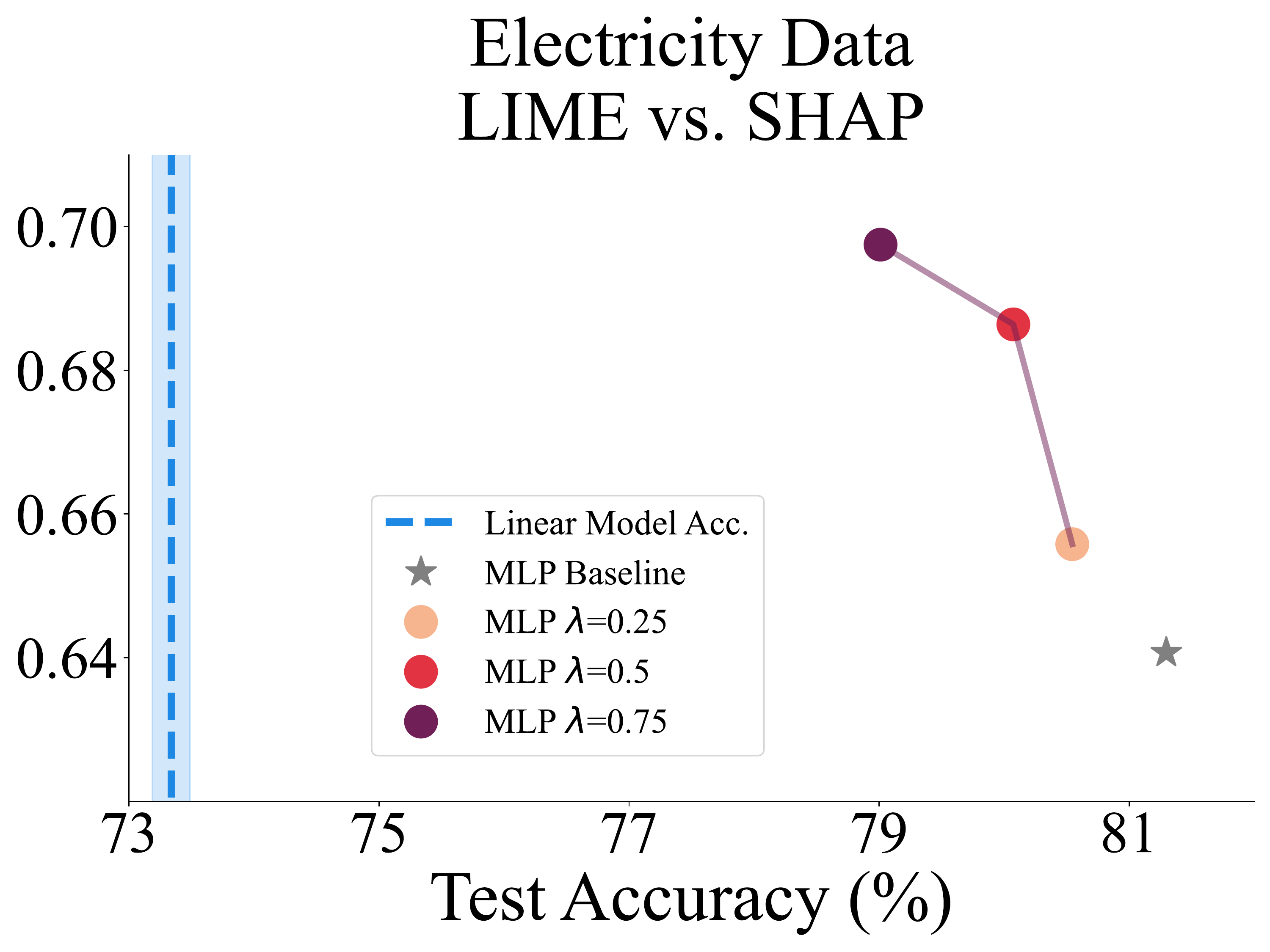}
    \caption{The trade-off curves of consensus and accuracy. Increasing the consensus comes with a drop in accuracy and the trade-off is such that we can achieve more agreement and still outperform linear baselines. Moreover, as we vary the $\lambda$ value, we move along the trade-off curve. In all three plots we measure agreement with the pairwise rank agreement metric and we show that increased consensus comes with a drop in accuracy, but all of our models are still more accurate than the linear baseline, indicated by the vertical dashed line (the shaded region shows $\pm$ one standard error).}
    \label{fig:pareto}
\end{figure*}

\subsection{Improving Consensus Metrics}
\label{sec:improving-metrics}

The intention of our consensus loss term is to improve agreement metrics. While the objective function explicitly includes only two explainers, we show generalization to unseen explainers as well as to the unseen test data. For example, we train for agreement between Grad and IntGrad and observe an increase in consensus between LIME and SHAP.

To evaluate the improvement in agreement metrics when using our consensus loss term, we compute explanations from each explainer on models trained naturally and on models trained with our consensus loss parameter using $\lambda=0.5$. 

In Figure~\ref{fig:redgrids}, using a visualization tool developed by \citet{krishna2022disagreement}, we show how we evaluate the change in an agreement metric (pairwise rank agreement) between all pairs of explainers on the California Housing data.

\textbf{Hypothesis:} \emph{We can increase consensus by deliberately training for post hoc explainer agreement.}

Through our experiments, we observe improved agreement metrics on unseen data and on unseen pairs of explainers. In Figure \ref{fig:redgrids} we show a representative example where Pairwise Rank Agreement between Grad and IntGrad improve from 87\% to 96\% on unseen data. Moreover, we can look at two other explainers and see that agreement between SmoothGrad and LIME improves from 56\% to 79\%. This shows both generalization to unseen data and to explainers other than those explicitly used in the loss term.
In Appendix~\ref{sec:app-redgrids}, we see more saturated disagreement matrices across all of our datasets and all six agreement metrics.

\subsection{Consistency At What Cost?}
\label{sec:cost}

While training for consensus works to boost agreement, a question remains: How accurate are these models?

To address this question, we first point out that there is a trade-off here, i.e., more consensus comes at the cost of accuracy. With this in mind we posit that there is a Pareto frontier on the accuracy-agreement axes. While we cannot assert that our models are on the Pareto frontier, we plot trade-off curves which represent the trajectory through accuracy-agreement space that is carved out by changing $\lambda$.

\textbf{Hypothesis:} \emph{We can increase consensus with an acceptable drop in accuracy.}

While this hypothesis is phrased as a subjective claim, in reality we define acceptable performance as better than a linear model as explained at the beginning of Section \ref{sec:experiments}. We see across all three datasets that increasing the consensus loss weight $\lambda$ leads to higher pairwise rank agreement between LIME and SHAP. Moreover, even with high values of $\lambda$, the accuracy stays well above linear models indicating that the loss in performance is acceptable. Therefore this experiment supports the hypothesis.

The results plotted in Figure \ref{fig:pareto} demonstrate that a practitioner concerned with agreement can tune $\lambda$ to meet their needs of accuracy and agreement. This figure serves in part to illuminate why our hyperparameter choice is sensible---$\lambda$ gives us control to slide along the trade-off curve, making post hoc explanation disagreement more of a controllable model parameter so that practitioners have more flexibility to make context-specific model design decisions.

\subsection{Are the Explanations Still Valuable?}
\label{sec:explanation-value}

Whether our proposed loss is useful in practice is not completely answered simply by showing accuracy and agreement. A question remains about how our loss might change the explanations in the end. Could we see boosted agreement as a result of some breakdown in how the explainers work? Perhaps models trained with our loss fool explainers into producing uninformative explanations just to appease the agreement term in the loss.

\textbf{Hypothesis:} \emph{We only get consensus trivially, i.e., with feature attributions scores that are uninformative.}

Since we have no ground truth for post hoc feature attribution scores, we cannot easily evaluate their quality \cite{sundararajan2017axiomatic}. 
Instead, we reject this hypothesis with an experiment wherein we add random ``junky'' features to the input data. In this experiment we show that when we introduce junky input features, which by definition have no predictive power, our explainers appropriately attribute near zero importance to them. 

Our experimental design is related to other efforts to understand explainers. \citet{fooling} demonstrate an experimental setup whereby a model is built with ground-truth knowledge that one feature is the only important feature to the model, and the other features are unused. They then adversarially attack the model-explainer pipeline and measure the frequency with which their explainers identify one of the truthfully unimportant features as the most important.
Our tactic works similarly, since a naturally trained model will not rely on random features which have no predictive power.

We measure the frequency with which our explainers place one of the junk features in the top-$k$ most important features, using $k=5$ throughout.

As a representative example, LIME explanations of MLPs trained on this augmented Electricity data put random features in the top five 11.8\% of the time on average. If our loss was encouraging models to permit uninformative explanations for the sake of agreement, we might see this number rise. However, when trained with $\lambda=0.5$, random features are only in the top five LIME features 9.1\% of the time -- and random chance would have at least one junk feature in the top five over 98\% of the time. For results on all three datasets and all six expalainers, see Appendix \ref{sec:app-junk}. 

The setting where junk features are most often labelled as one of the top five is when using SmoothGrad to explain models trained on Bank Marketing data with $\lambda = 0$, where for 43.1\% of the samples, at least one of the top five is in fact a junk feature. 
Interestingly, for the same explainer and dataset models trained with $\lambda = 0.5$ lead to explanations that have a junk feature as one of the top five less than 1\% of the time, indicating that our loss can even improve this behavior in some settings.

Therefore, we reject this hypothesis and conclude that the explanations are not corrupted by training with our loss.

\subsection{Consensus and Linearity}
\label{sec:linearity}

Since linear models are the gold standard in model explainability, one might wonder if our loss is pushing models to be more like linear models. 
We conduct a quantitative and qualitative test to see whether our method indeed increases linearity. 

\textbf{Hypothesis:} \emph{Encouraging explanation consensus during training encourages linearity.}

\begin{figure}
    \centering
    \begin{tabular}{cccc}
       $\lambda$ = 0.00& $\lambda$ = 0.75 & $\lambda$ = 0.95 & Linear \\
    \includegraphics[width=0.2\columnwidth]{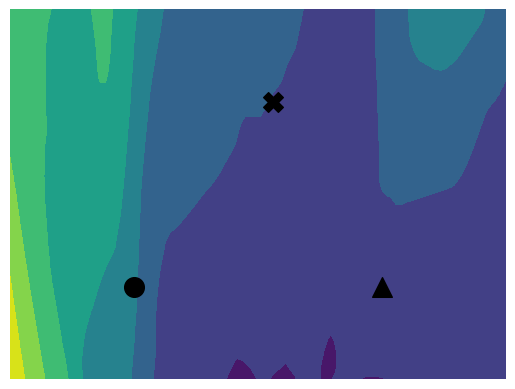} &
    \includegraphics[width=0.2\columnwidth]{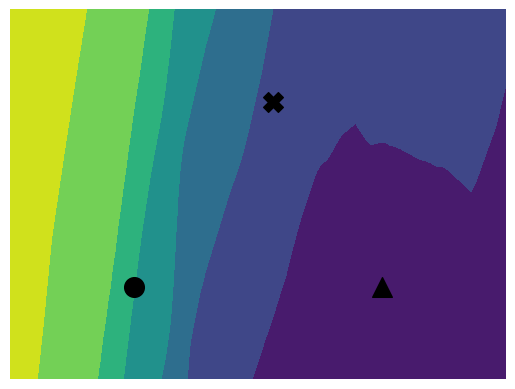} &
    \includegraphics[width=0.2\columnwidth]{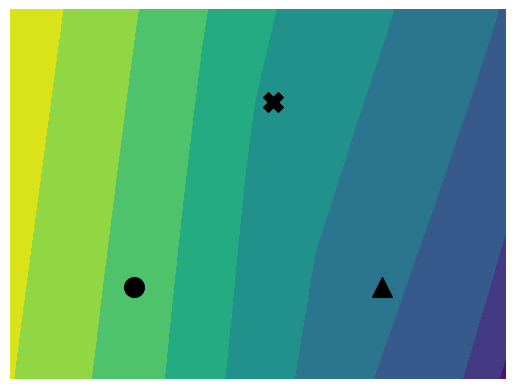} &
    \includegraphics[width=0.2\columnwidth]{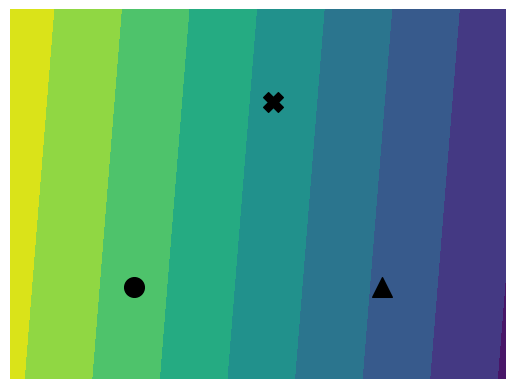}
    \end{tabular}
    \caption{Logit surface contour plots on a plane spanning three real data points from four different models. Left to right: MLPs trained with $\lambda$ = 0, $\lambda$ = 0.75 and $\lambda$ = 0.95 as well as a linear model. Notice that as we increase $\lambda$, and move from left to right, we get straighter contours in the logit surface.}
    \label{fig:planes}
\end{figure}

\textbf{Qualitative analysis.} In their work on model reproducibility, \citet{somepalli2022can} describe a visualization technique wherein a high-dimensional decision surface is plotted in two dimensions. Rather than more complex distance preserving projection tactics, they argue that the subspace of input space defined by a plane spanning three real data points can be a more informative way to visualize how a model's outputs change in high dimensional input space. We take the same approach to study how the logit surface of our model changes with $\lambda$. We take three random points from the test set, and interpolate between the three of them to get a planar slice of input space. We then compute the logit surface on this plane (we arbitrarily choose the logit corresponding to the first class). We visualize the contour plots of the logit surface in Figure \ref{fig:planes} (more visualizations in
Section \ref{sec:app-three-point}). As we increase $\lambda$, we see that the shape of the contours often tends toward the contour pattern that a linear model takes on that same plane slice of input space.

\begin{figure}
    \centering
    \includegraphics[width=\columnwidth]{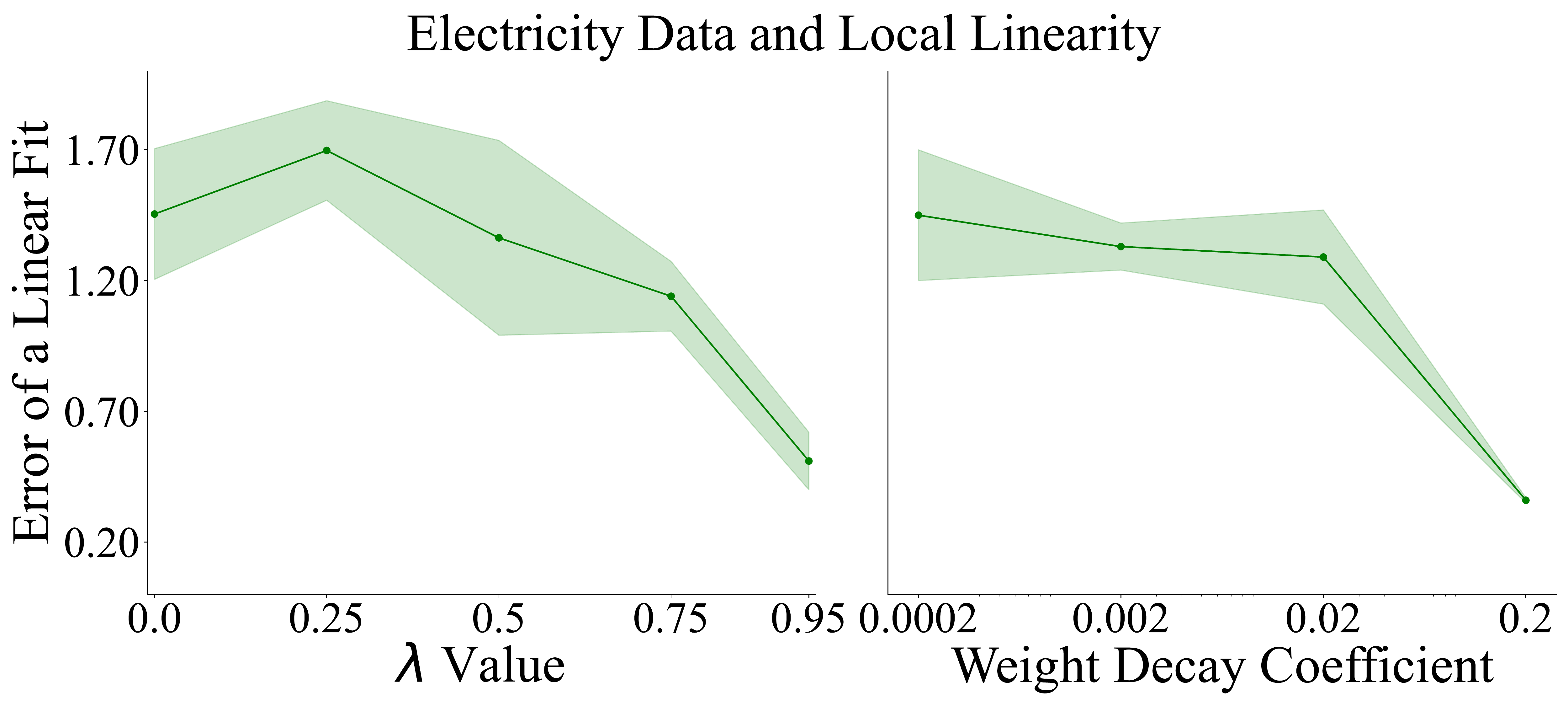}
    \caption{Sampled linear-best-fit error (MAE) measurements as $\lambda$ increases and as the weight decay coefficient increases. Both approaches lead to lower error of a linear approximation on the Electricity dataset. This indicates that both weight decay and consensus training are correlated with linear fit. The shaded region corresponds to $\pm$ one standard error.}
    \label{fig:linearity}
\end{figure}

\textbf{Quantitative analysis.} We can also measure how close to linear a model is quantitatively. The extent to which our models trained with higher $\lambda$ values are close to linear can be measured as follows. For each of ten random planes in input space (constructed using the three-point method described above), we fit a linear regression model to predict the logit value at each point of the plane, and measure the mean absolute error. The closer this error term is to zero, the more our model's logits on this input subspace resemble a linear model. In Figure~\ref{fig:linearity} we show the error values of the linear fit drop as we increase the weight on the consensus loss for the Electricity dataset. Thus, these analyses support the hypothesis that encouraging consensus encourages linearity.

But if our consensus training pushes models to be closer to linear, does any method that increases the linearity measurement also lead to increased consensus?
We consider the possibility that any approach to make models closer to linear improves consensus metrics. 

\textbf{Hypothesis:} \emph{Linearity implies more explainer consistency.}

To explore another path toward more linear models, we train a set of MLPs without our consensus loss but with various weight decay coefficients. In Figure \ref{fig:linearity}, we show a drop in linear-best-fit error across the random three-point planes which is similar to the drop observed by increasing $\lambda$, showing that increasing weight decay also encourages models to be closer to linear.

But when evaluating these MLPs with increasing weight decay by their consensus metrics, they show near-zero improvement. We therefore reject this hypothesis---linearity alone does not seem to be enough to improve consensus on post hoc explanations.

\begin{figure}
    \centering
    \includegraphics[width=\columnwidth]{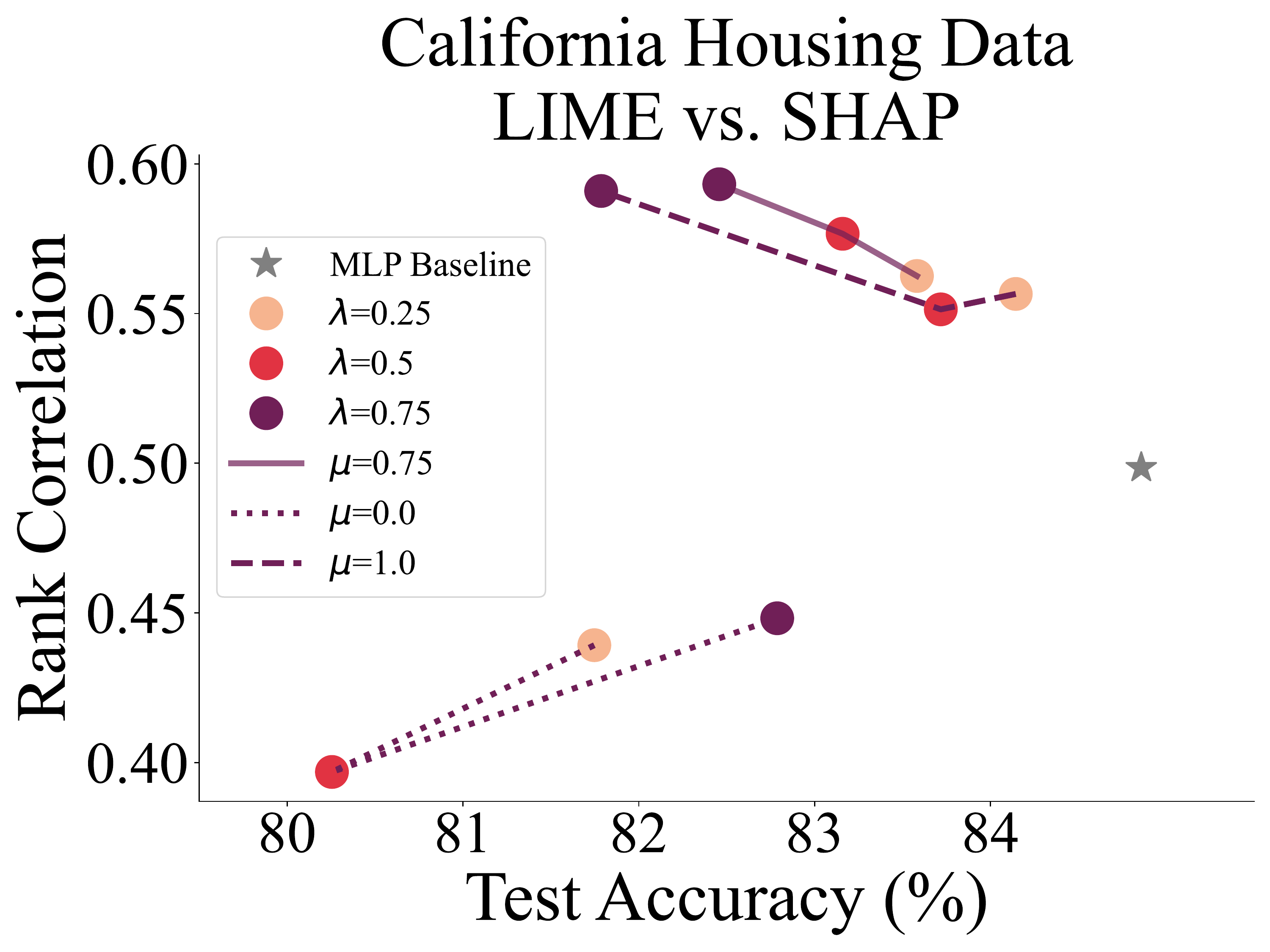}
    \caption{We perform an ablation study of our loss term parameter $\mu$ to show why, when training to improve correlation between feature attribution scores, using both Spearman and Pearson correlation can be better than using just one type of correlation.}
    \label{fig:busy-pareto}
\end{figure}

\subsection{Two Loss Terms}
\label{sec:ablations}

For the majority of experiments, we set $\mu=0.75$, which is determined by a coarse grid search. And while it may not be optimal for every dataset on every agreement metric, we seek to show that the extreme values $\mu=0$ and $\mu=1$, which each correspond to only one correlation term in the loss, can be suboptimal. This ablation study serves to justify our choice of incorporating two terms in the loss. In Figure~\ref{fig:busy-pareto}, we show the agreement-accuracy trade-off for multiple values of $\mu$ and of $\lambda$. We see that $\mu = 0.75$ shows the more optimal trade-off curve. 

In Appendix \ref{sec:app-three-point}, where we show more plots like Figure \ref{fig:busy-pareto} for other datasets and metrics, we see that the best value of $\mu$ varies case by case. This demonstrates the importance of having a tunable parameter within our consensus loss term to be tweaked for better performance.

\section{Discussion}
\label{sec:discussion}

The empirical results we present demonstrate that our loss term is effective in its goal of boosting consensus among explainers. As with any first attempt at introducing a new objective to neural network training, we see modest results in some settings and evidence that hyperparameters can likely be tuned on a case-by-case basis. It is not our aim to leave practitioners with a how-to guide, but rather to begin exploring how practitioners can control where a model lies along the accuracy-agreement trade-off curve.

We introduce a loss term measuring two types of correlation between explainers, which unfortunately adds more complexity to the machine learning engineer's job of tuning models. But, we show conclusively that there are settings in which using both types of correlation is better than using only one when encouraging explanation consensus.

Another limitation of these experiments as a guide on how to train for consensus is that we only trained with one pair of explainers. Our loss is defined for any pair and perhaps another choice would better suit specific applications. 

In light of the contentious debate on whether deep models or decision-tree-based methods are better for tabular data \citep{gorishniy2021revisiting, shwartz2022tabular, bojan}, we argue that developing new tools for training deep models can help promote wider adoption for tabular deep learning. Moreover, with the results we present in this work, it is our hope that future work improves these trends, which could possibly lead to neural models that have more agreement (and possibly more accuracy) than their tree-based counterparts (such as XGBoost).

\subsection{Future Work}
\label{sec:future-work}

Armed with the knowledge that training for consensus with PEAR is possible, we describe several exciting directions for future work. 
First, as alluded to above, we explored training with only one pair of explainers, but other pairs may help data scientists who have a specific type of target agreement. 
Work to better understand how a given pair of explainers in the loss affects the agreement of other explainers at test time could lead to principled decisions about how to use our loss in practice.
Indeed, PEAR could fit into larger learning frameworks~\citep{Nanda21:LearningHCXAI} that aim to select user- and task-specific explanation methods automatically.

It will be crucial to study the quality of explanations produced with PEAR from a human perspective. Ultimately, both the efficacy of a single explanation and the efficacy of agreement between multiple explanations is tied to how the explanations are used and interpreted. Since our work only takes a quantitative approach to demonstrate improvement when regularizing for explanation consensus, it remains to be seen whether actual human practitioners would make better judgments about models trained with PEAR vs models trained naturally.

In terms of model architecture, we chose standard sized MLPs for the experiments on our tabular datasets. Recent work proposes transformers \citep{somepalli2021saint} and even ResNets \cite{gorishniy2021revisiting} for tabular data, so completely different architectures could also be examined in future work as well.

Finally, research into developing better explainers could lead to an even more powerful consensus loss term. Recall that IntGrad integrates the gradients over a path in input space. The designers of that algorithm point out that a straight path is the canonical choice due to its simplicity and symmetry \citep{sundararajan2017axiomatic}. Other paths through input space that include more realistic data points, instead of paths of points constructed via linear interpolation, could lead to even better agreement metrics on actual data.

\subsection{Conclusion}
\label{sec:conclusion}

In the quest for fair and accessible deep learning, balancing interpretability and performance are key. It is known that common explainers may return conflicting results on the same model and input, to the detriment of an end user. The gains in explainer consensus we achieve with our method, however modest, serve to kick start others to improve on our work in aligning machine learning models with the practical challenge of interpreting complex models for real-life stakeholders.

\section*{Acknowledgements}

We thank Teresa Datta and Daniel Nissani at Arthur for their insights throughout the course of the project. We also thank Satyapriya Krishna, one of the authors of the original Disagreement Problem paper, for informative email exchanges that helped shape our experiments.




{
\small
\bibliography{main}
\bibliographystyle{plainnat} 
}

\newpage
\appendix
\onecolumn

\section{Appendix}
\label{sec:appendix}

\subsection{Datasets}
\label{sec:app-datasets}

In our experiments we use tabular datasets originally from OpenML and compiled into a set of benchmark datasets from the Inria-Soda team on \href{https://huggingface.co/datasets/inria-soda/tabular-benchmark}{HuggingFace} \citep{Grinsztajn22:Why}. 
We provide some details about each dataset:

\textbf{Bank Marketing} This is a binary classification dataset with six input features and is approximately class balanced. We train on 7,933 training samples and test on the remaining 2,645 samples.

\textbf{California Housing} This is a binary classification dataset with seven input features and is approximately class balanced. We train on 15,475 training samples and test on the remaining 5,159 samples.

\textbf{Electricity} This is a binary classification dataset with seven input features and is approximately class balanced. We train on 28,855 training samples and test on the remaining 9,619 samples.

\subsection{Hyperparamters}
\label{sec:app-hyperparameters}

Many of our hyperparameters are constant across all of our experiments. For example, all MLPs are trained with a batch size of 64, and initial learning rate of $0.0005$. Also, all the MLPs we study are 3 hidden layers of 100 neurons each. We always use the AdamW  optimizer \citep{loshchilov2017decoupled}. The number of epochs varies from case to case. For all three datasets, we train for 30 epochs when $\lambda \in \{0.0, 0.25\}$ and 50 epochs otherwise. When training linear models, we use 10 epochs and an initial learning rate of 0.1.

\subsection{Disagreement Metrics}
\label{sec:app-metrics}

We define each of the six agreement metrics used in our work here. 

The first four metrics depend on the top-$k$ most important features in each explanation. Let $top\_features(E, k)$ represent the top-$k$ most important features in an explanation $E$, let $rank(E, s)$ be the importance rank of the feature $s$ within explanation $E$, and let $sign(E, s)$ be the sign (positive, negative, or zero) of the importance score of feature $s$ in explanation $E$.

\textbf{Feature Agreement} 
\begin{equation}
    \frac{| top\_features(E_1, k) \cap top\_features(E_2, k) |}{k}
\end{equation}

\textbf{Rank Agreement} 
\begin{equation}
\frac{| \bigcup_{s \in S} \{s \in top\_features(E_1, k) \wedge s \in top\_features(E_2, k) \wedge rank(E_1, s) = rank(E_2, s) \}| }{k}
\end{equation}

\textbf{Sign Agreement} 
\begin{equation}
\frac{| \bigcup_{s \in S} \{s \in top\_features(E_1, k) \wedge s \in top\_features(E_2, k) \wedge sign(E_1, s) =signrank(E_2, s) \}| }{k}
\end{equation}

\textbf{Signed Rank Agreement}
\begin{equation}\frac{| \bigcup_{s \in S} \{s \in top\_features(E_1, k) \wedge s \in top\_features(E_2, k) \wedge rank(E_1, s) = rank(E_2, s) \wedge sign(E_1, s) = sign(E_2, s) \}| }{k}
\end{equation}

The next two agreement metrics depend on all features within each explanation, not just the top-$k$. Let $R$ be a function that computes the ranking of features within an explanation by importance.

\textbf{Rank Correlation}
\begin{equation}
    \sum_i \frac{(R(a)_i - \overline{R(a)})(R(b)_i - \overline{R(b)})}{\Vert R(a) \Vert \Vert R(b) \Vert}
    \label{eq:app-spearman}
\end{equation}

Lastly, let $RelR(E, f_i, f_j)$ be a relative ranking function that returns 1 when feature $f_i$ has higher importance than feature $f_j$ in explanation $E$, and let $F$ be any set of features.

\textbf{Pairwise Rank Agreement}
\begin{equation}
\frac{\sum_{i < j} \mathbb{1} [RelR(E_1, f_i, f_j) = RelR(E_2, f_i, f_j)]}{\binom{|F|}{2}}
\end{equation}

(Note: \citet{krishna2022disagreement} specify in their paper that $F$ is to be a set of features specified by an end user, but in our experiments we use all features with this metric).

\subsection{Junk Feature Experiment Results}
\label{sec:app-junk}

When we add random features for the experiment in Section \ref{sec:explanation-value}, we double the number of features. We do this to check whether our consensus loss damages explanation quality by placing irrelevant features in the top-$K$ more often than models trained naturally. In Table \ref{tab:app-junk}, we report the percentage of the time that each explainer included one of the random features in the top-5 most important features. We observe that across the board, we do not see a systematic increase of these percentages between $\lambda = 0.0$ (a baseline MLP without our consensus loss) and $\lambda = 0.5$ (an MLP trained with our consensus loss).

\begin{table*}[ht!]
\centering
\caption{Frequency of junk features getting top-$5$ ranks, measured in percent.}
\label{tab:app-junk}
\begin{tabular}{llccccccc}
\toprule
                                                                               &               & LIME & SHAP & GRAD & Input*Grad & IntGrad & SmoothGrad & Random Chance \\ 
\midrule
\multirow{2}{*}{\begin{tabular}[c]{@{}l@{}}Bank Marketing\end{tabular}}        & $\lambda=0.0$ & 30.4 & 17.1 & 1.1  & 43.2       & 0.0     & 43.1       & \multirow{2}{*}{98.9}          \\
                                                                               & $\lambda=0.5$ & 25.1 & 12.0 & 0.1  & 34.9       & 0.0     & 0.1                        \\

\midrule
\multirow{2}{*}{\begin{tabular}[c]{@{}l@{}}California  Housing\end{tabular}}   & $\lambda=0.0$ & 22.6 & 8.7  & 0.0  & 24.8       & 0.0     & 0.3        & \multirow{2}{*}{98.5}          \\
                                                                               & $\lambda=0.5$ & 21.2 & 20.4 & 1.4  & 25.9       & 1.4     & 0.9                        \\

\midrule
\multirow{2}{*}{Eelectricity}                                                  & $\lambda=0.0$ & 11.8 & 16.0 & 4.0  & 15.8       & 0.9     & 6.8        & \multirow{2}{*}{98.5}          \\
                                                                               & $\lambda=0.5$ & 9.1  & 9.5  & 1.7  & 8.6        & 0.8     & 3.1                        \\ 
\bottomrule
\end{tabular}
\end{table*}

\subsection{More Disagreement Matrices}
\label{sec:app-redgrids}

\begin{figure*}[ht!]
\centering
\fbox{
\parbox[c]{0.28\textwidth}{
\includegraphics[width=0.13\textwidth]{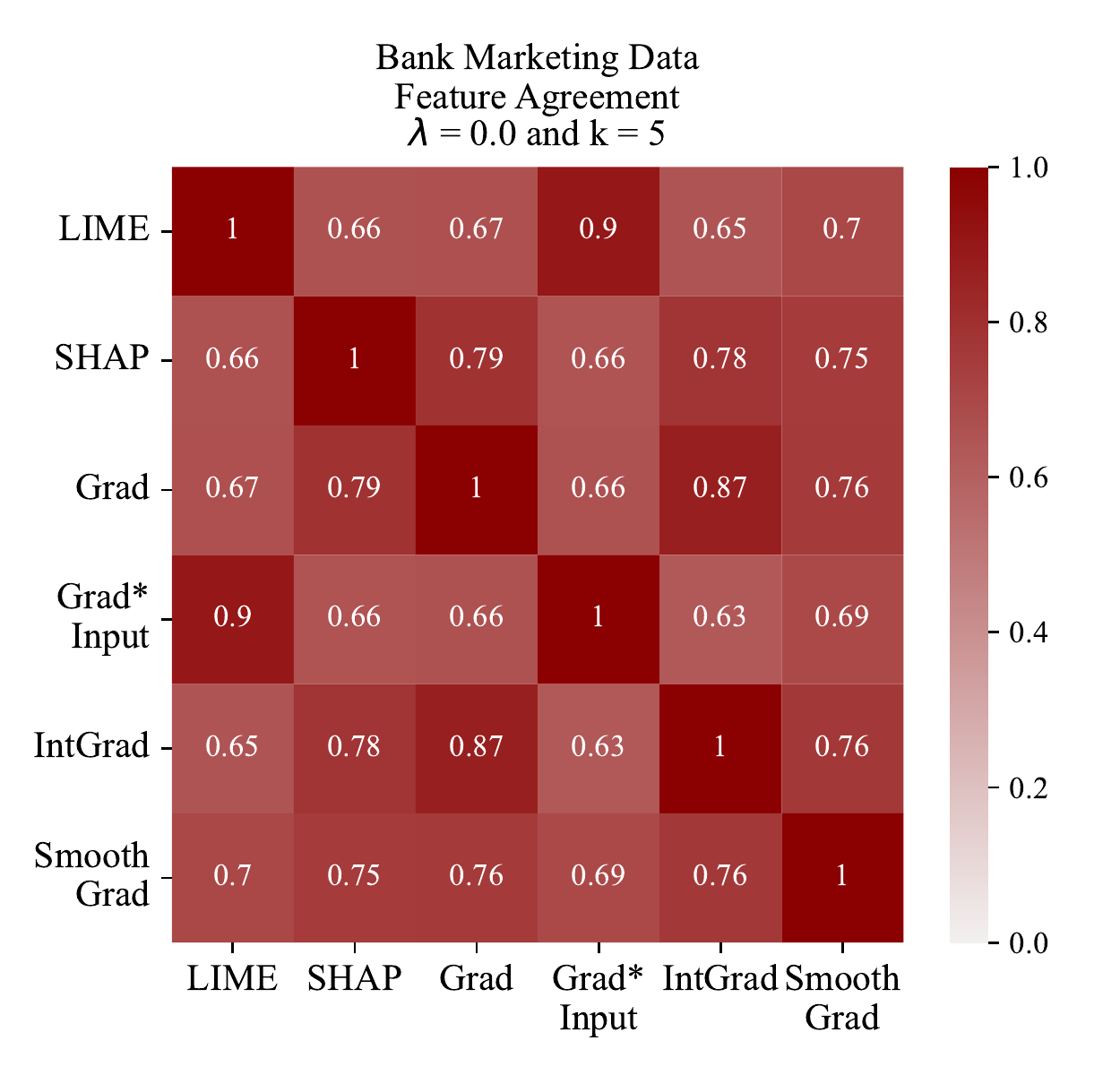}
\includegraphics[width=0.13\textwidth]{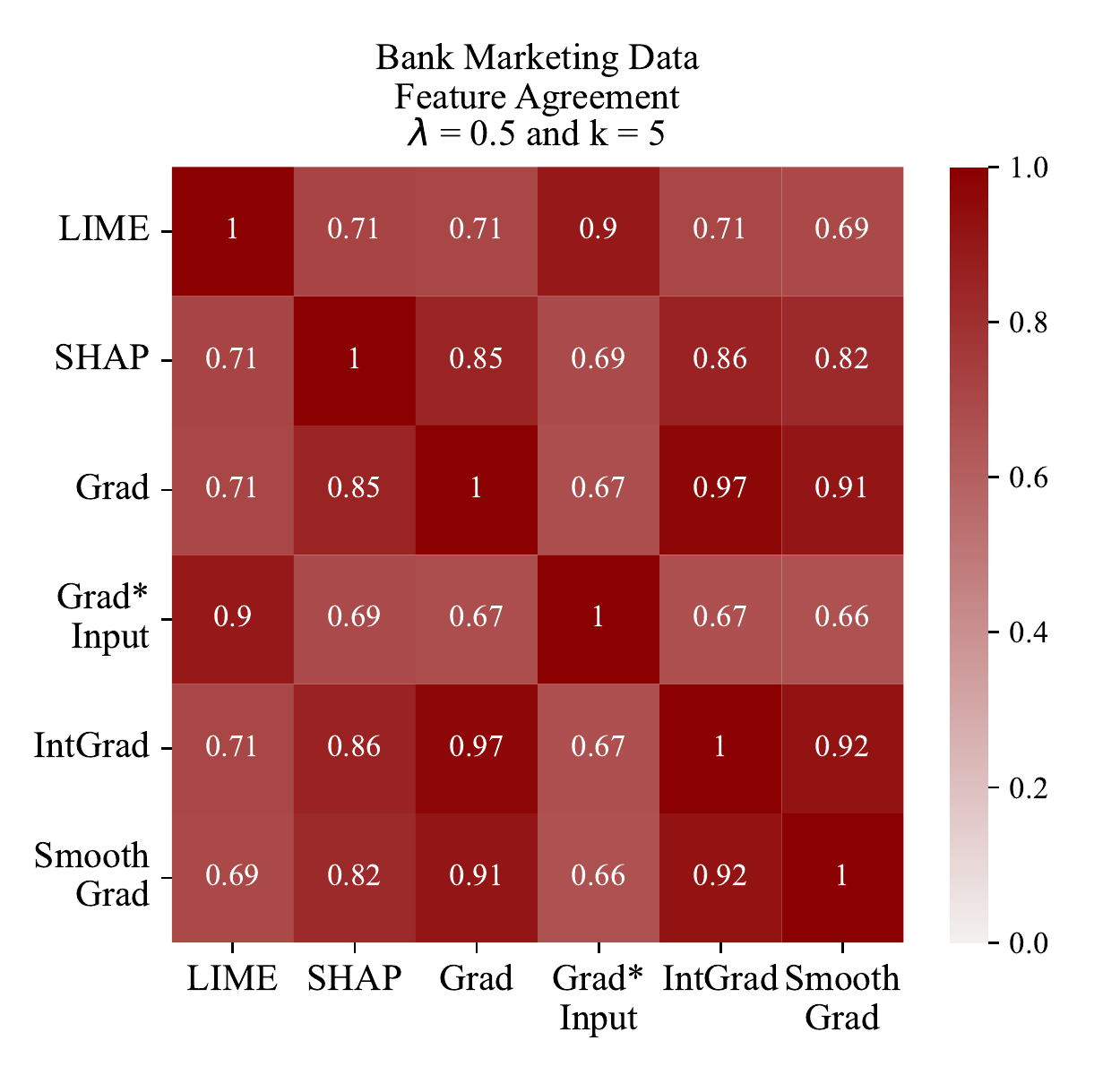}\\
\includegraphics[width=0.13\textwidth]{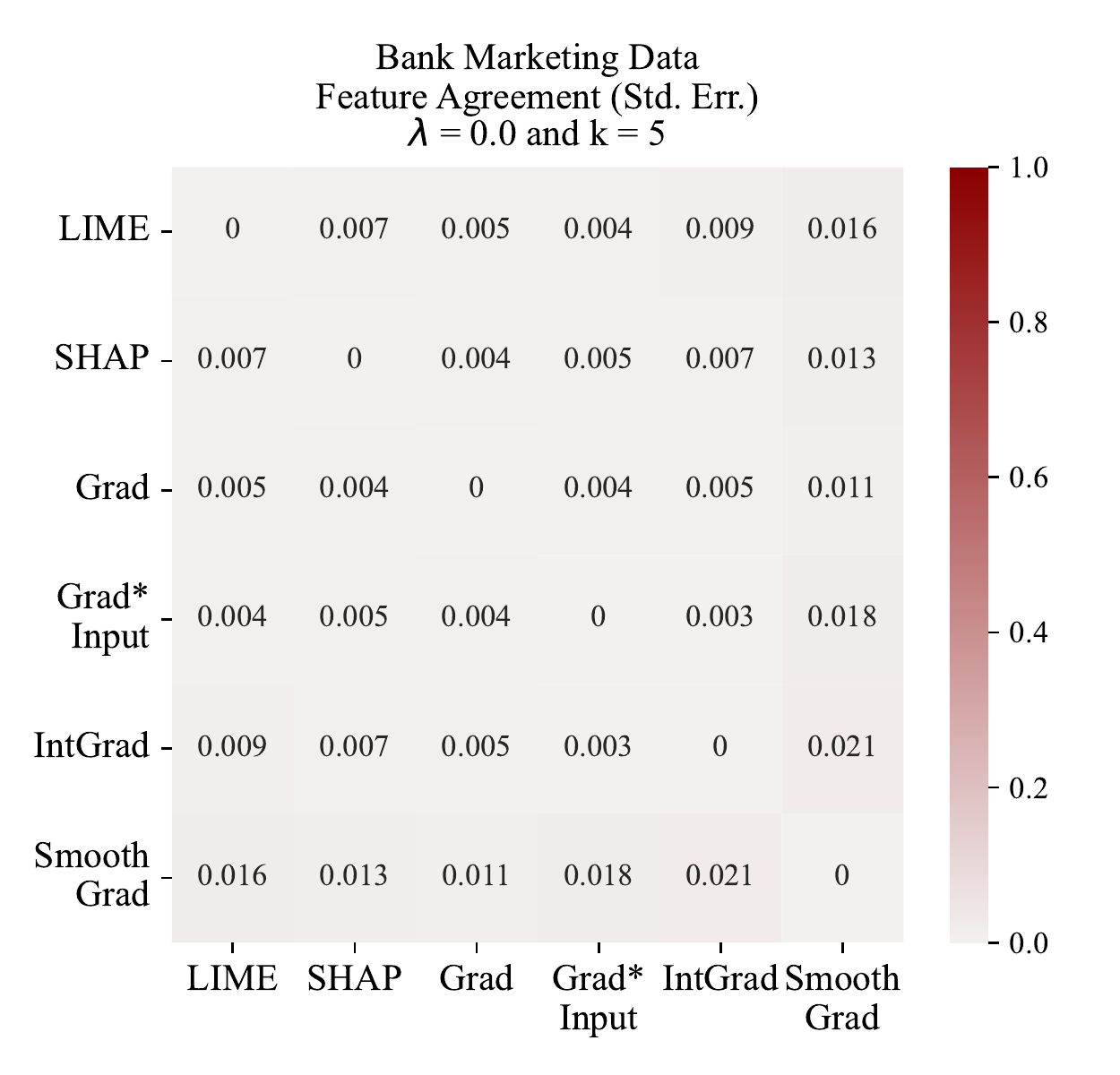}
\includegraphics[width=0.13\textwidth]{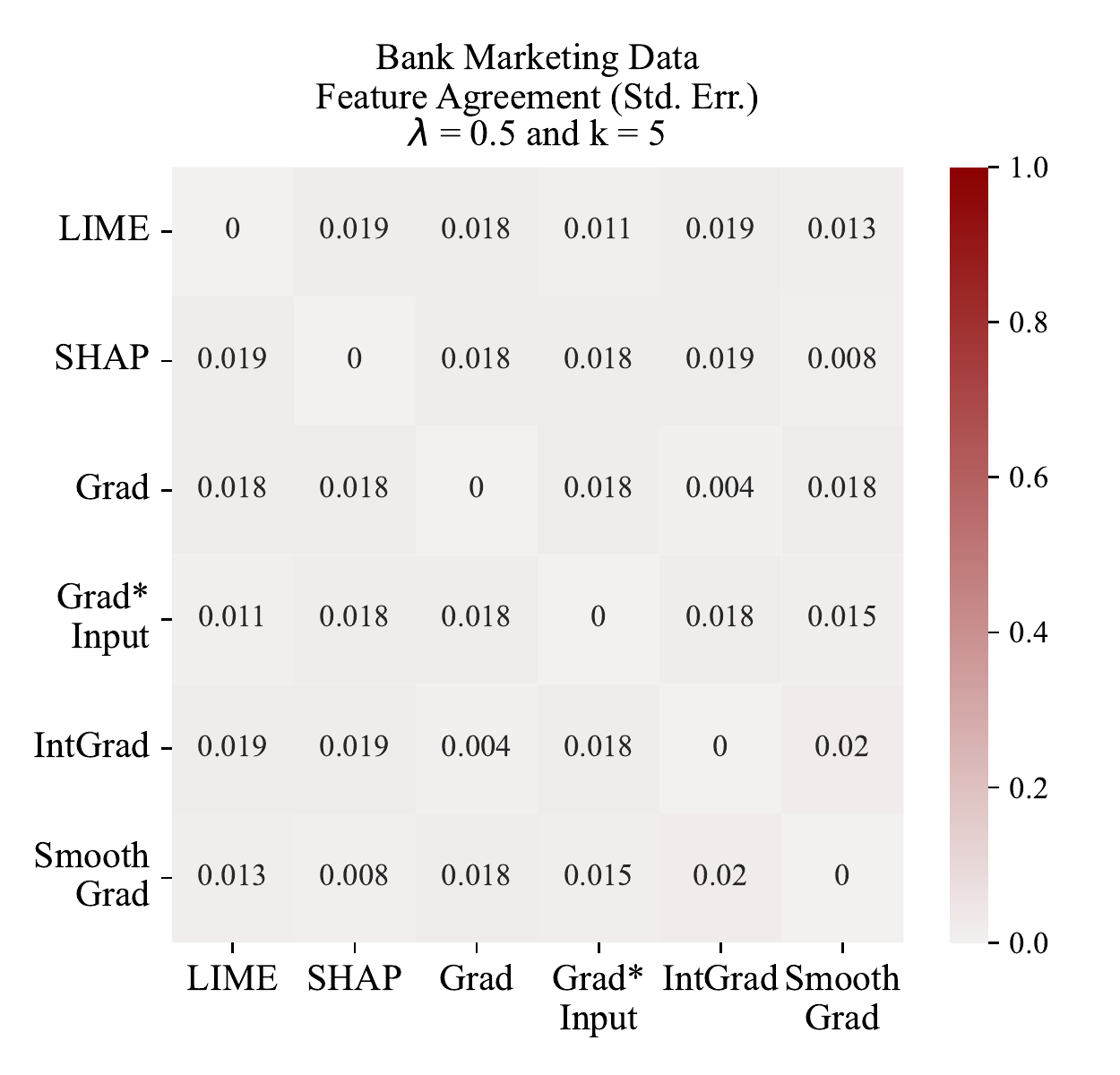}
}}
\fbox{
\parbox[c]{0.28\textwidth}{
\includegraphics[width=0.13\textwidth]{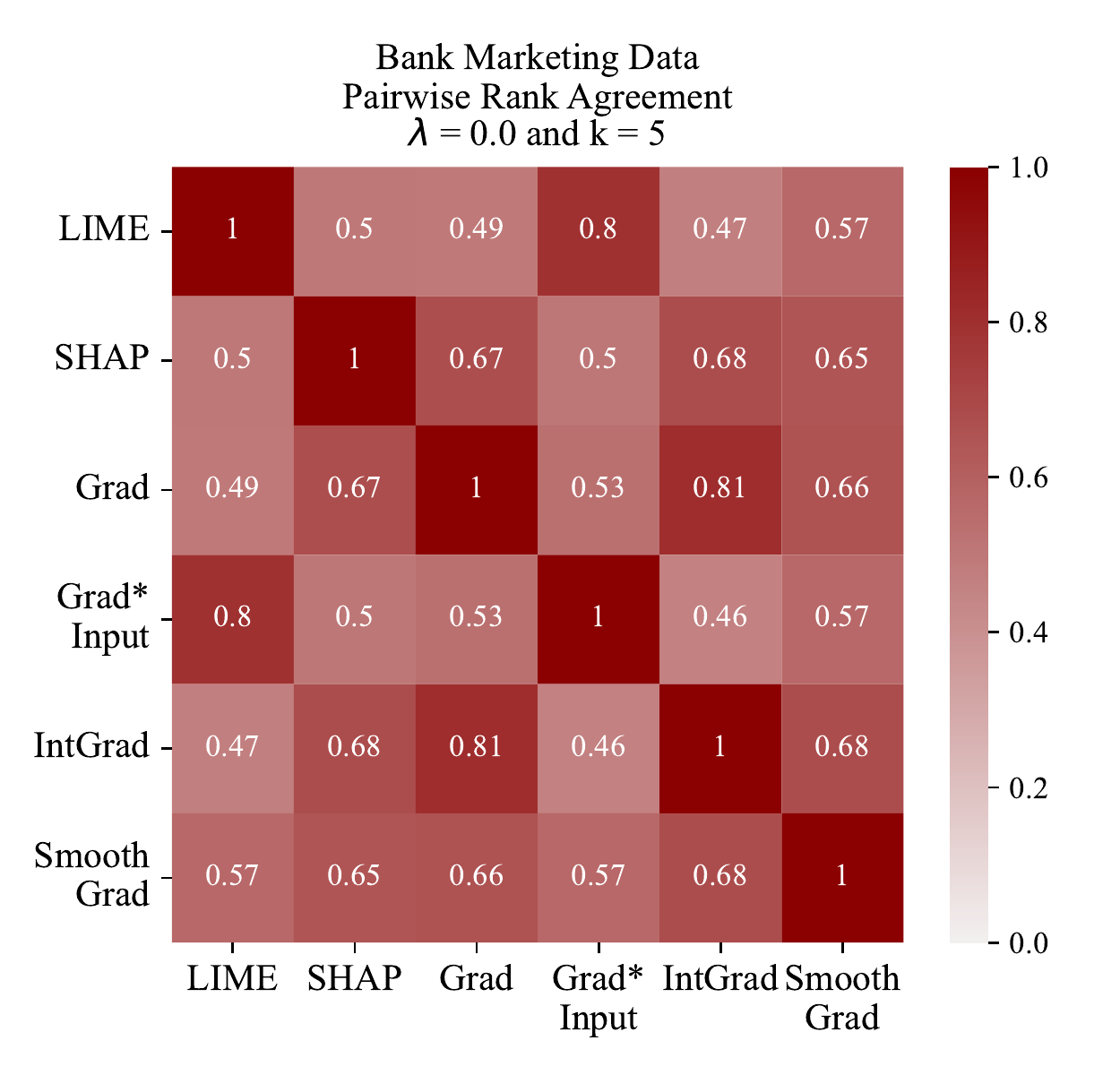}
\includegraphics[width=0.13\textwidth]{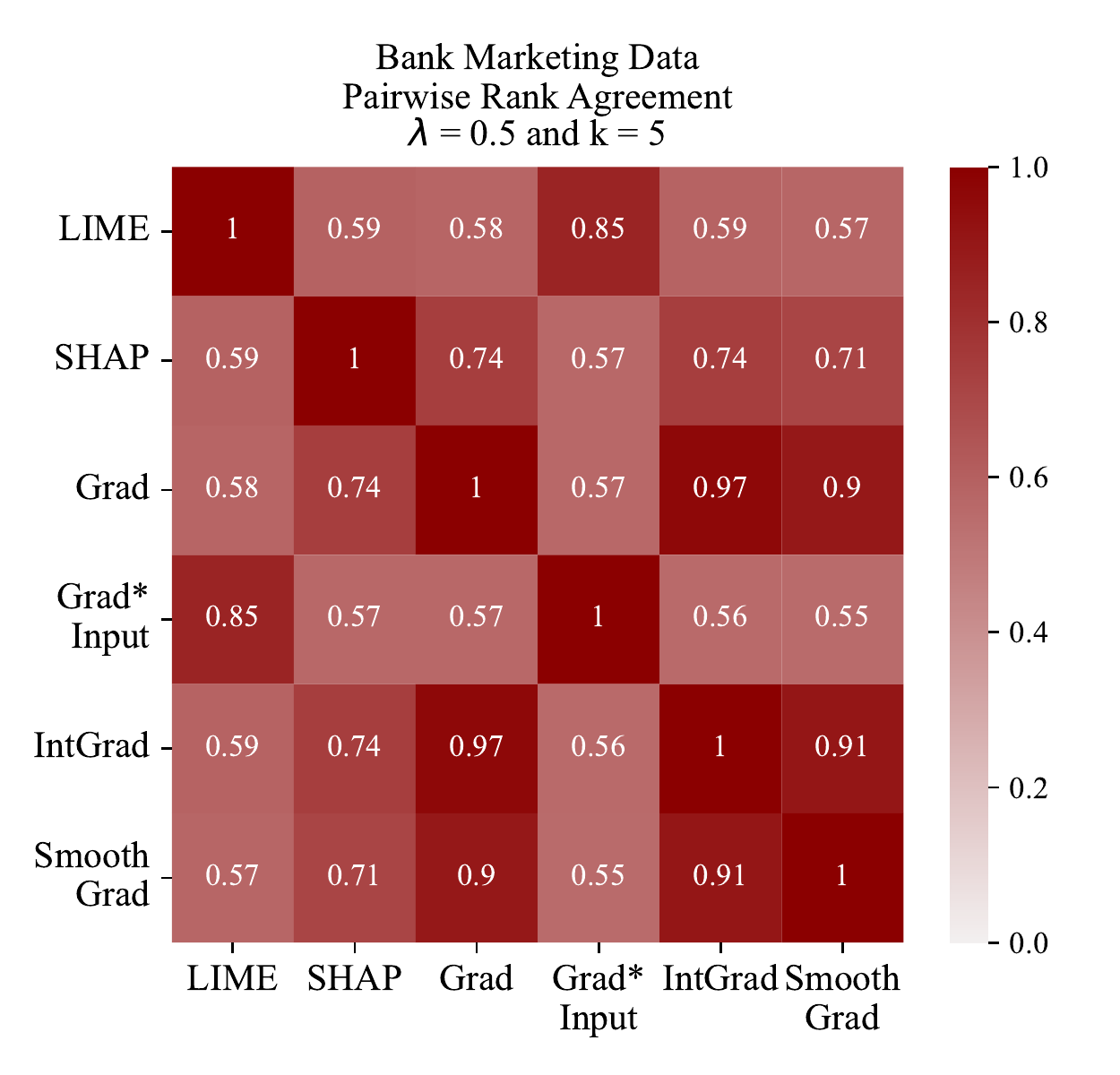}\\
\includegraphics[width=0.13\textwidth]{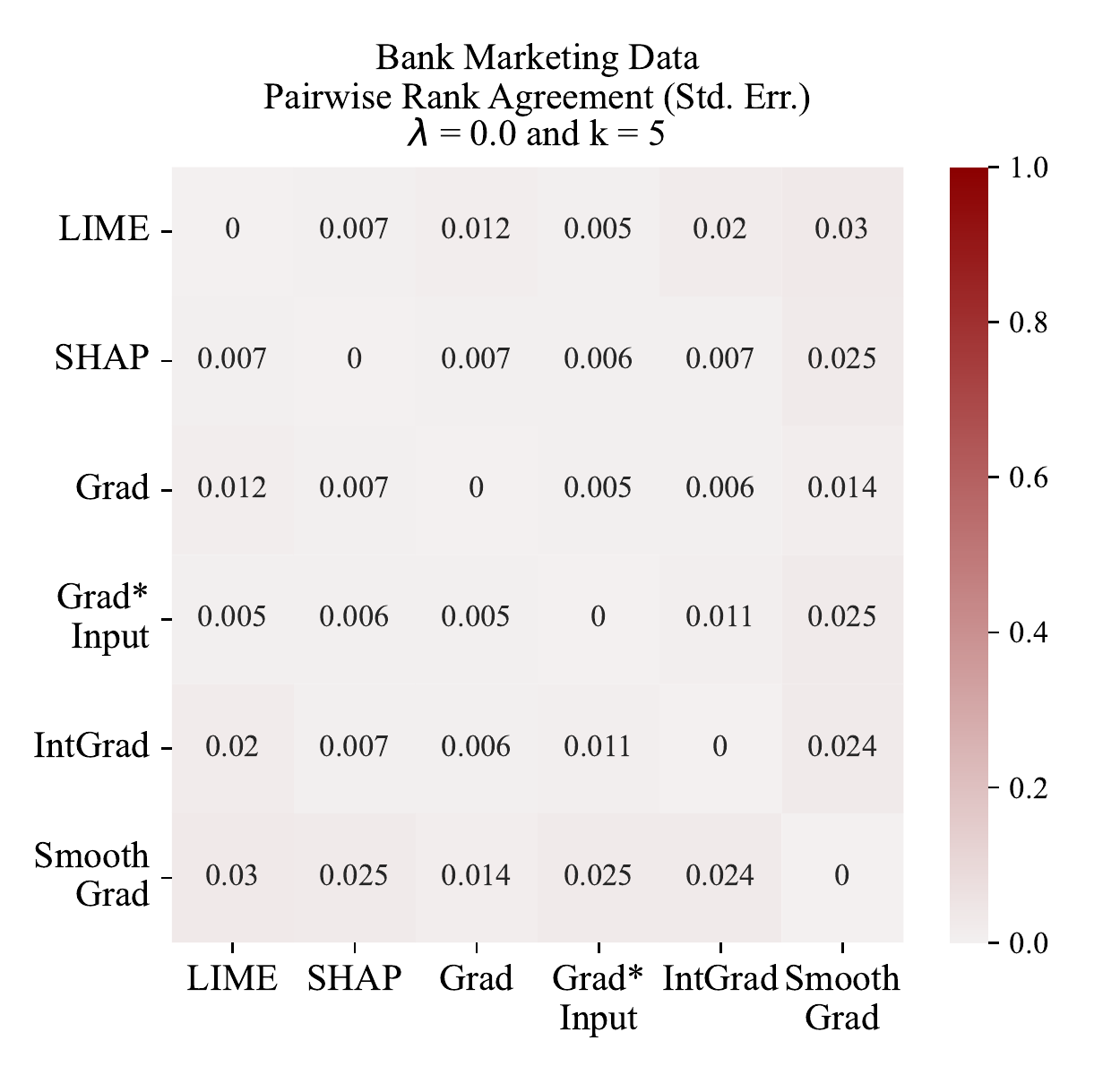}
\includegraphics[width=0.13\textwidth]{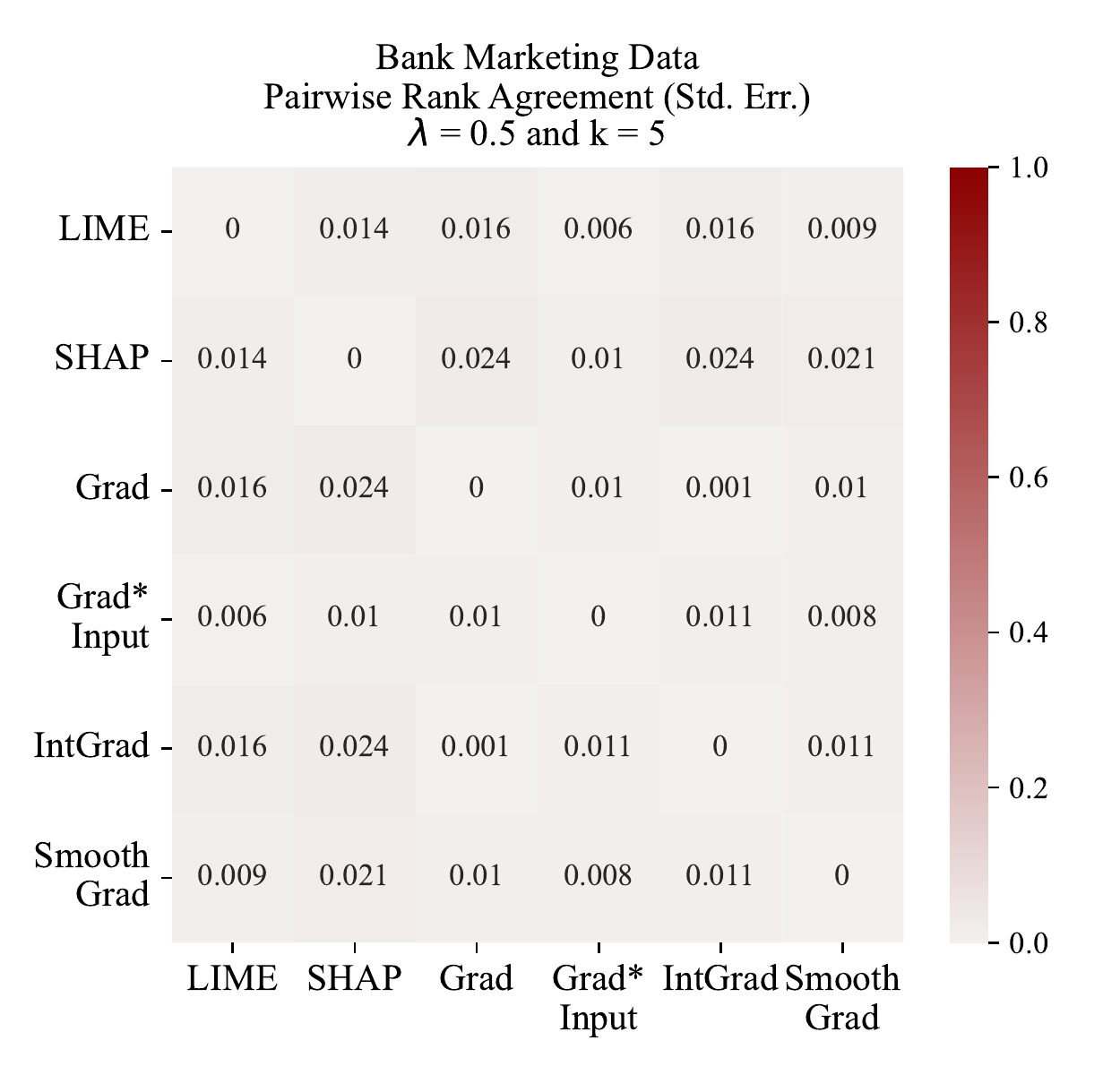}
}}
\fbox{
\parbox[c]{0.28\textwidth}{
\includegraphics[width=0.13\textwidth]{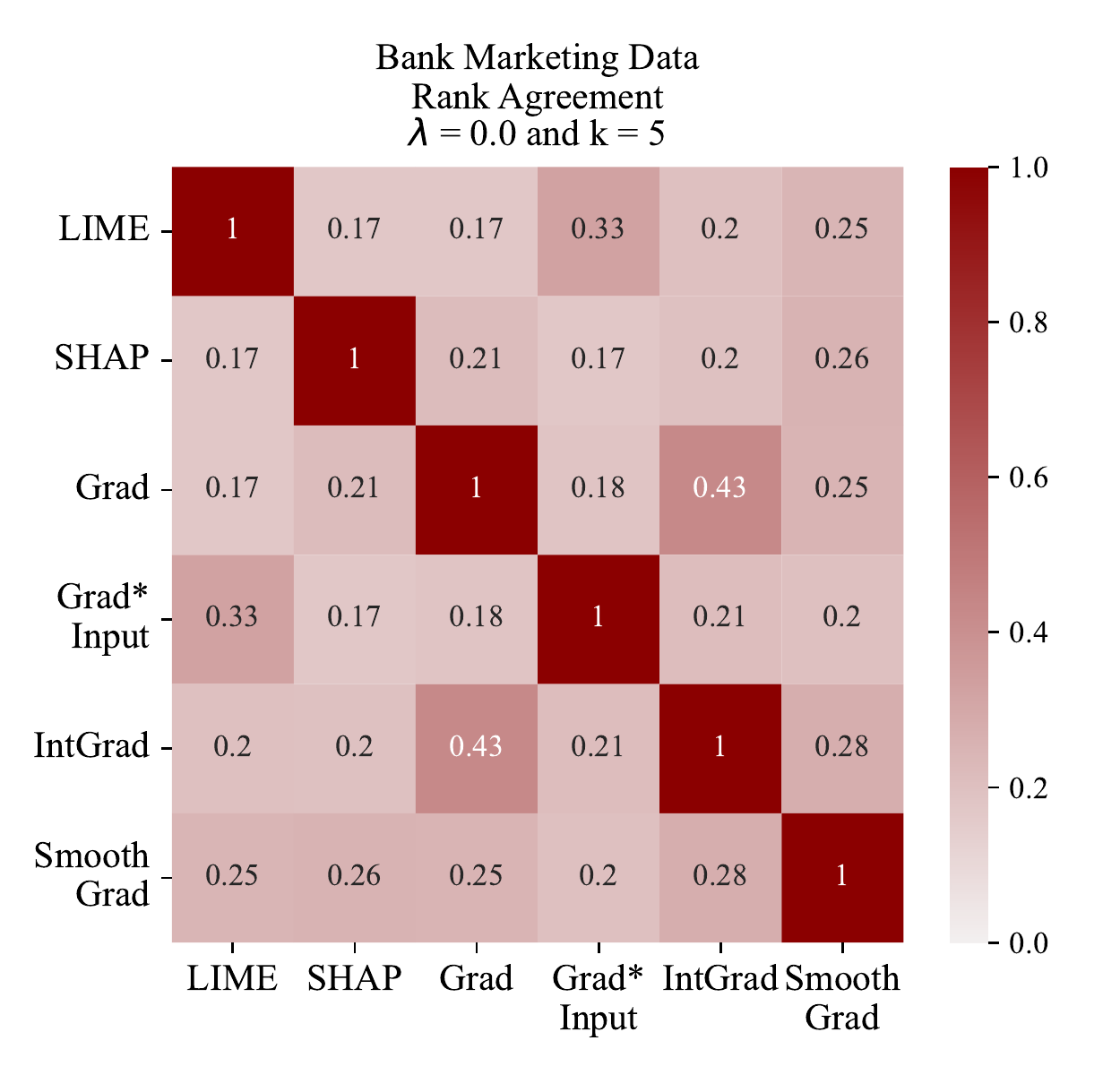}
\includegraphics[width=0.13\textwidth]{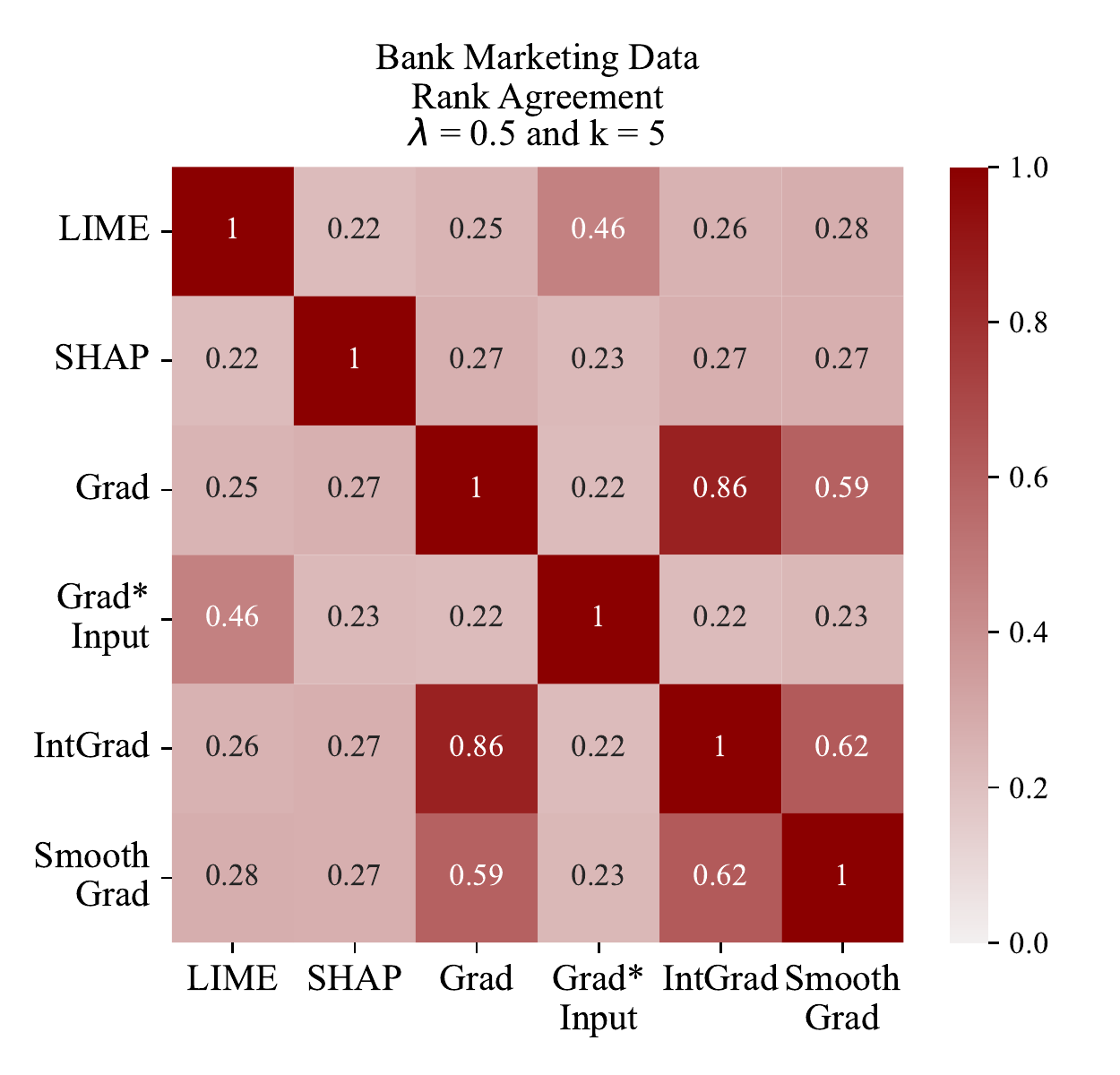}\\
\includegraphics[width=0.13\textwidth]{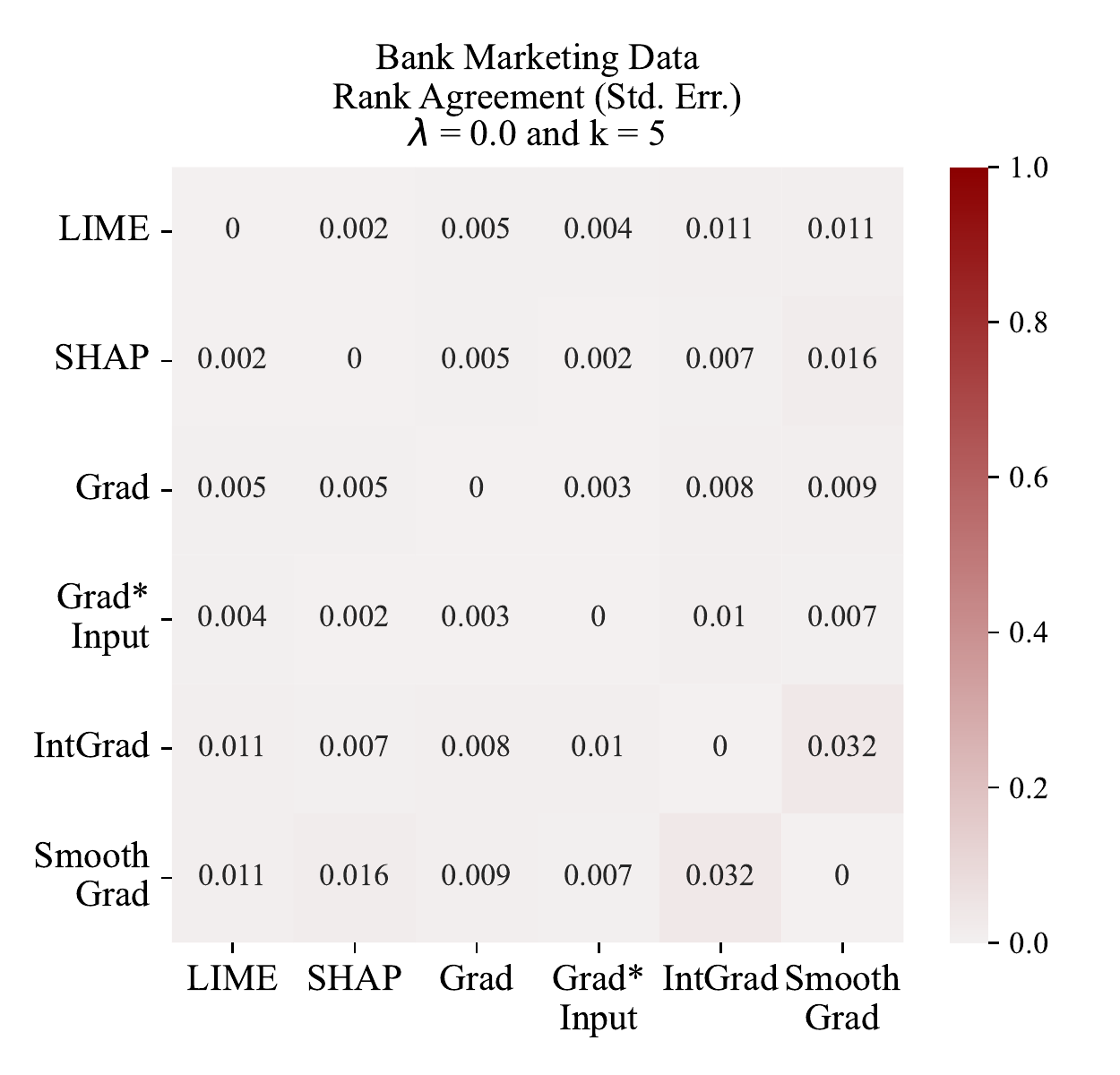}
\includegraphics[width=0.13\textwidth]{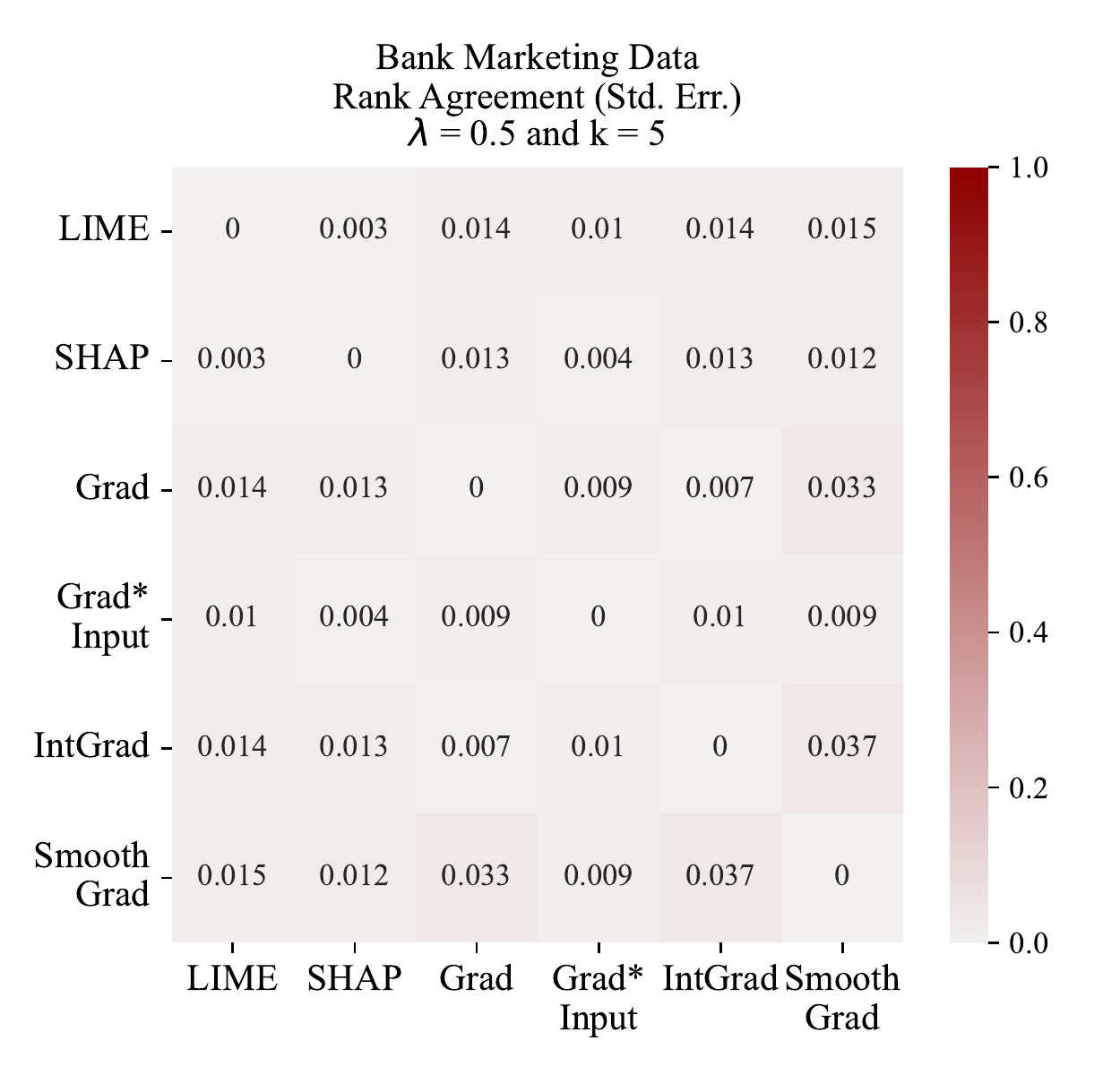}
}}
\\
\fbox{
\parbox[c]{0.28\textwidth}{
\includegraphics[width=0.13\textwidth]{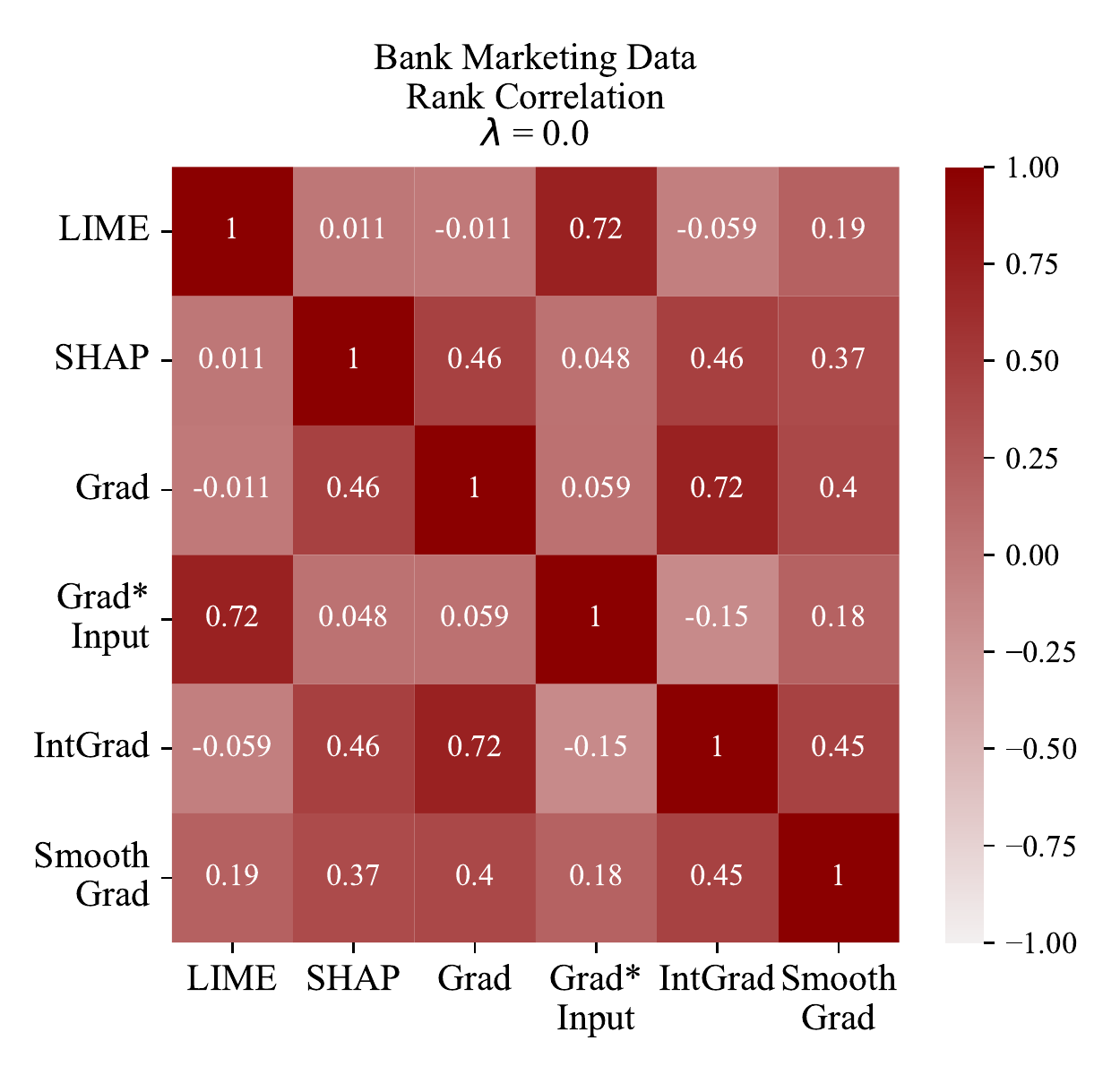}
\includegraphics[width=0.13\textwidth]{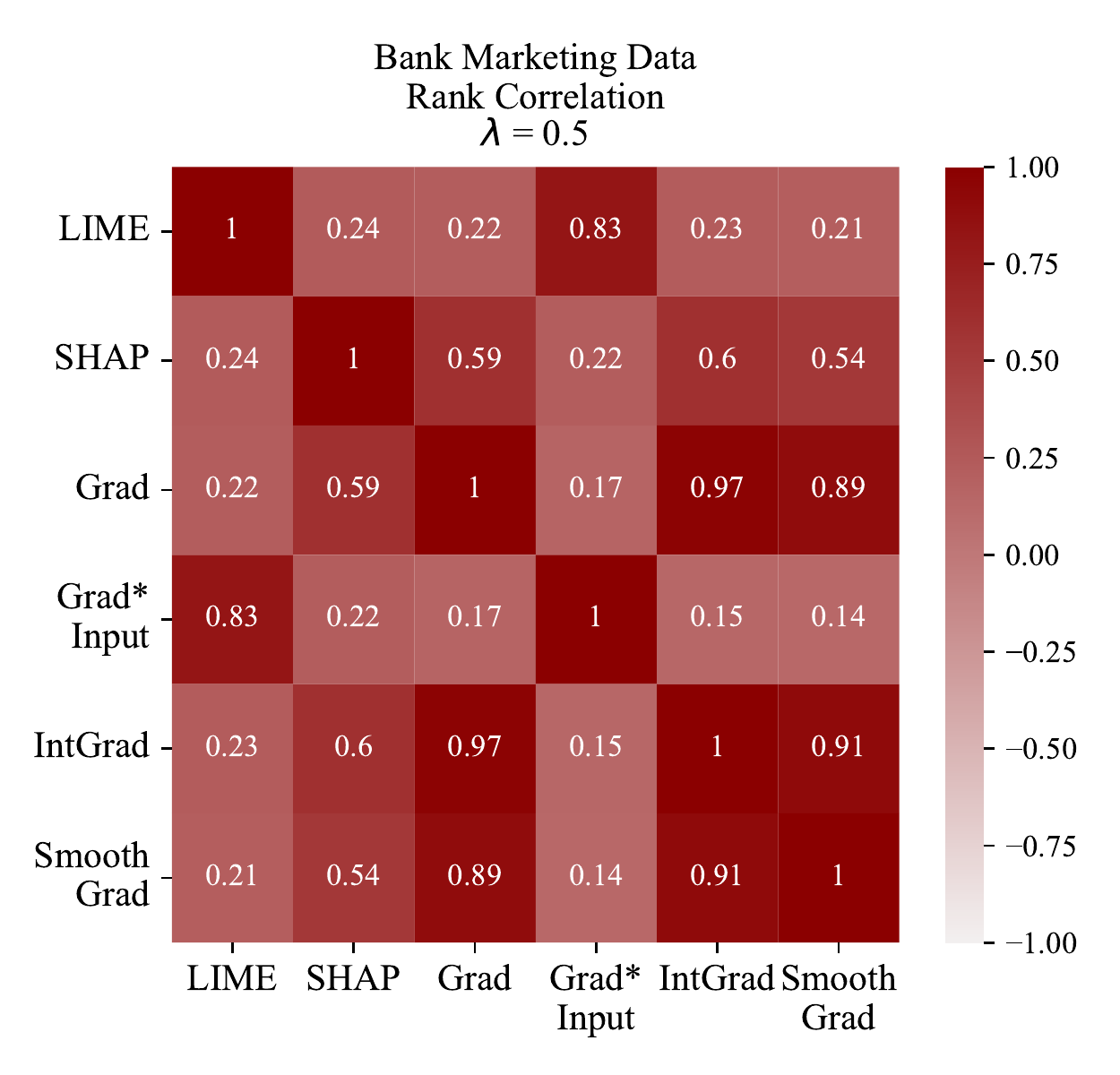}\\
\includegraphics[width=0.13\textwidth]{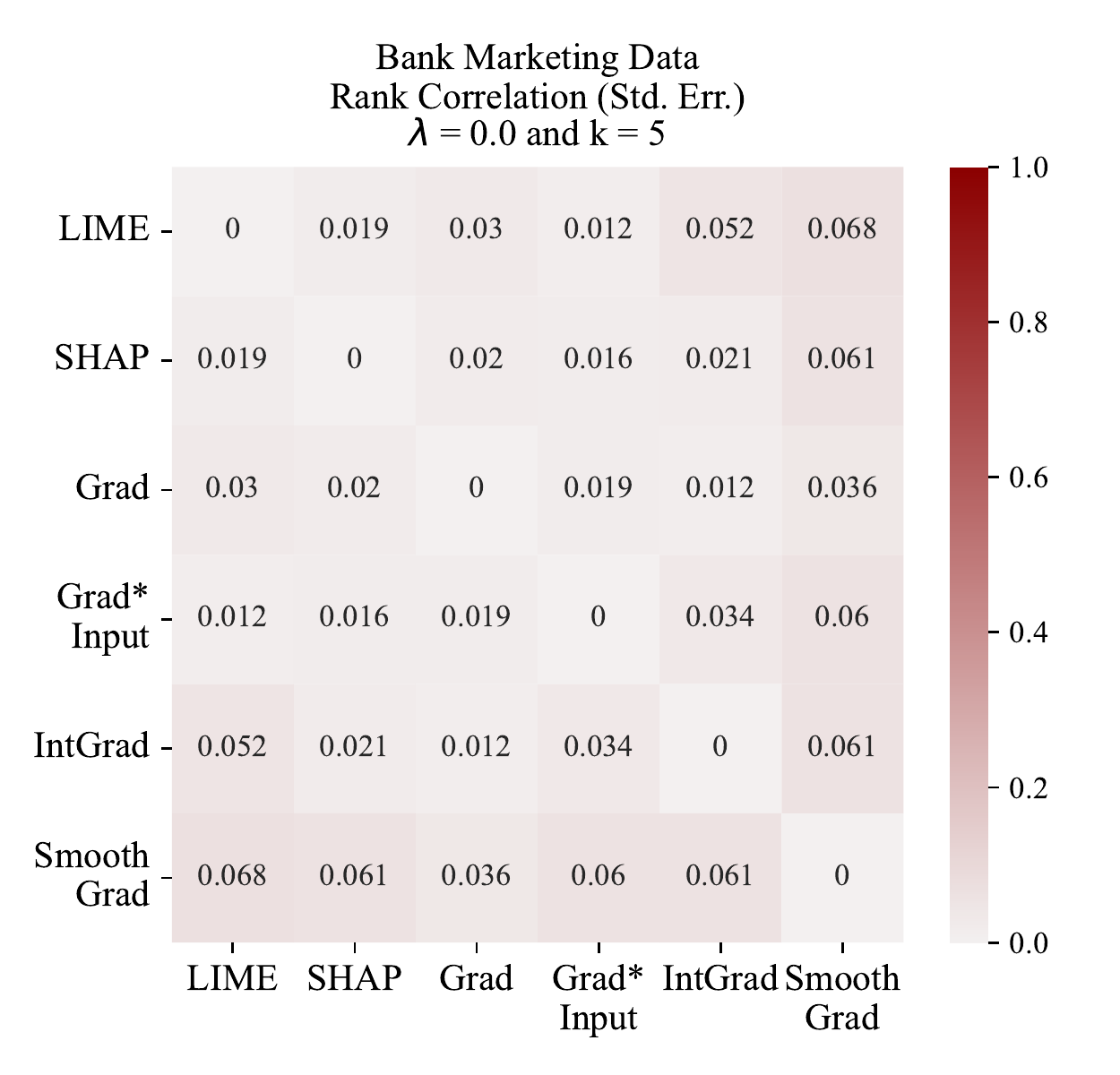}
\includegraphics[width=0.13\textwidth]{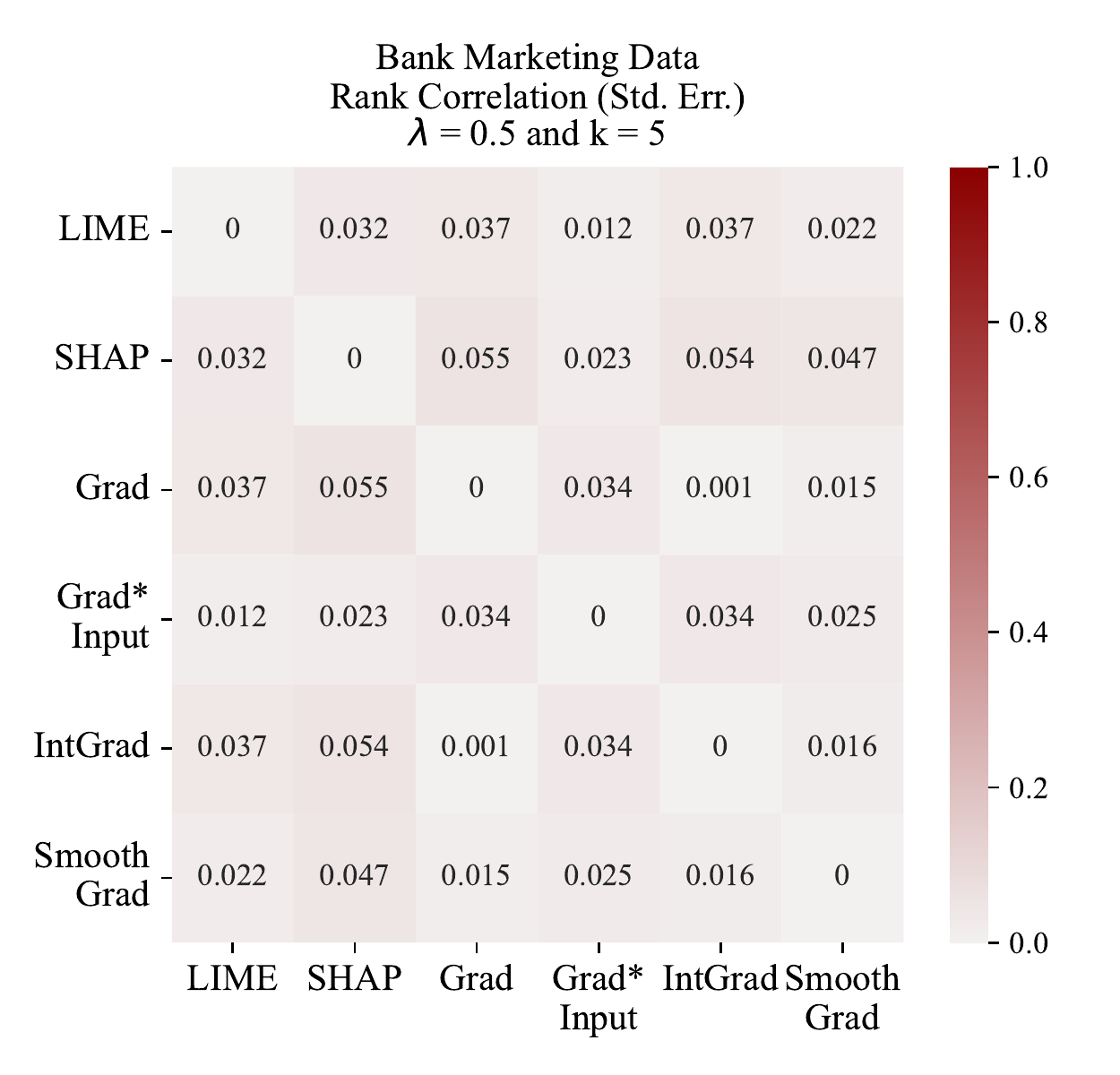}
}}
\fbox{
\parbox[c]{0.28\textwidth}{
\includegraphics[width=0.13\textwidth]{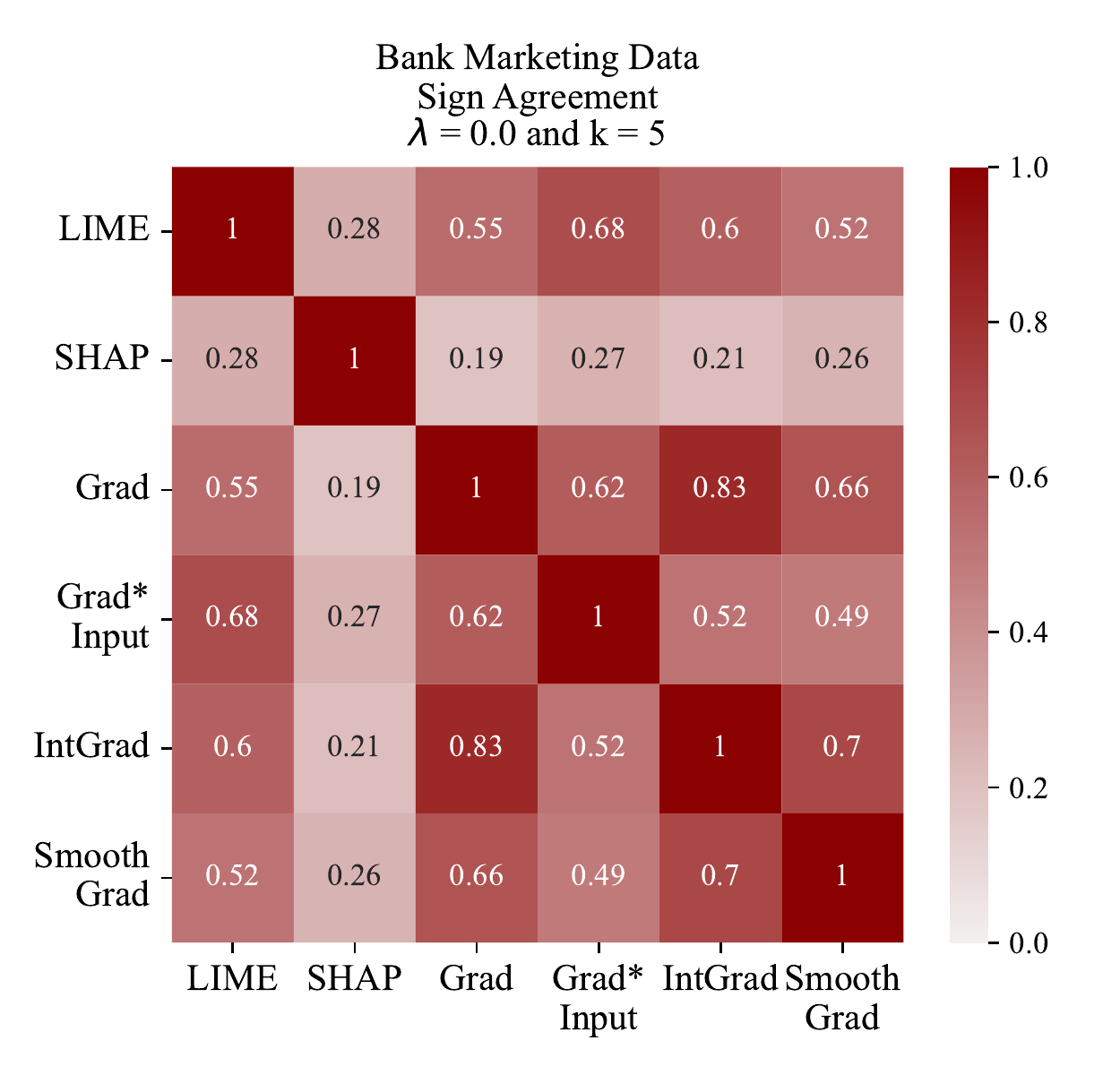}
\includegraphics[width=0.13\textwidth]{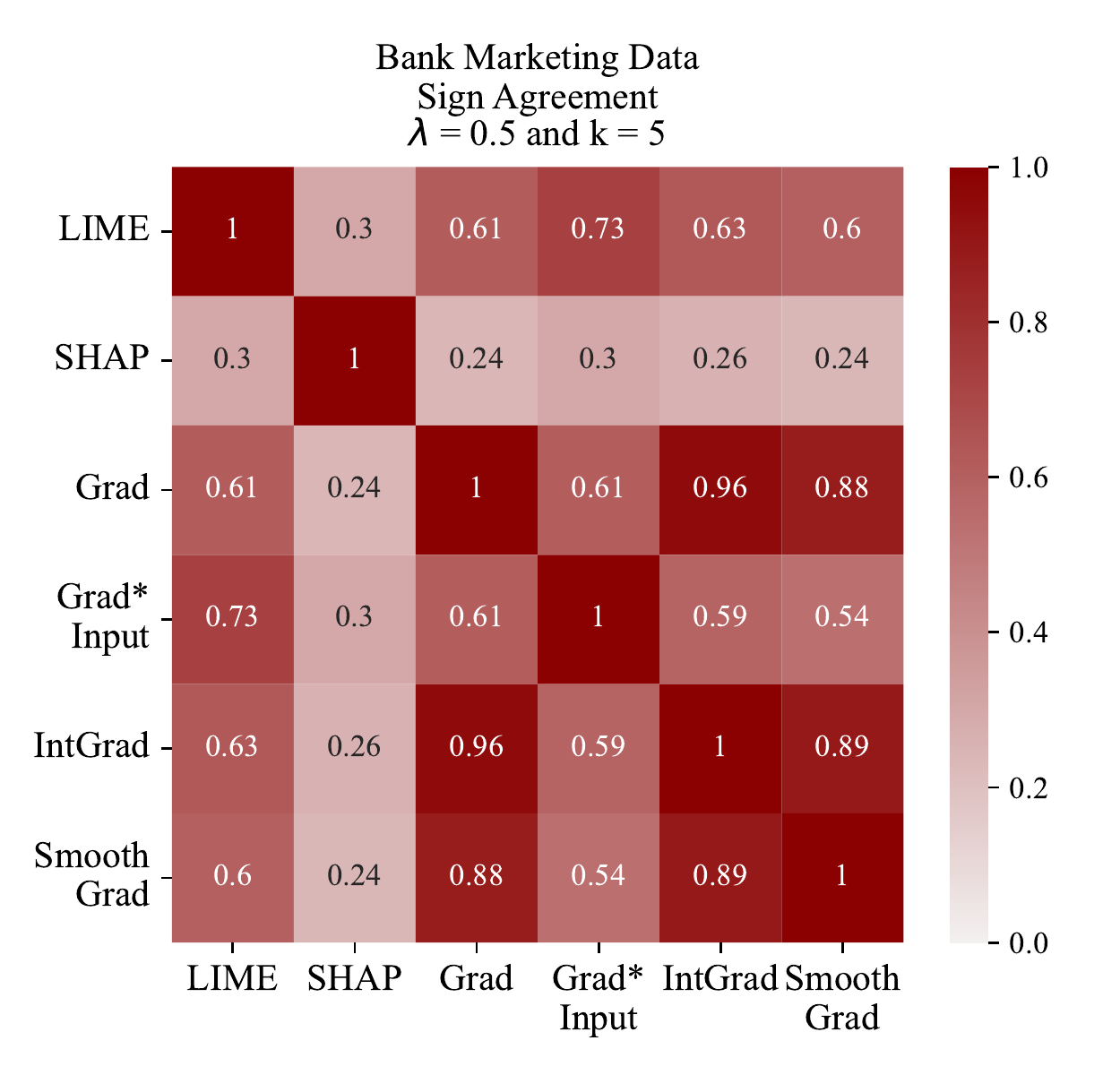}\\
\includegraphics[width=0.13\textwidth]{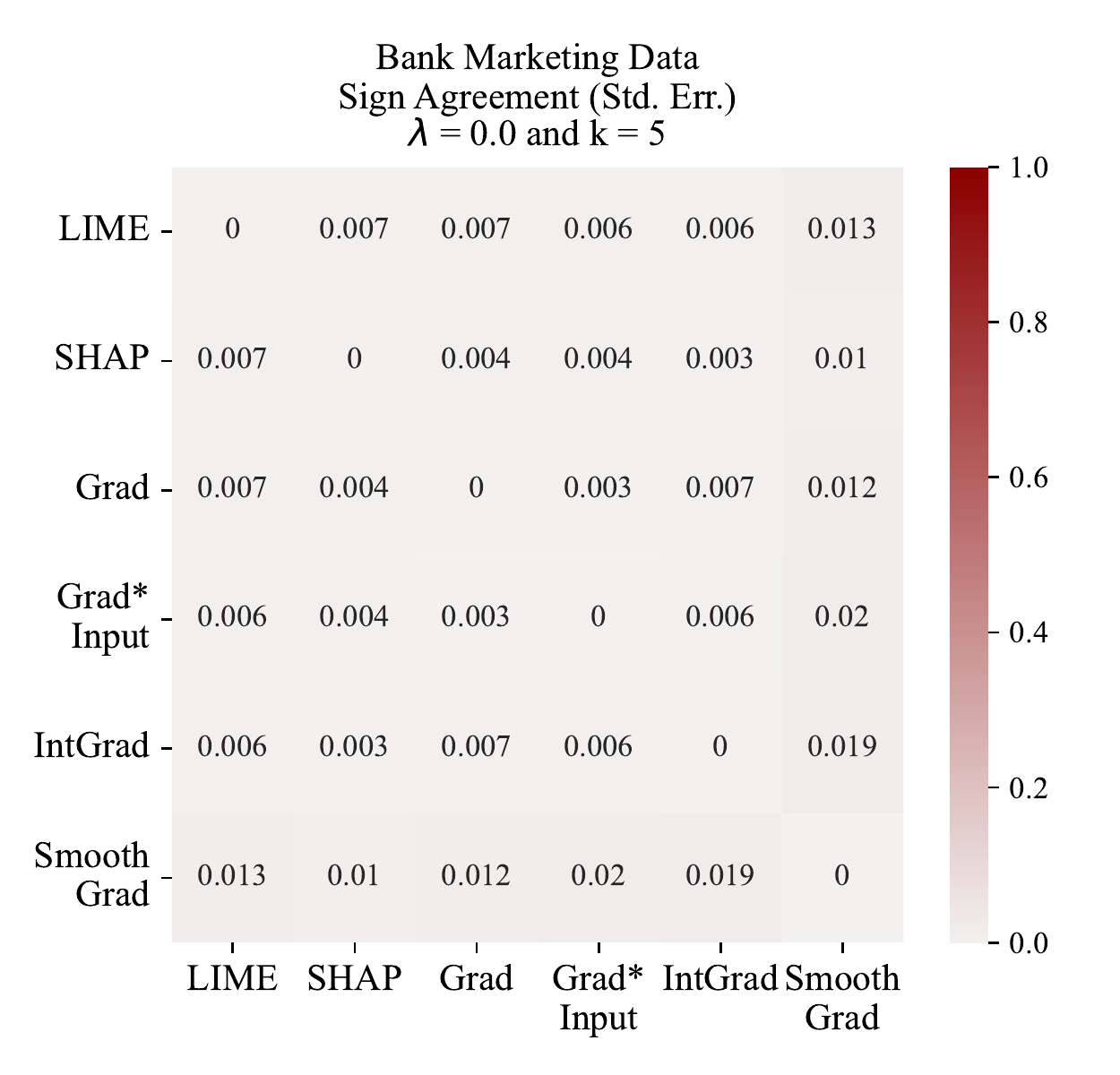}
\includegraphics[width=0.13\textwidth]{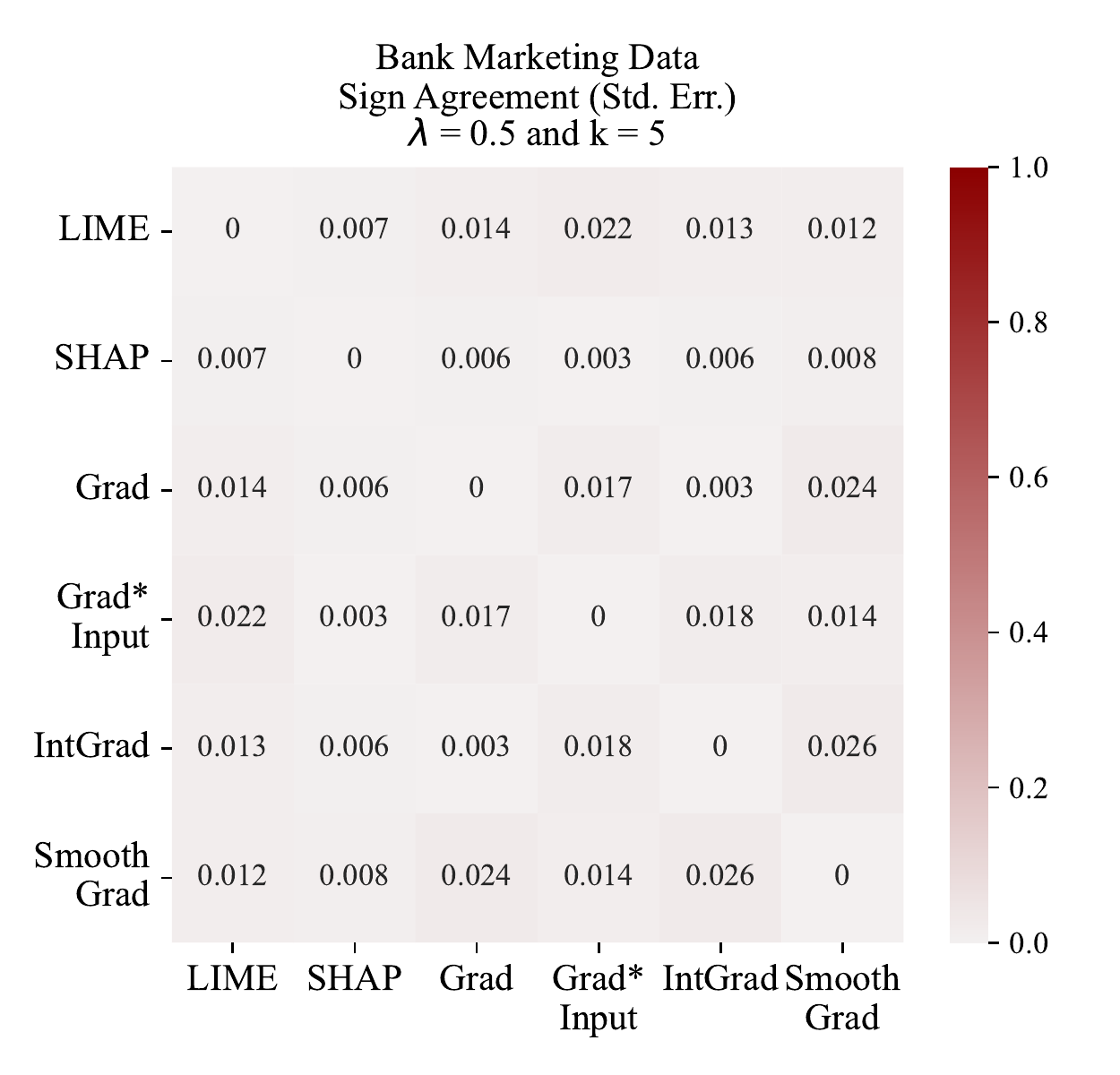}
}}
\fbox{
\parbox[c]{0.28\textwidth}{
\includegraphics[width=0.13\textwidth]{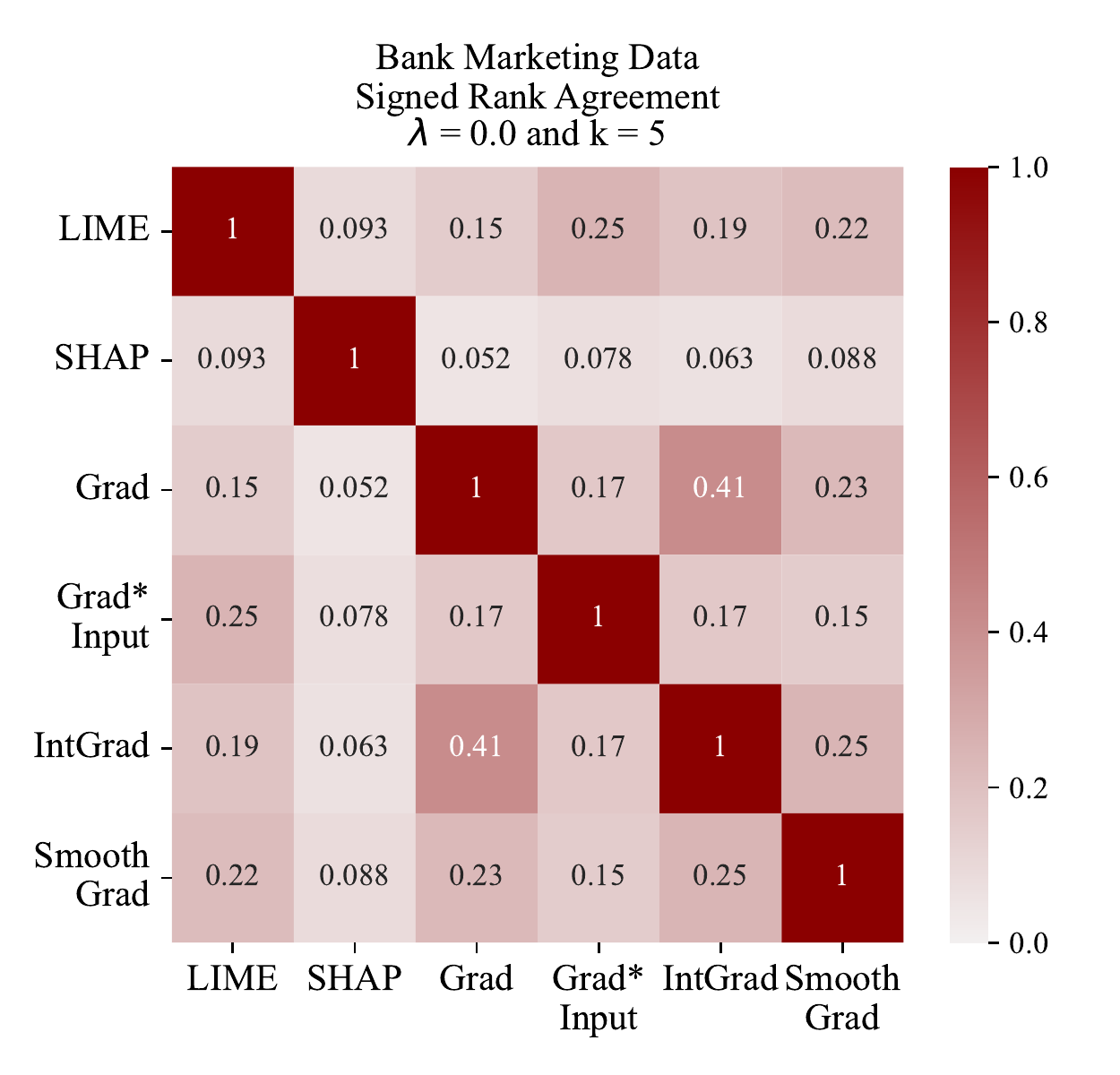}
\includegraphics[width=0.13\textwidth]{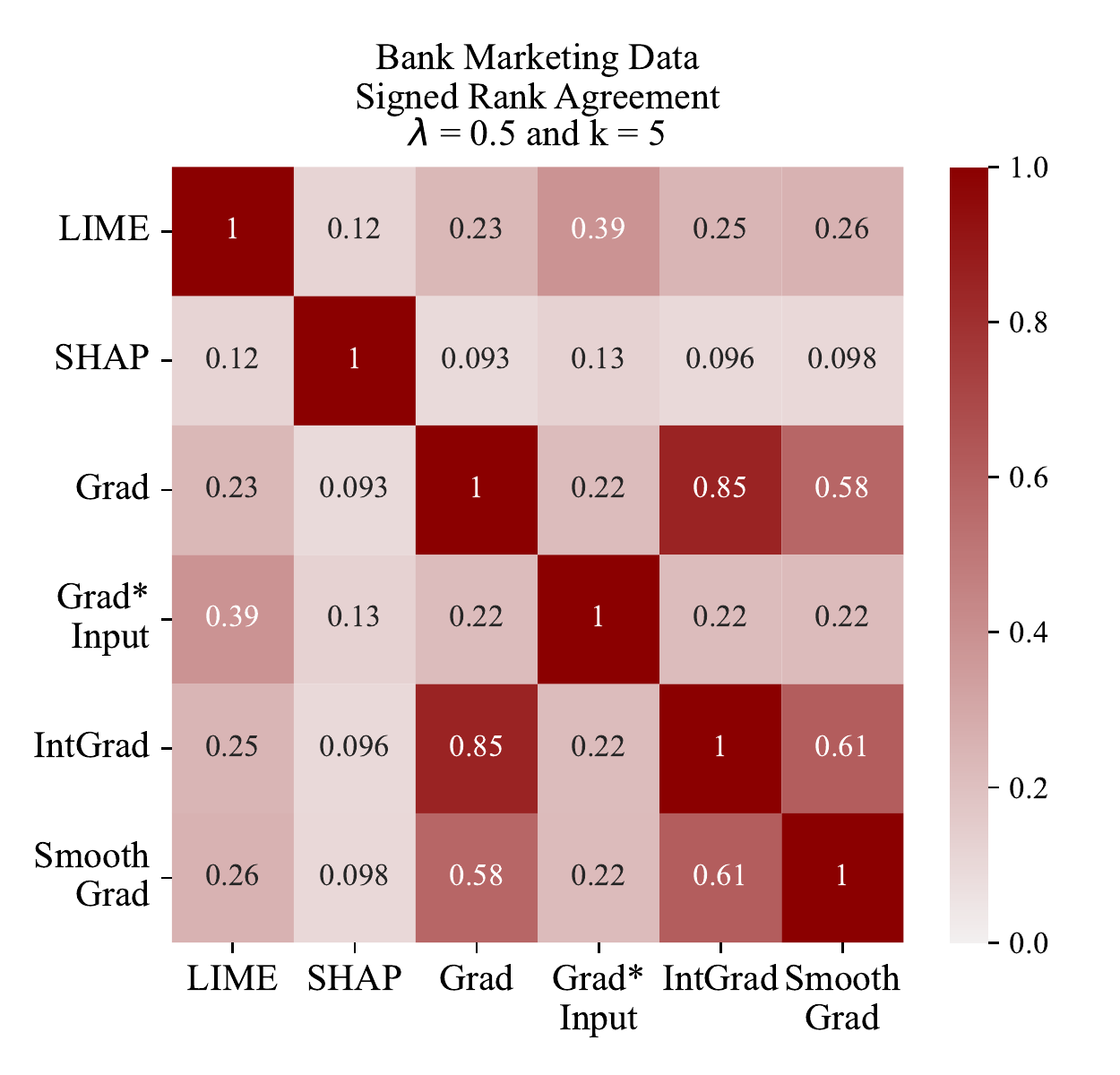}\\
\includegraphics[width=0.13\textwidth]{latest_images/more_red_grids/SEM-Bankmarketing-sign_agreement-lambda=0.0-mu=0.0-test.pdf}
\includegraphics[width=0.13\textwidth]{latest_images/more_red_grids/SEM-Bankmarketing-sign_agreement-lambda=0.5-mu=0.75-test.pdf}
}}
\caption{Disagreement matrices for all metrics considered in this paper on Bank Marketing data.}
\label{fig:more-redgrids-bank}
\end{figure*}
\begin{figure*}[ht!]
\centering
\fbox{
\parbox[c]{0.28\textwidth}{
\includegraphics[width=0.13\textwidth]{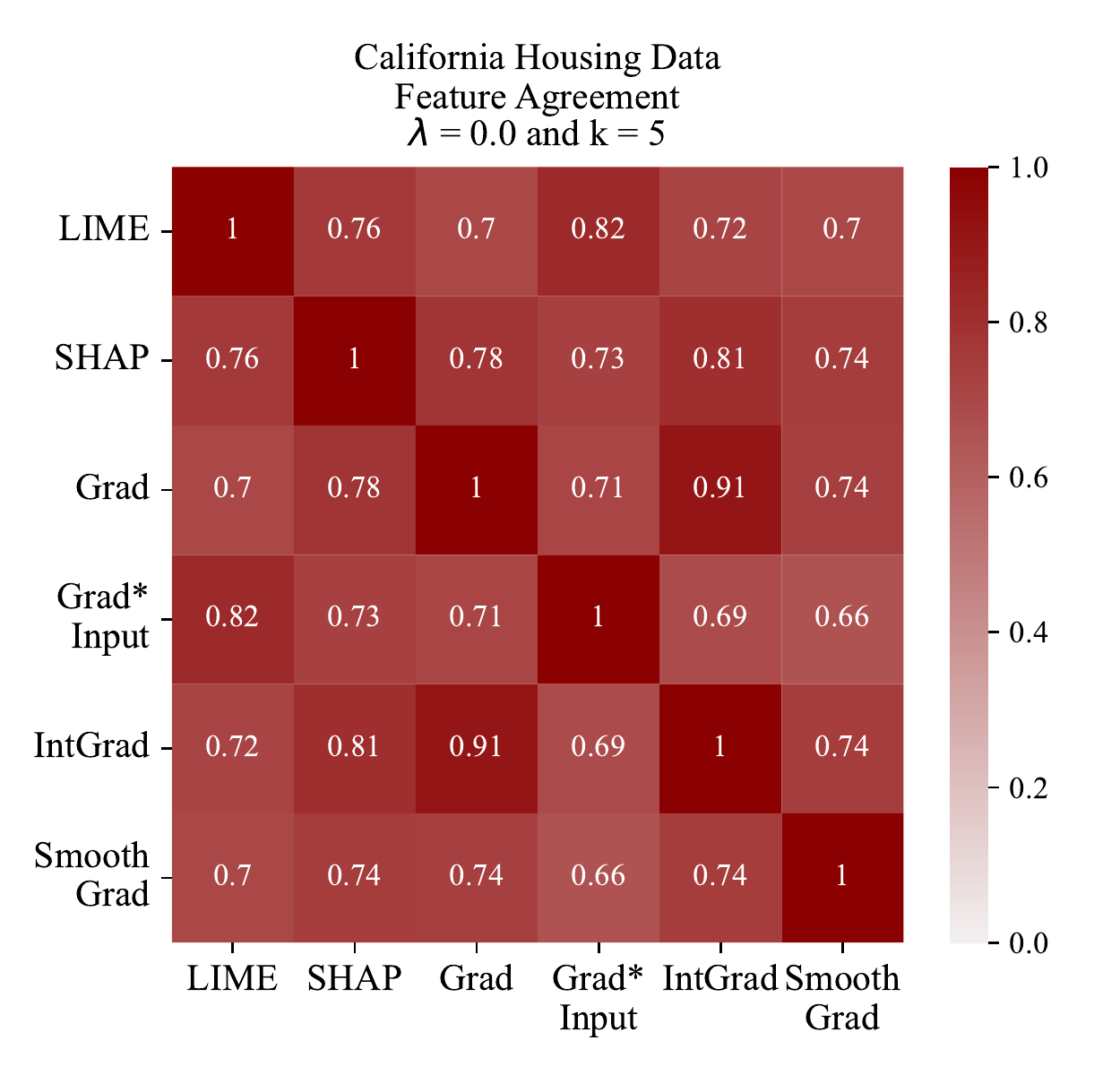}
\includegraphics[width=0.13\textwidth]{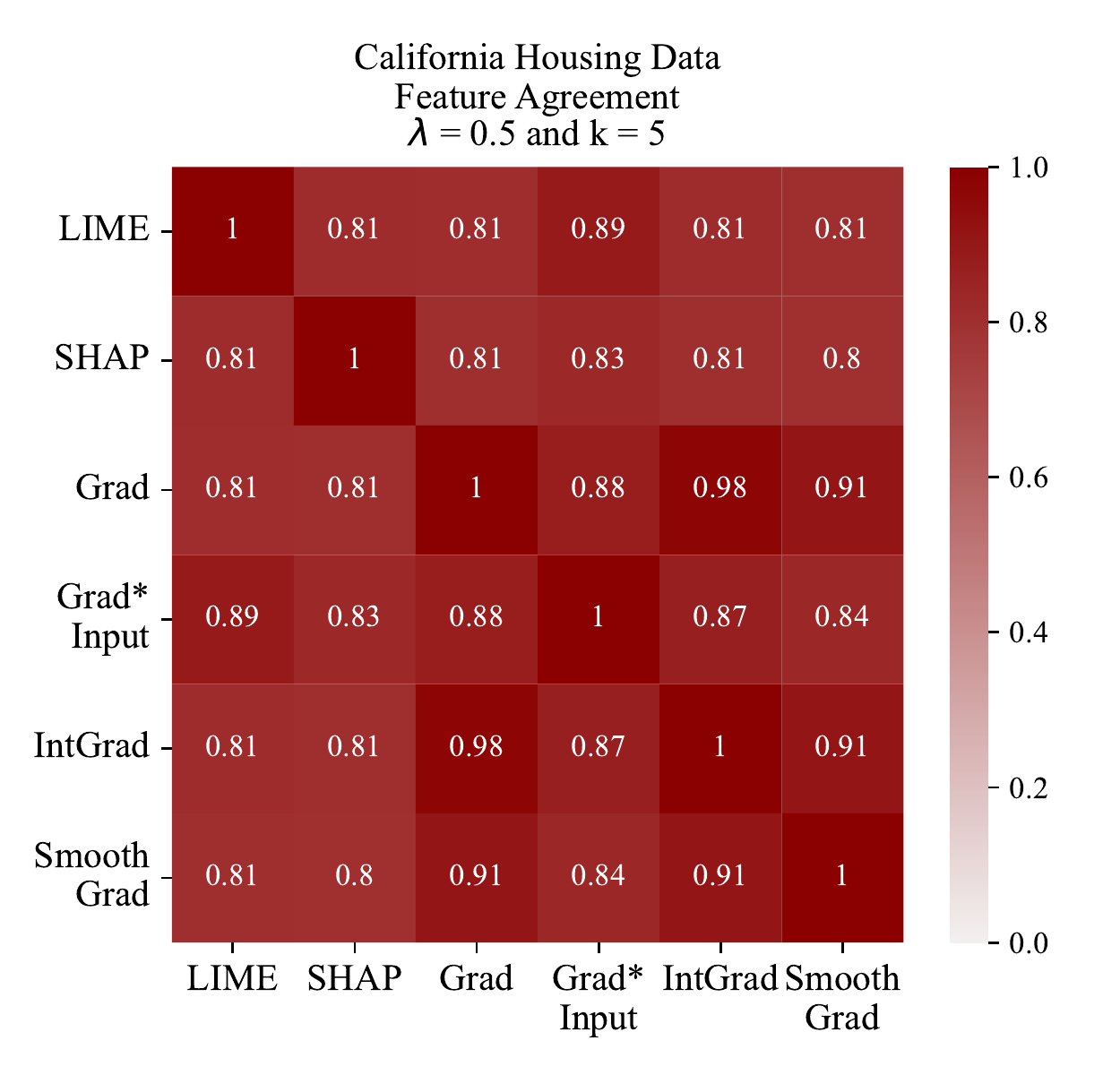}\\
\includegraphics[width=0.13\textwidth]{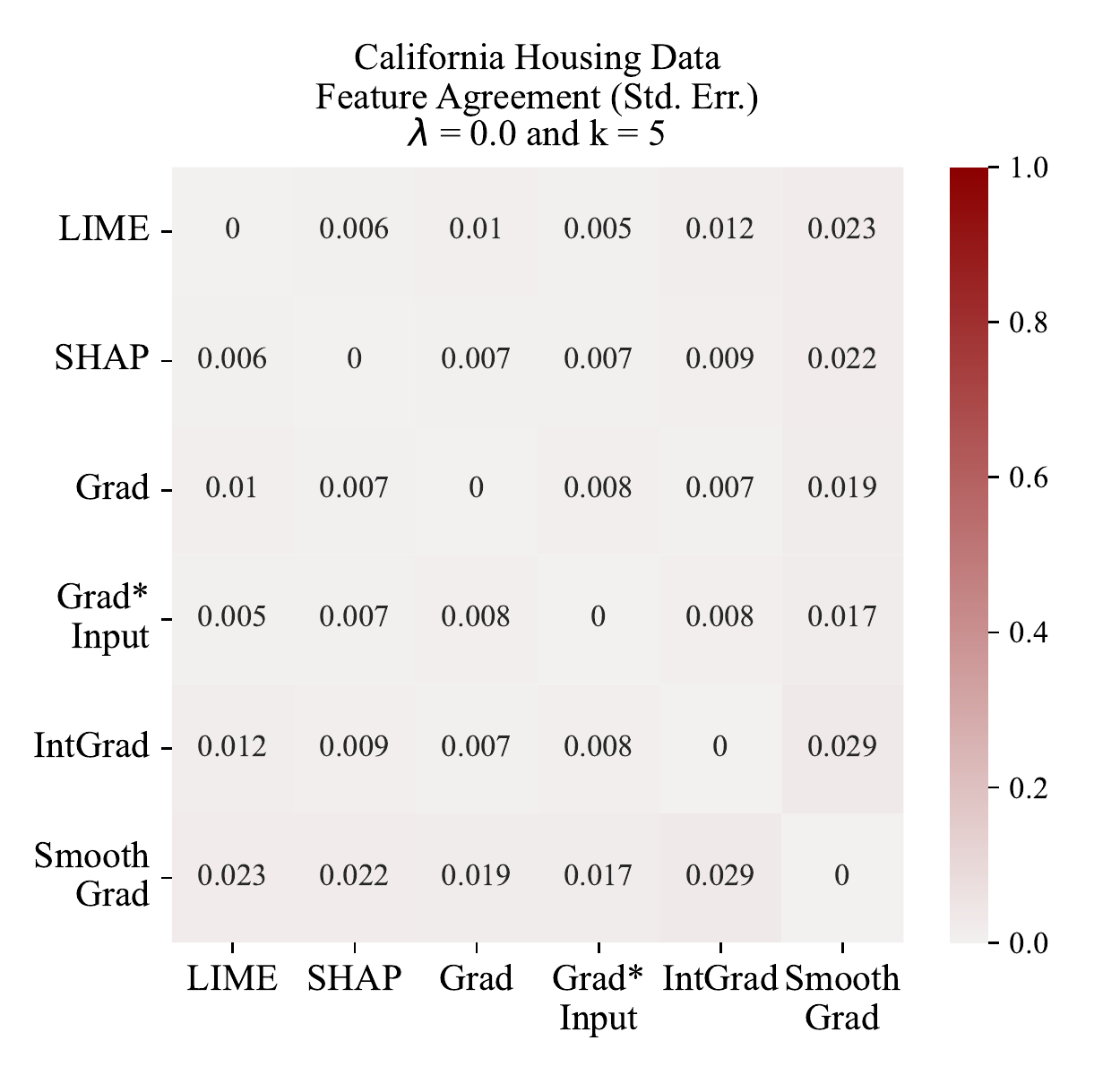}
\includegraphics[width=0.13\textwidth]{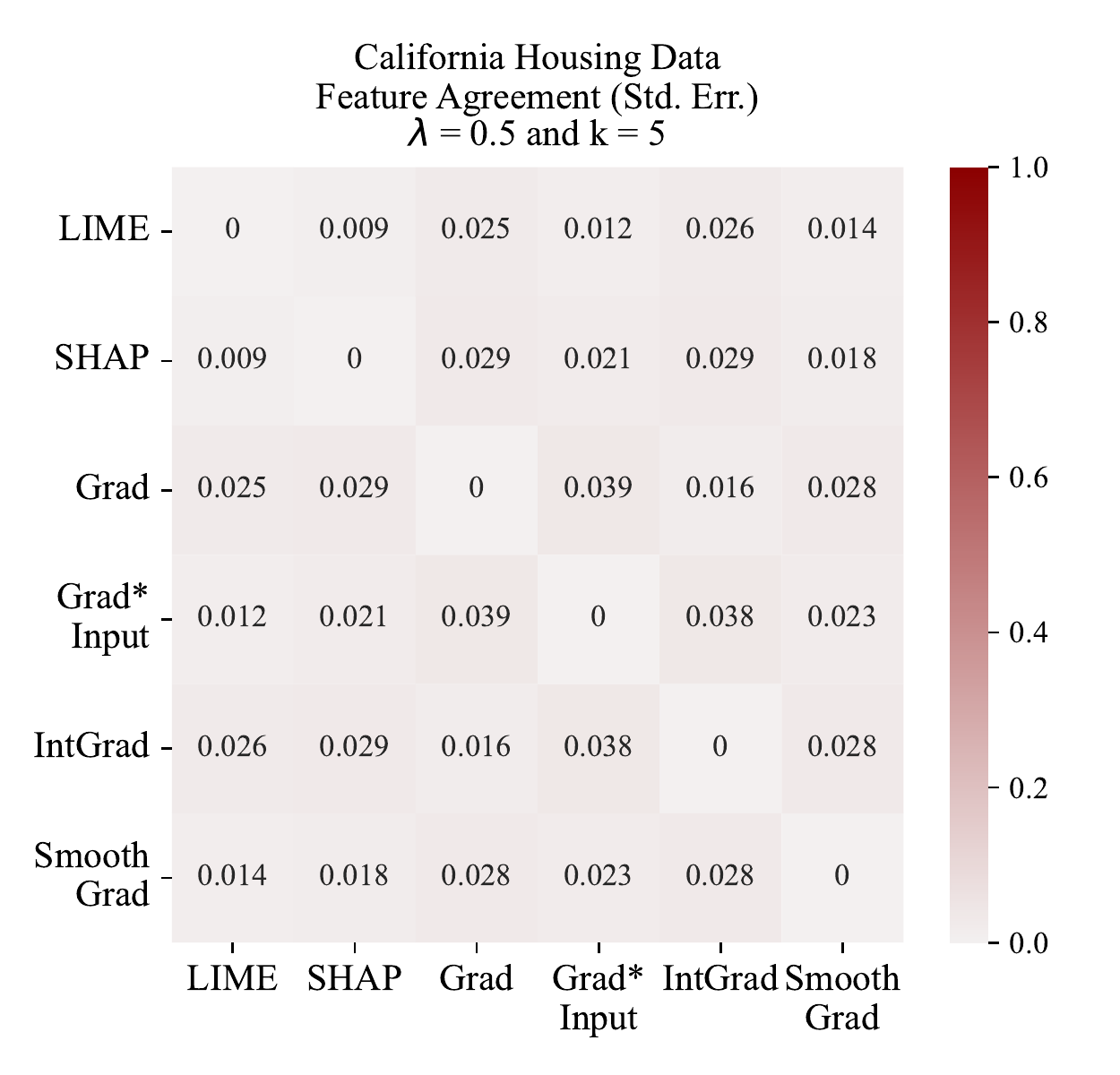}
}}
\fbox{
\parbox[c]{0.28\textwidth}{
\includegraphics[width=0.13\textwidth]{latest_images/more_red_grids/Californiahousing-pairwise_rank-lambda=0.0-mu=0.0-test.pdf}
\includegraphics[width=0.13\textwidth]{latest_images/more_red_grids/Californiahousing-pairwise_rank-lambda=0.5-mu=0.75-test.pdf}\\
\includegraphics[width=0.13\textwidth]{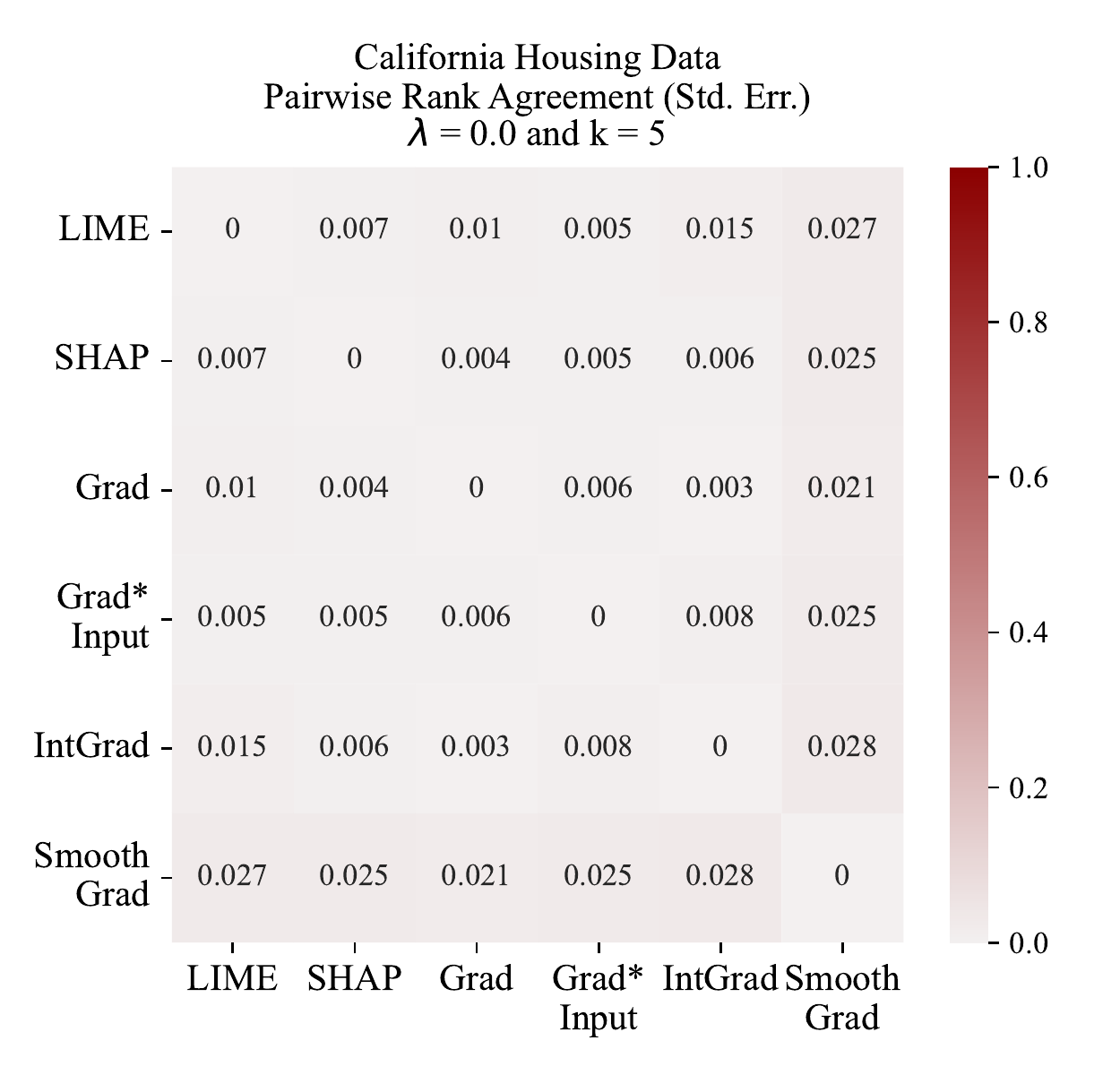}
\includegraphics[width=0.13\textwidth]{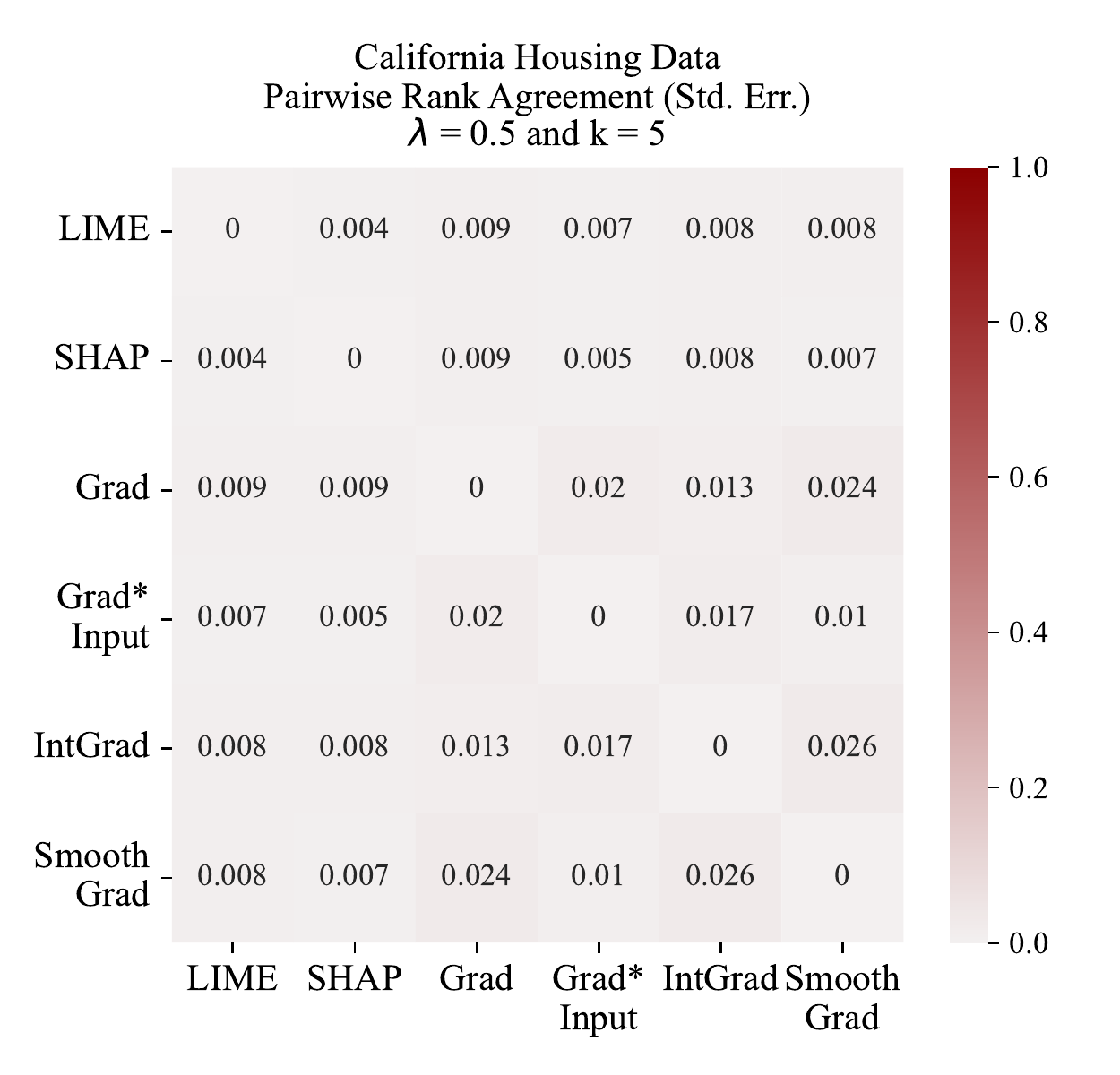}
}}
\fbox{
\parbox[c]{0.28\textwidth}{
\includegraphics[width=0.13\textwidth]{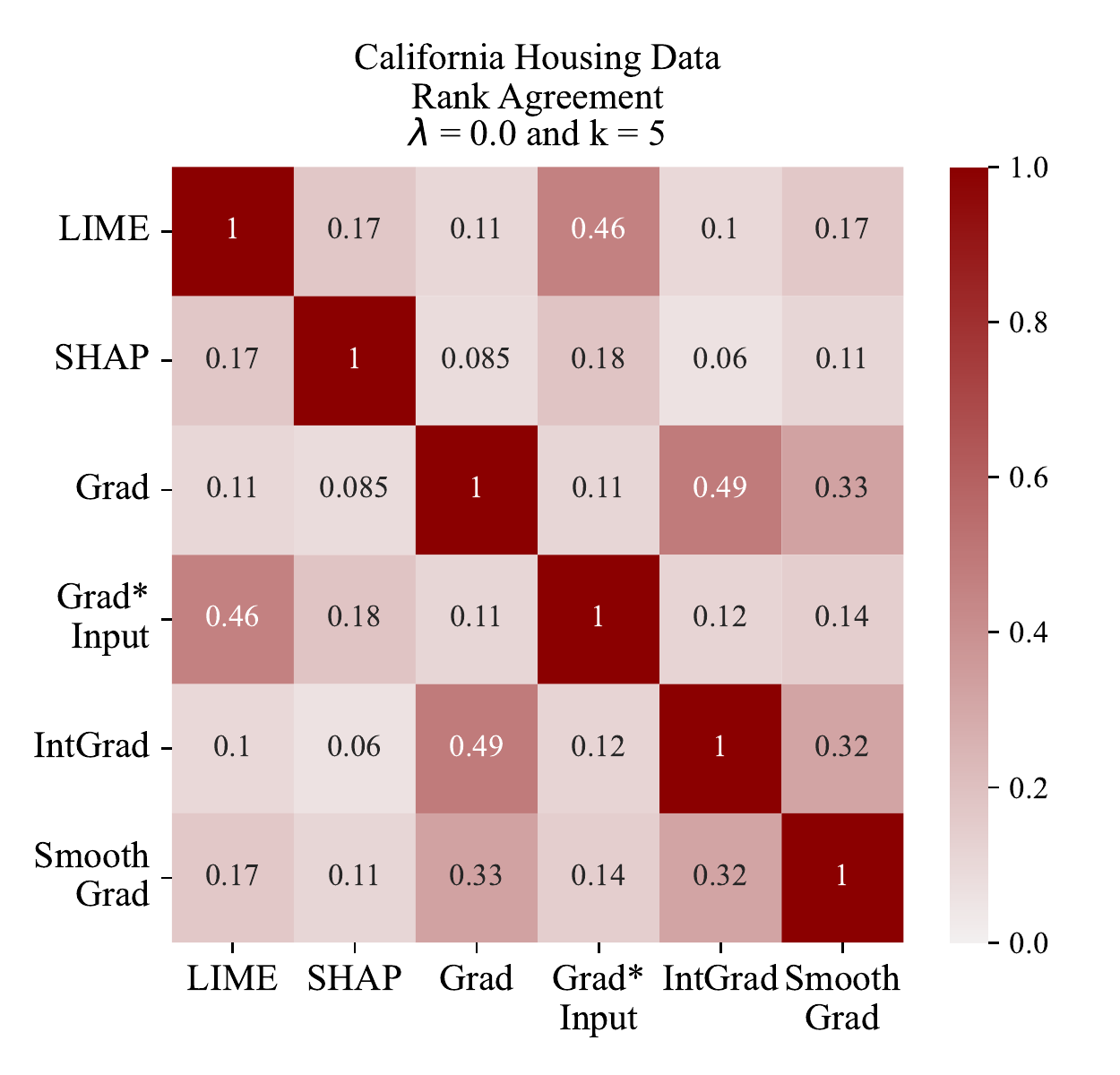}
\includegraphics[width=0.13\textwidth]{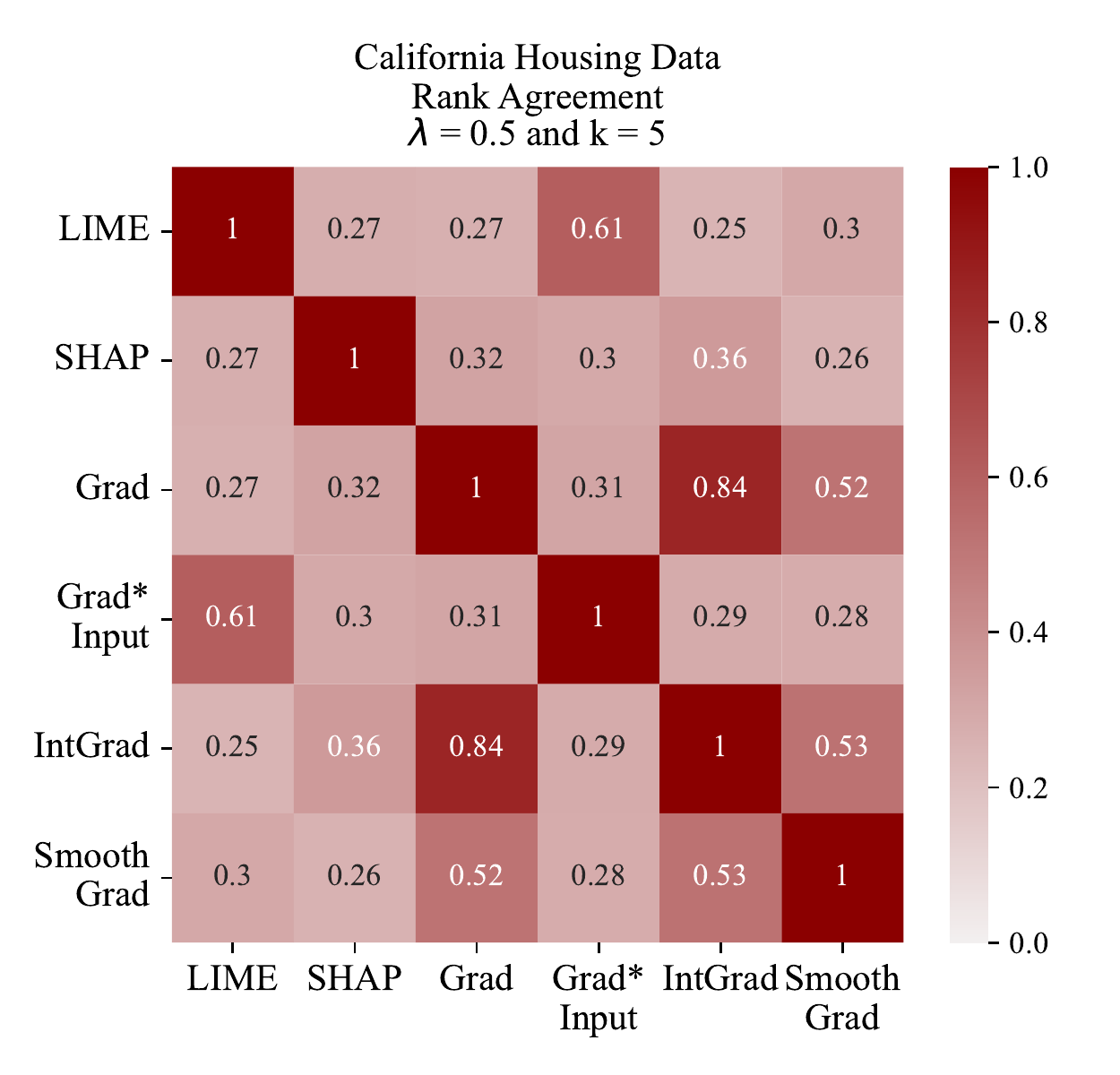}\\
\includegraphics[width=0.13\textwidth]{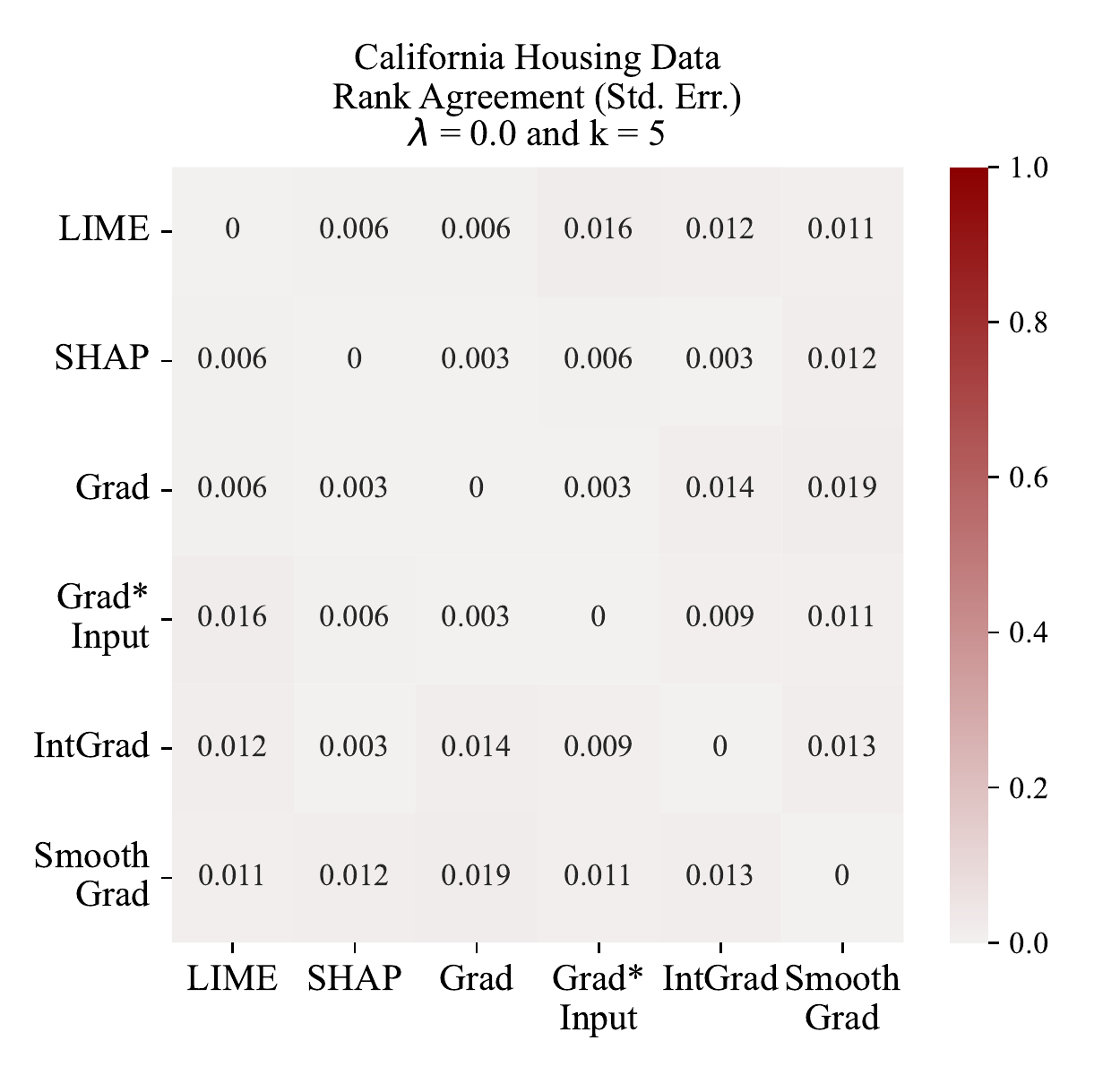}
\includegraphics[width=0.13\textwidth]{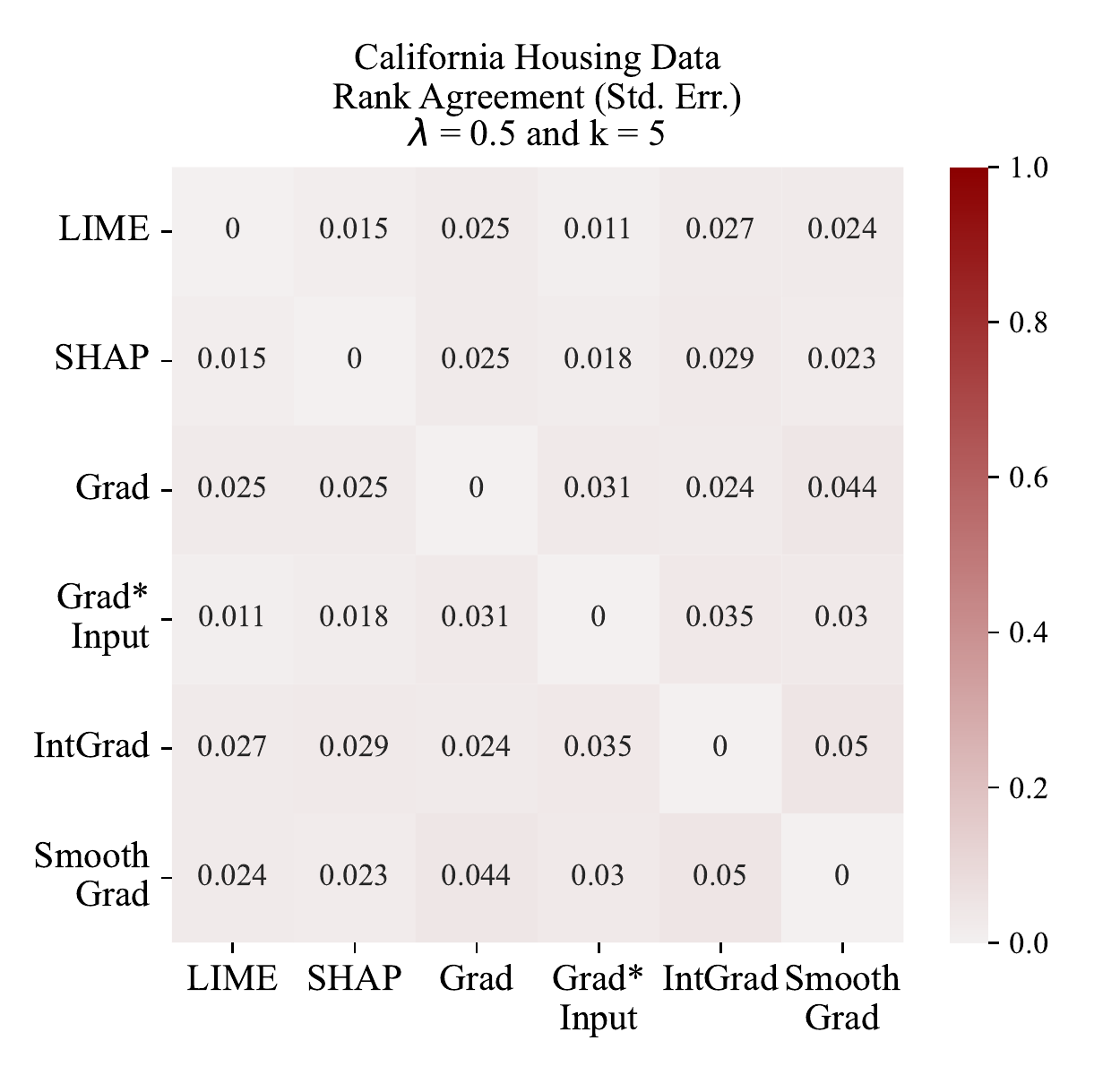}
}}
\\
\fbox{
\parbox[c]{0.28\textwidth}{
\includegraphics[width=0.13\textwidth]{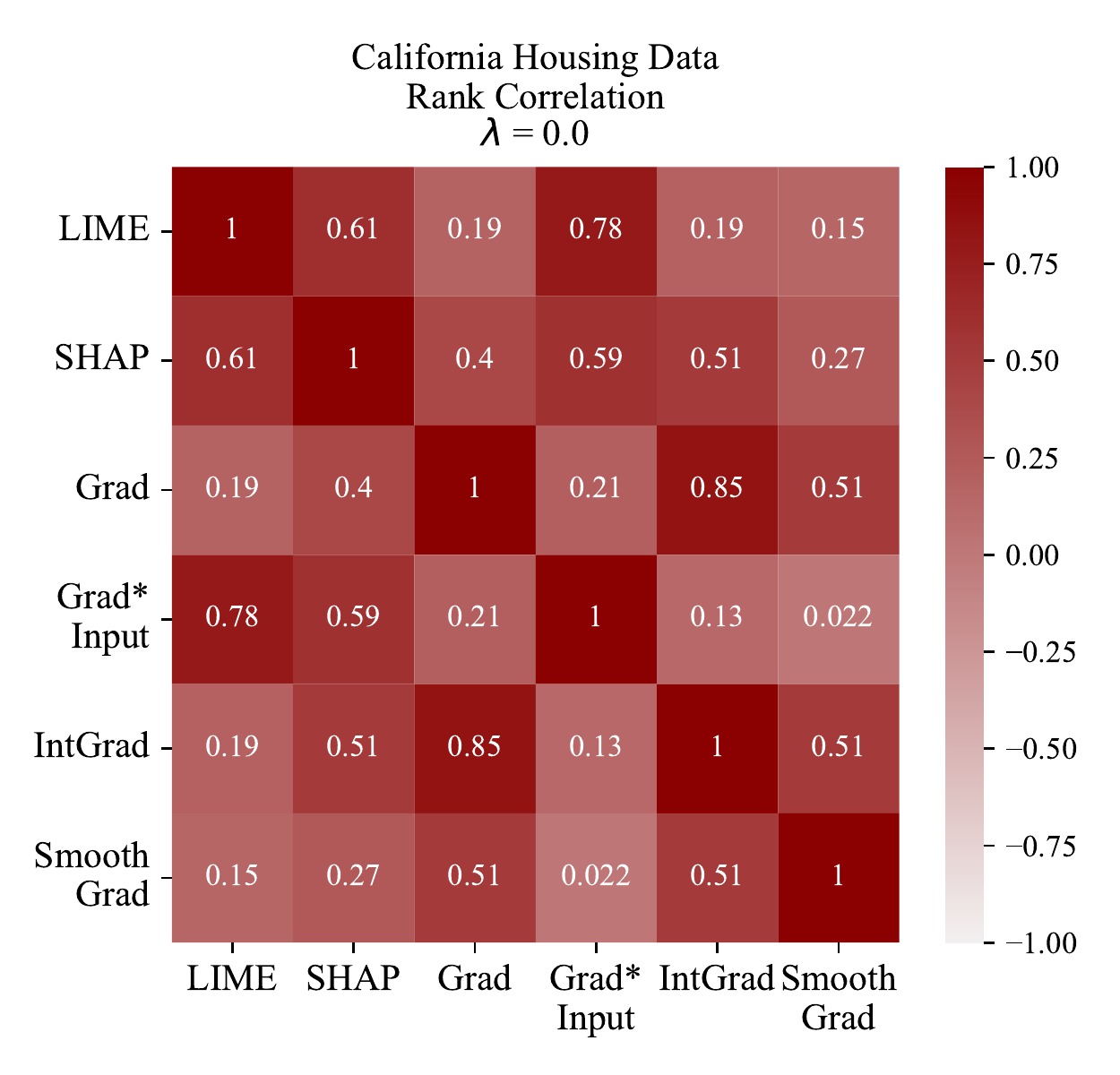}
\includegraphics[width=0.13\textwidth]{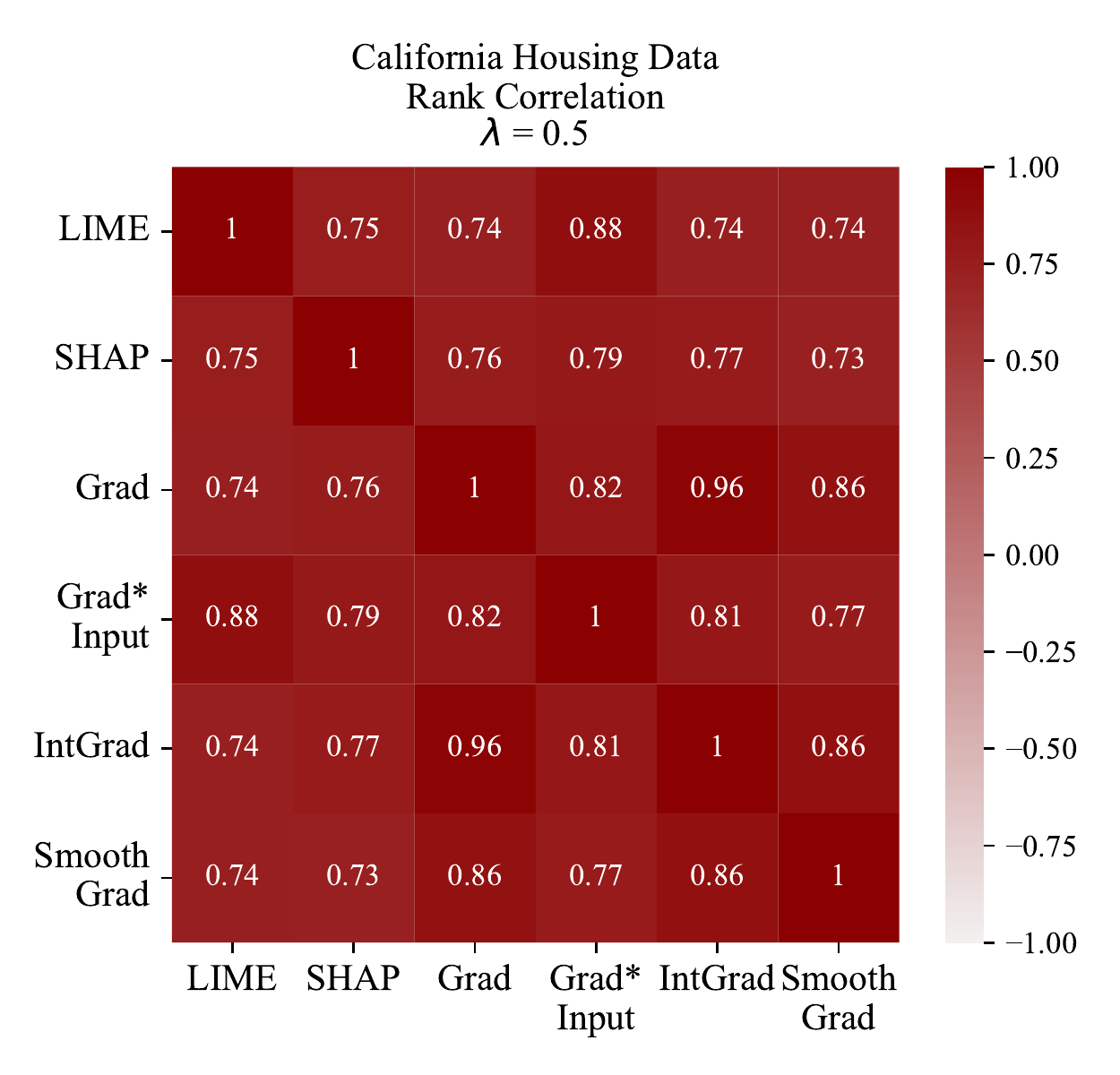}\\
\includegraphics[width=0.13\textwidth]{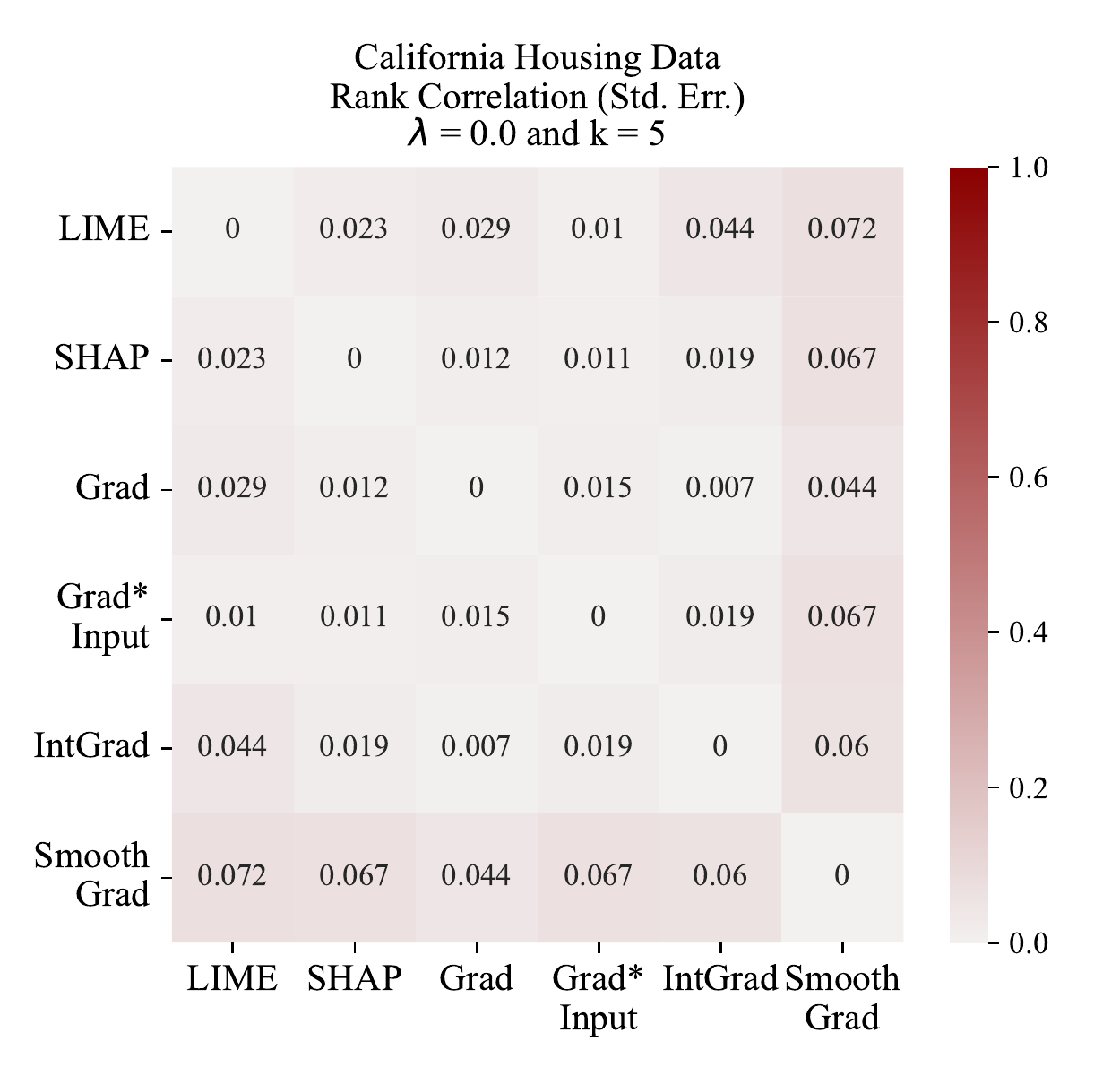}
\includegraphics[width=0.13\textwidth]{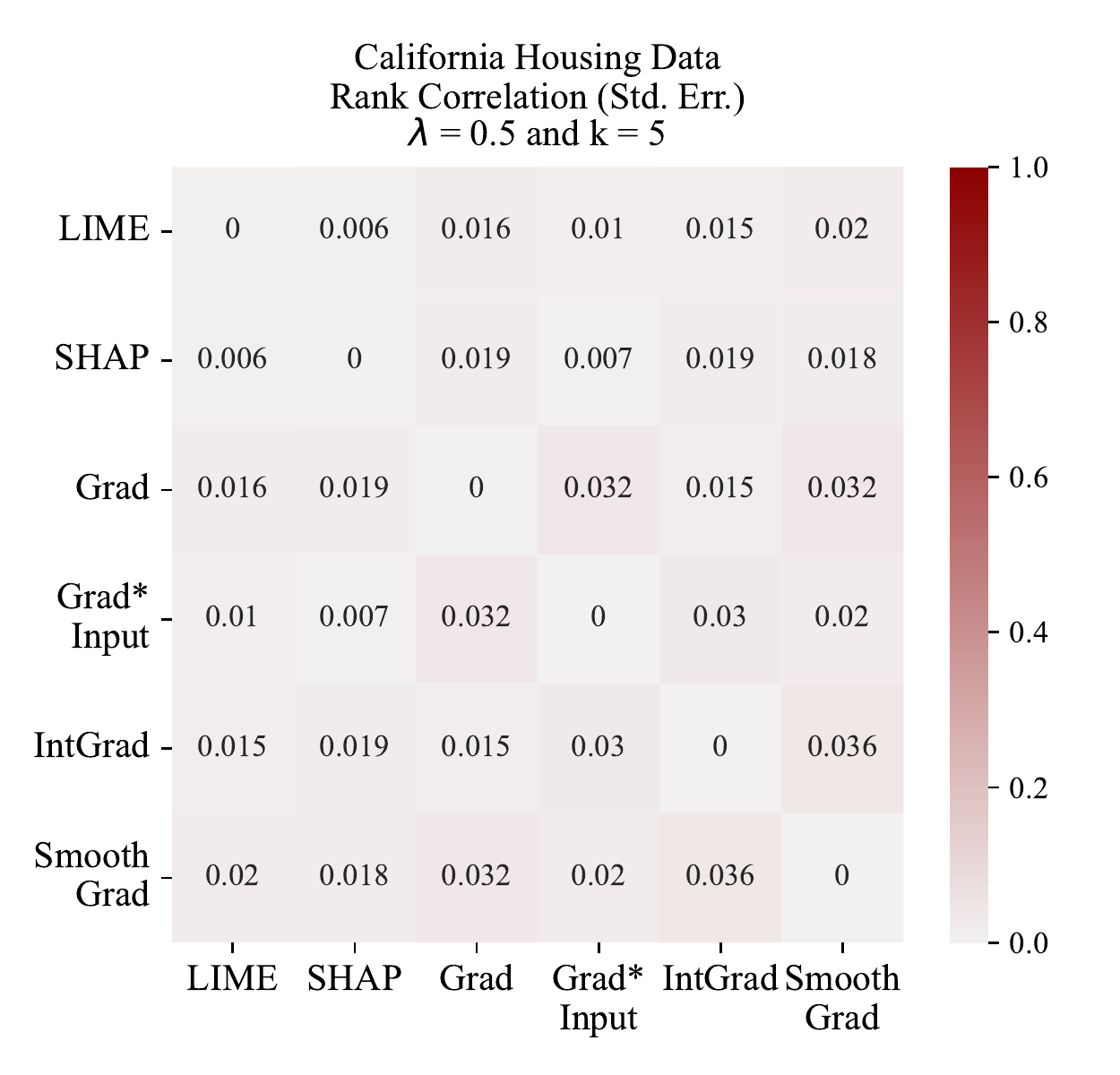}
}}
\fbox{
\parbox[c]{0.28\textwidth}{
\includegraphics[width=0.13\textwidth]{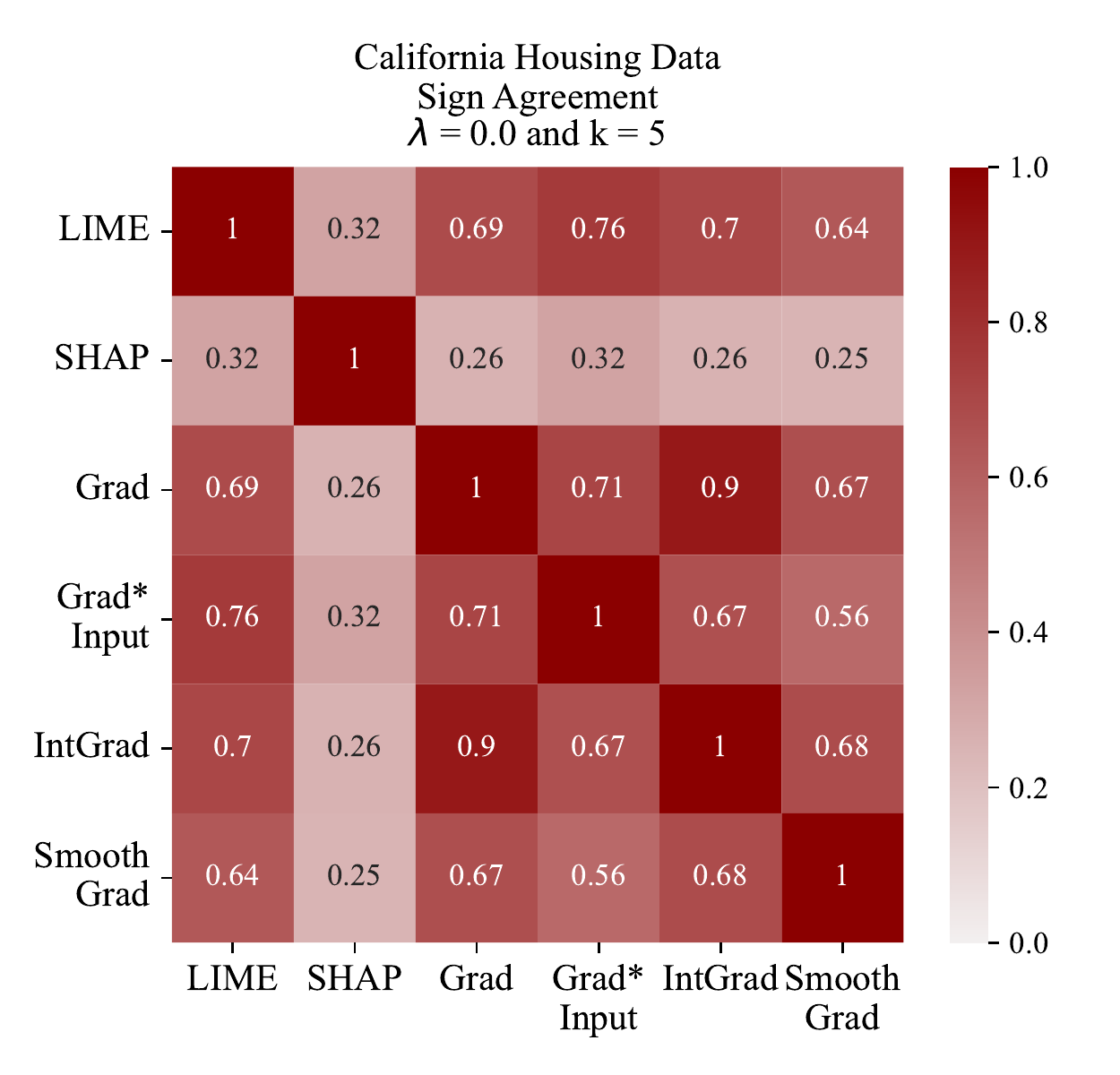}
\includegraphics[width=0.13\textwidth]{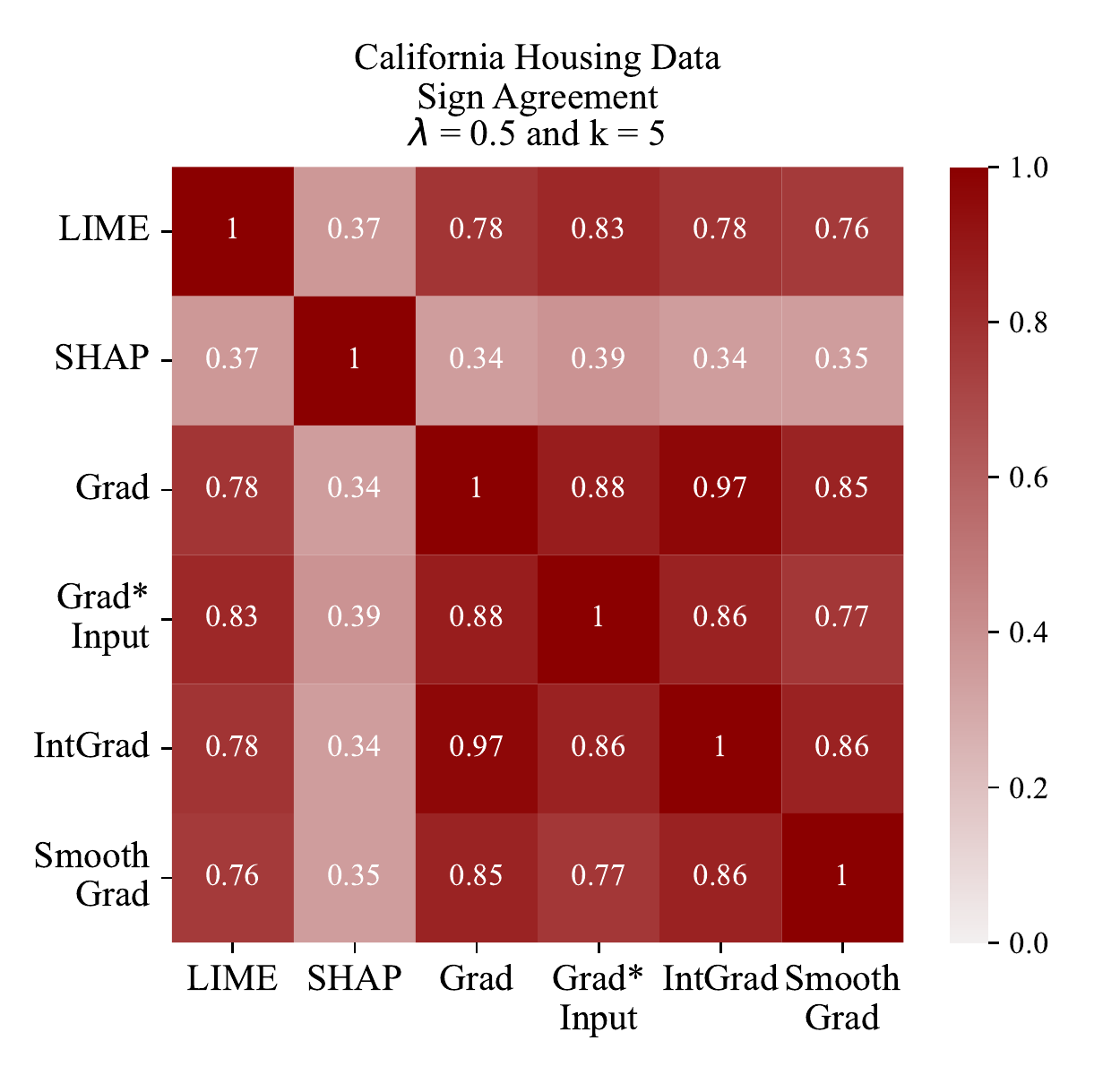}\\
\includegraphics[width=0.13\textwidth]{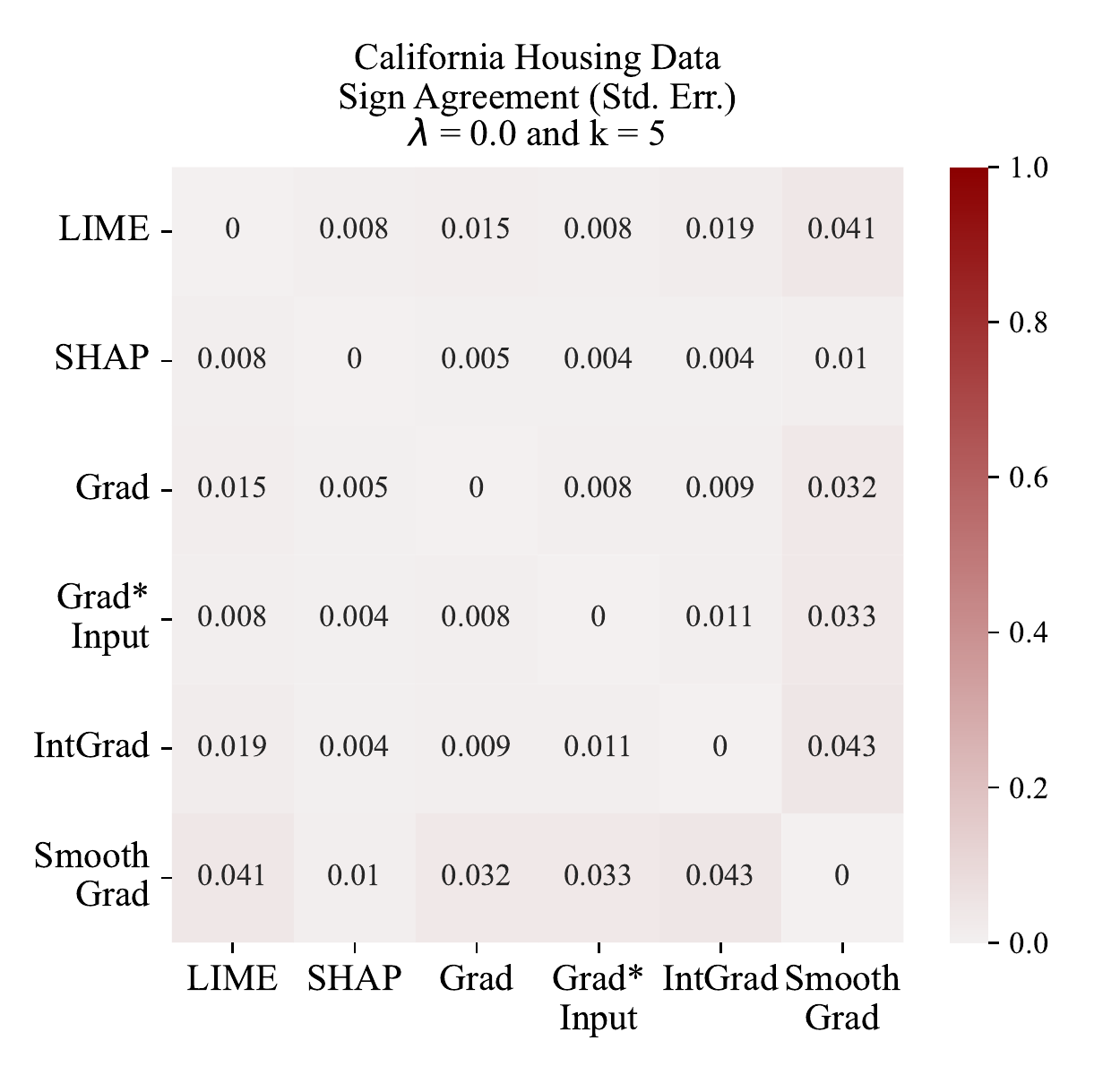}
\includegraphics[width=0.13\textwidth]{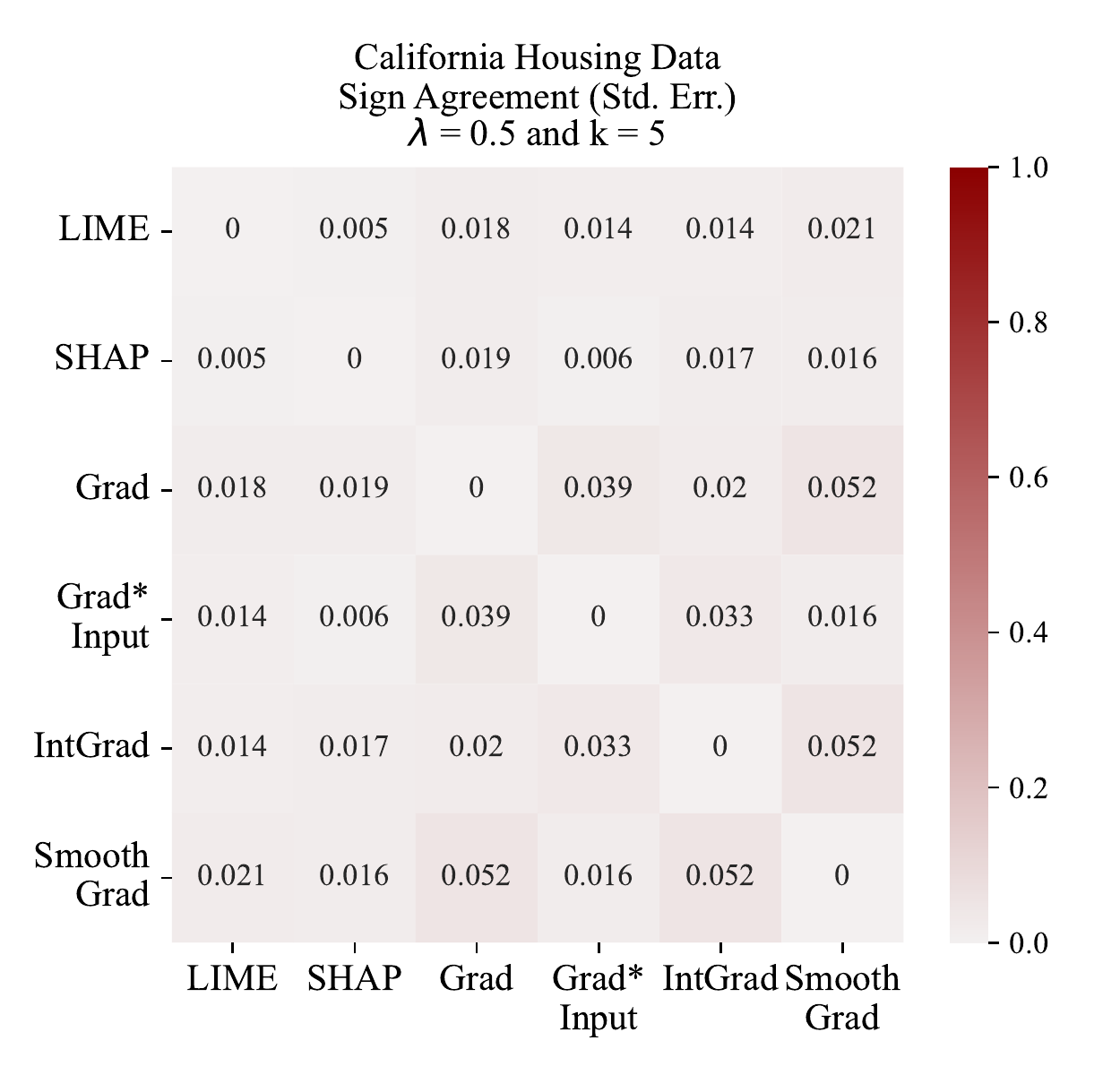}
}}
\fbox{
\parbox[c]{0.28\textwidth}{
\includegraphics[width=0.13\textwidth]{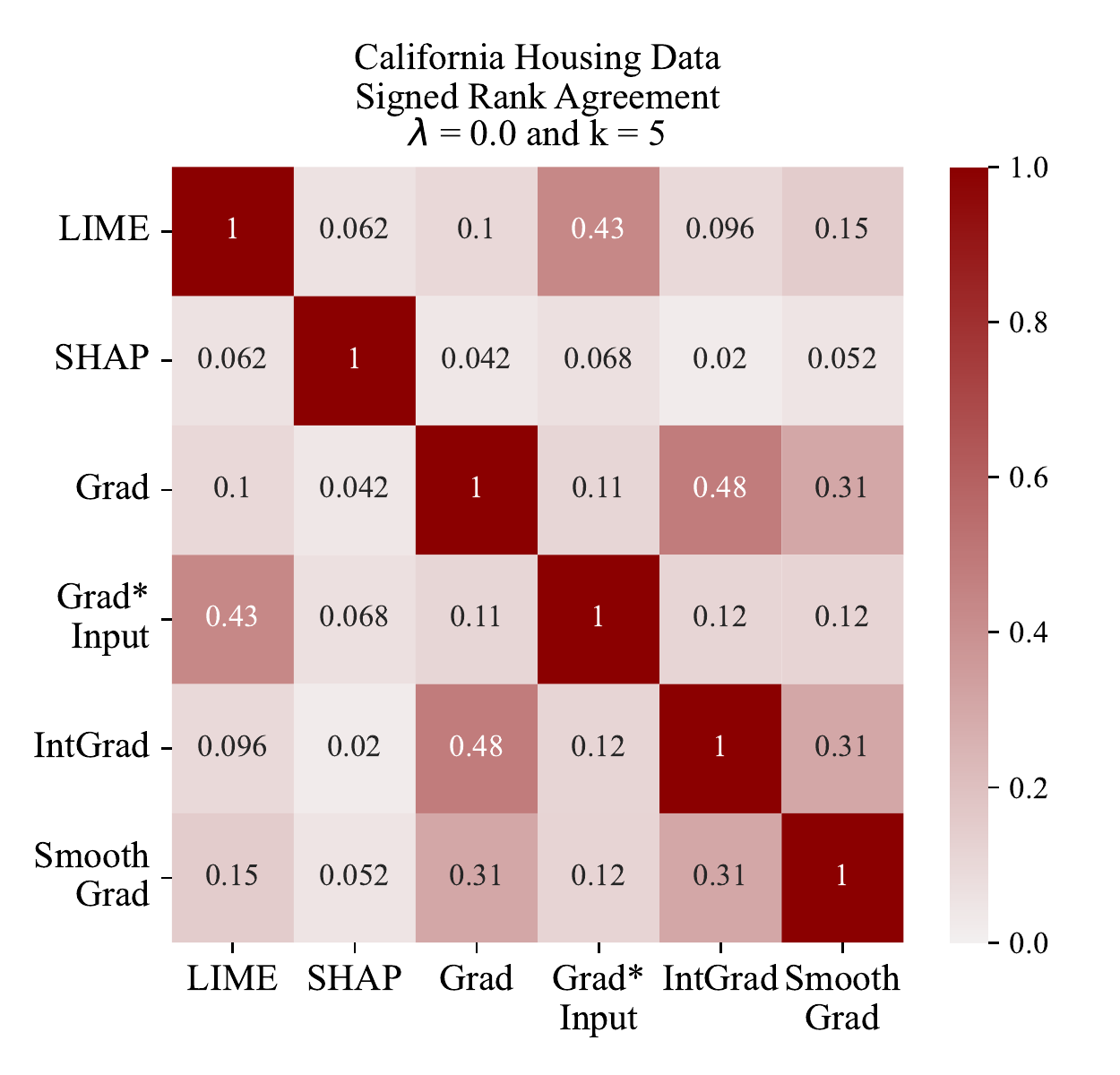}
\includegraphics[width=0.13\textwidth]{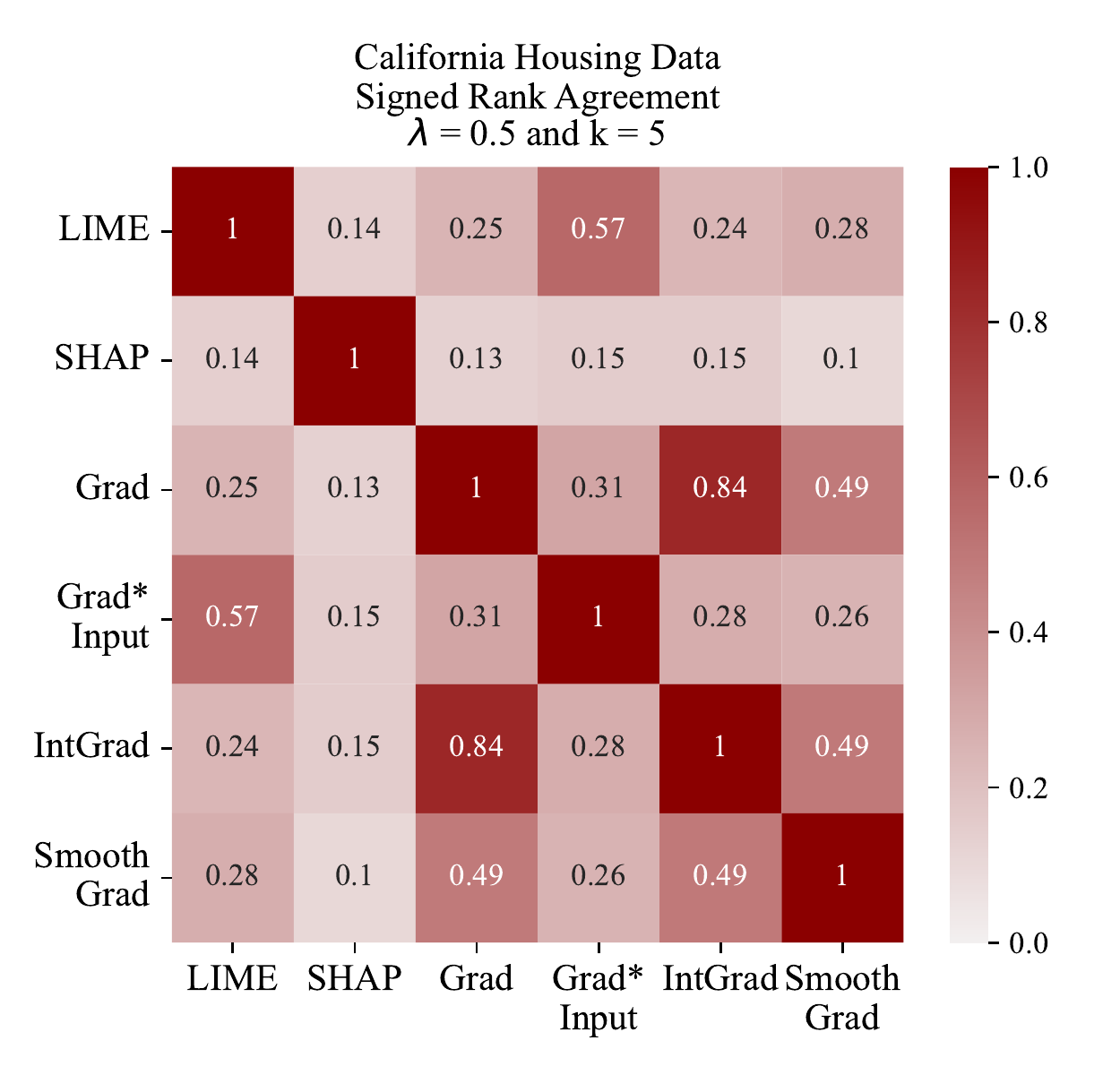}\\
\includegraphics[width=0.13\textwidth]{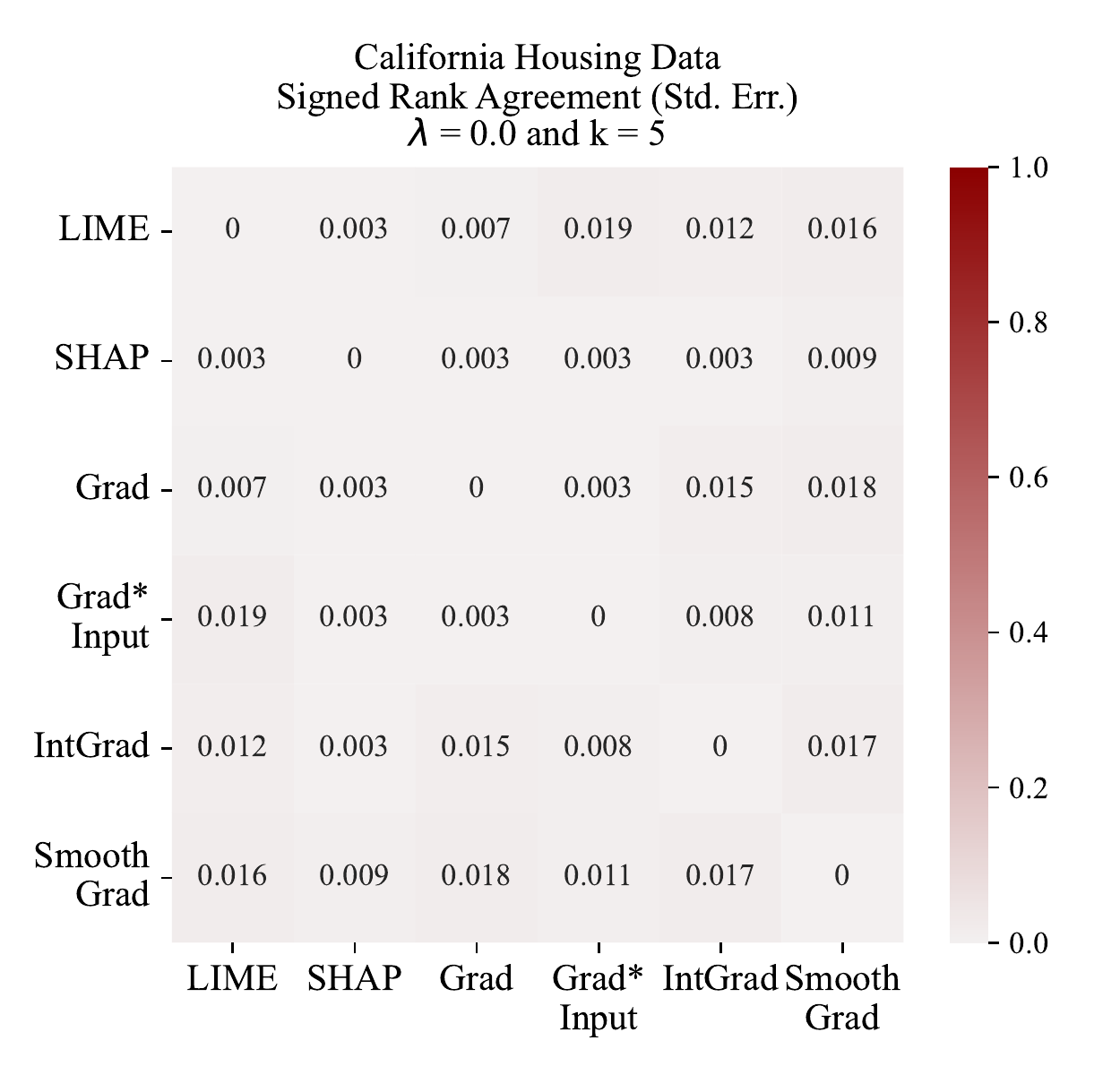}
\includegraphics[width=0.13\textwidth]{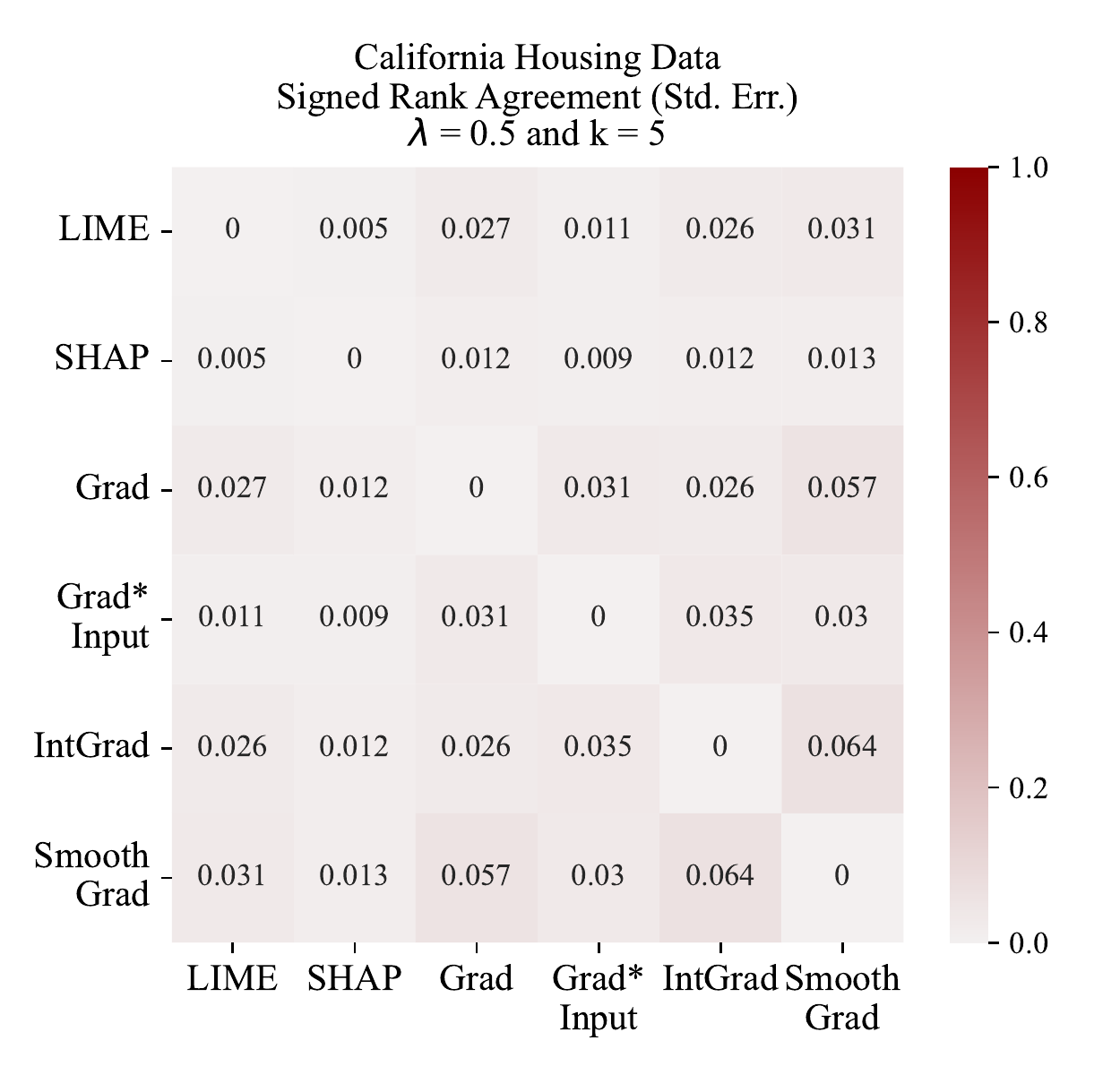}
}}
\caption{Disagreement matrices for all metrics considered in this paper on California Housing data.}
\label{fig:more-redgrids-cali}
\end{figure*}
\begin{figure*}[ht!]
\centering
\fbox{
\parbox[c]{0.28\textwidth}{
\includegraphics[width=0.13\textwidth]{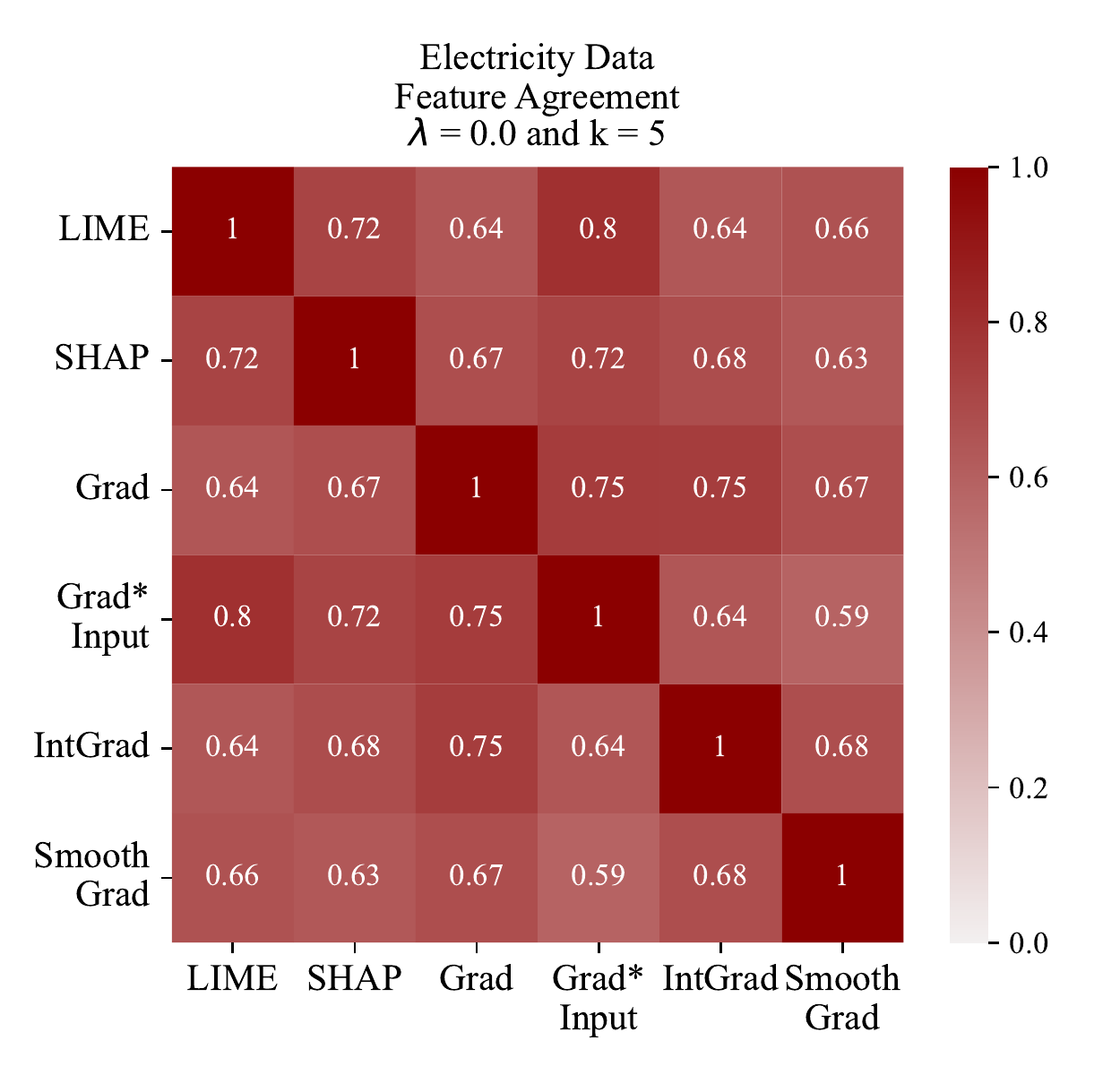}
\includegraphics[width=0.13\textwidth]{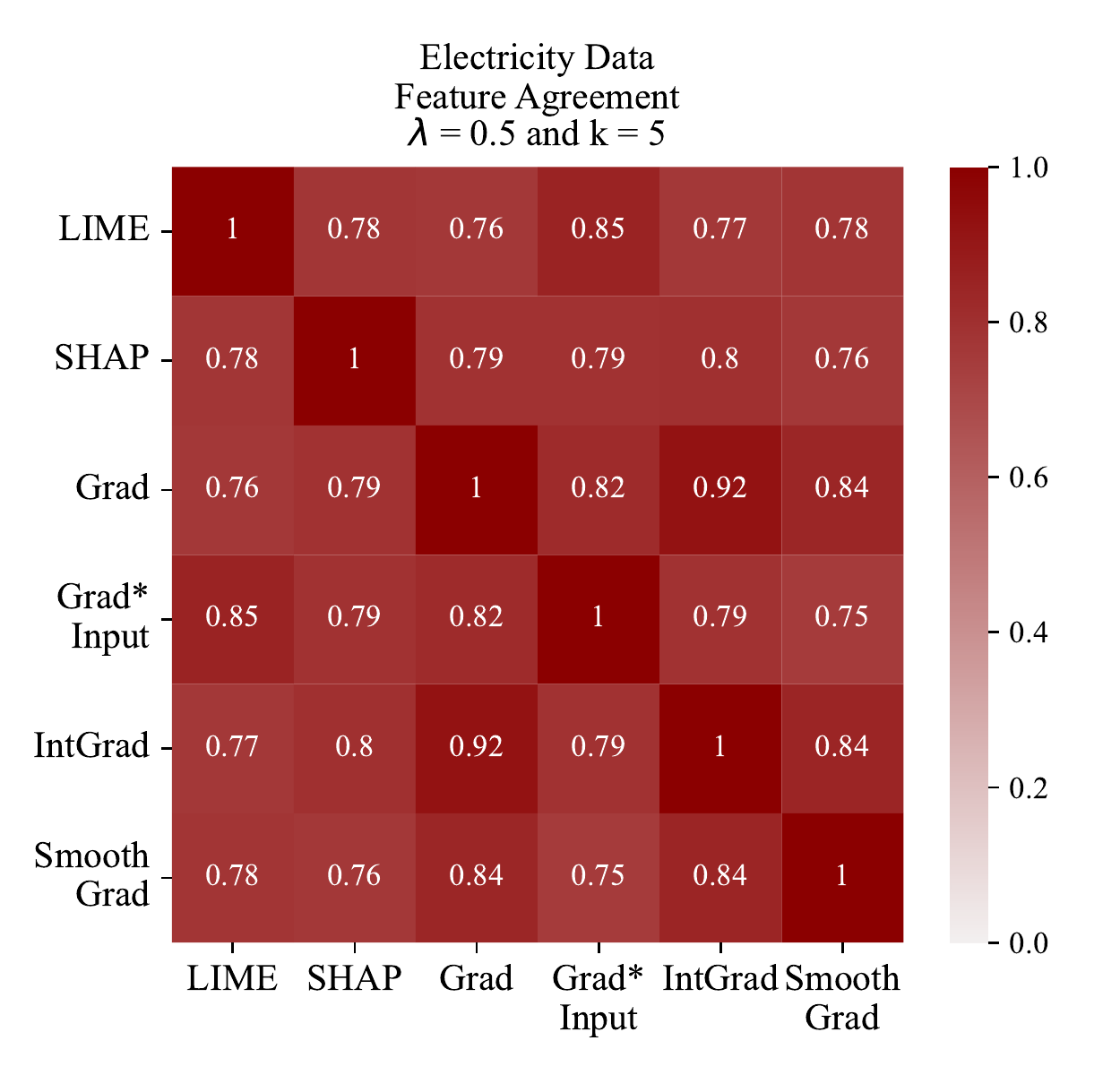}\\
\includegraphics[width=0.13\textwidth]{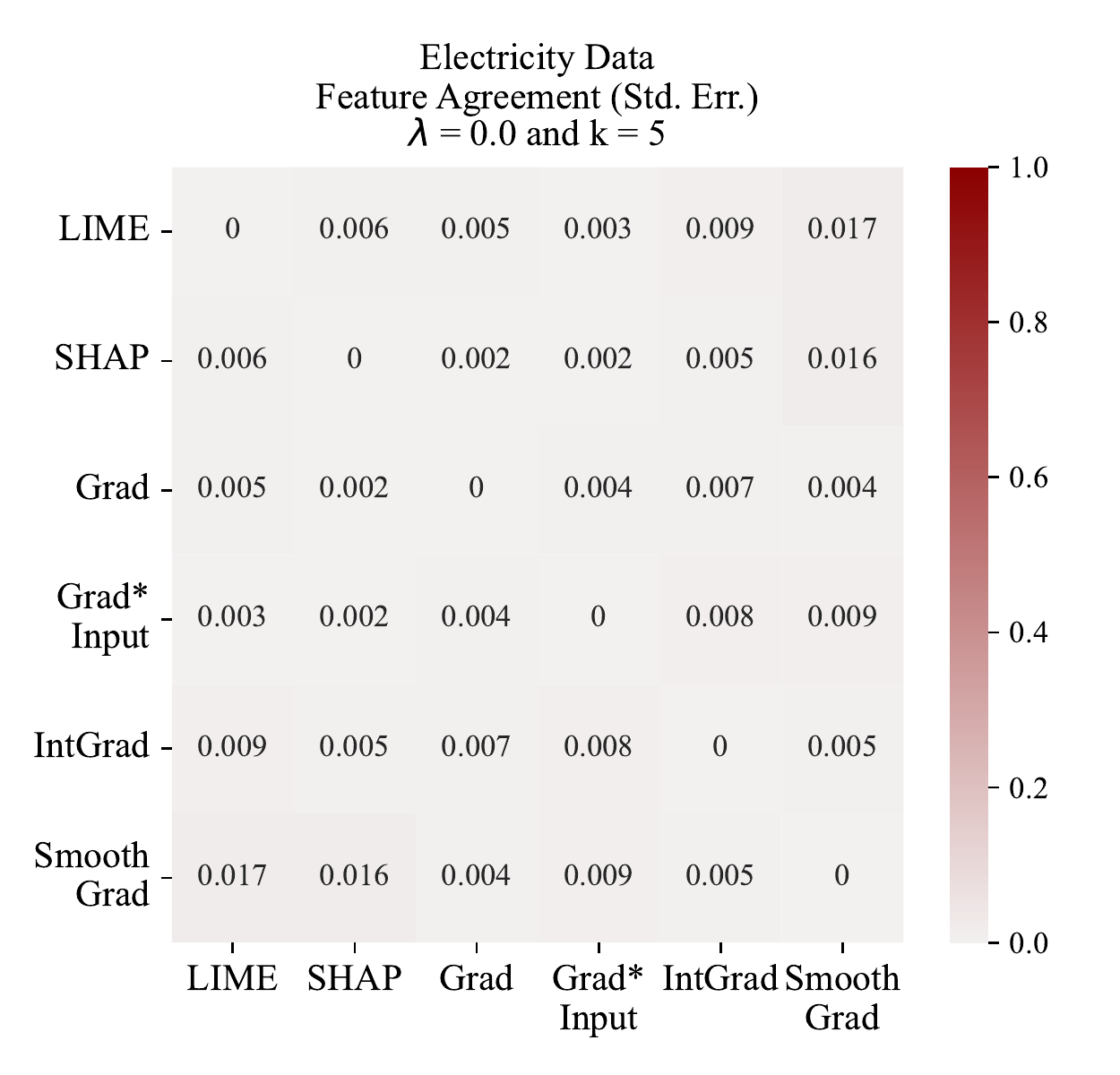}
\includegraphics[width=0.13\textwidth]{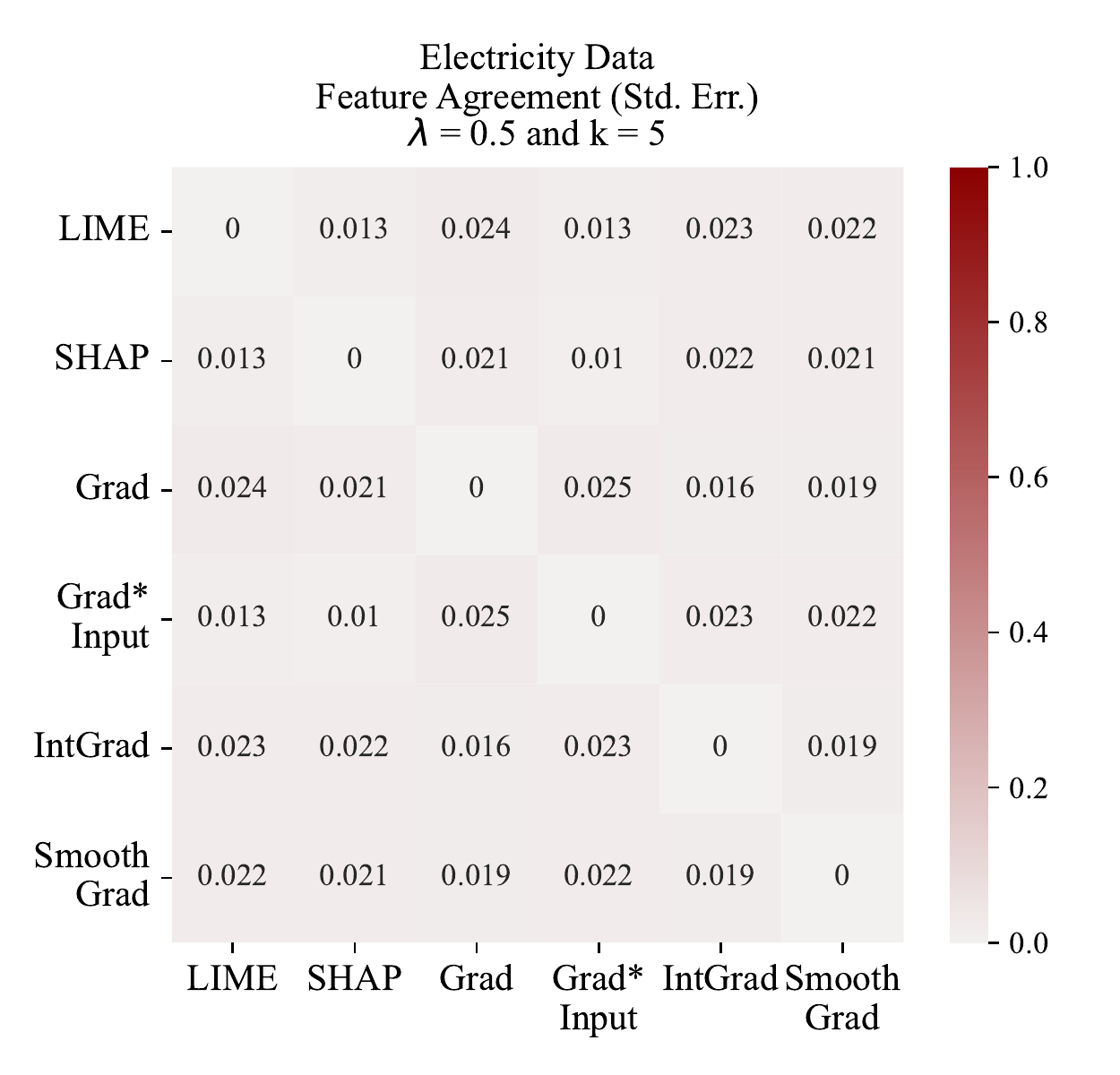}
}}
\fbox{
\parbox[c]{0.28\textwidth}{
\includegraphics[width=0.13\textwidth]{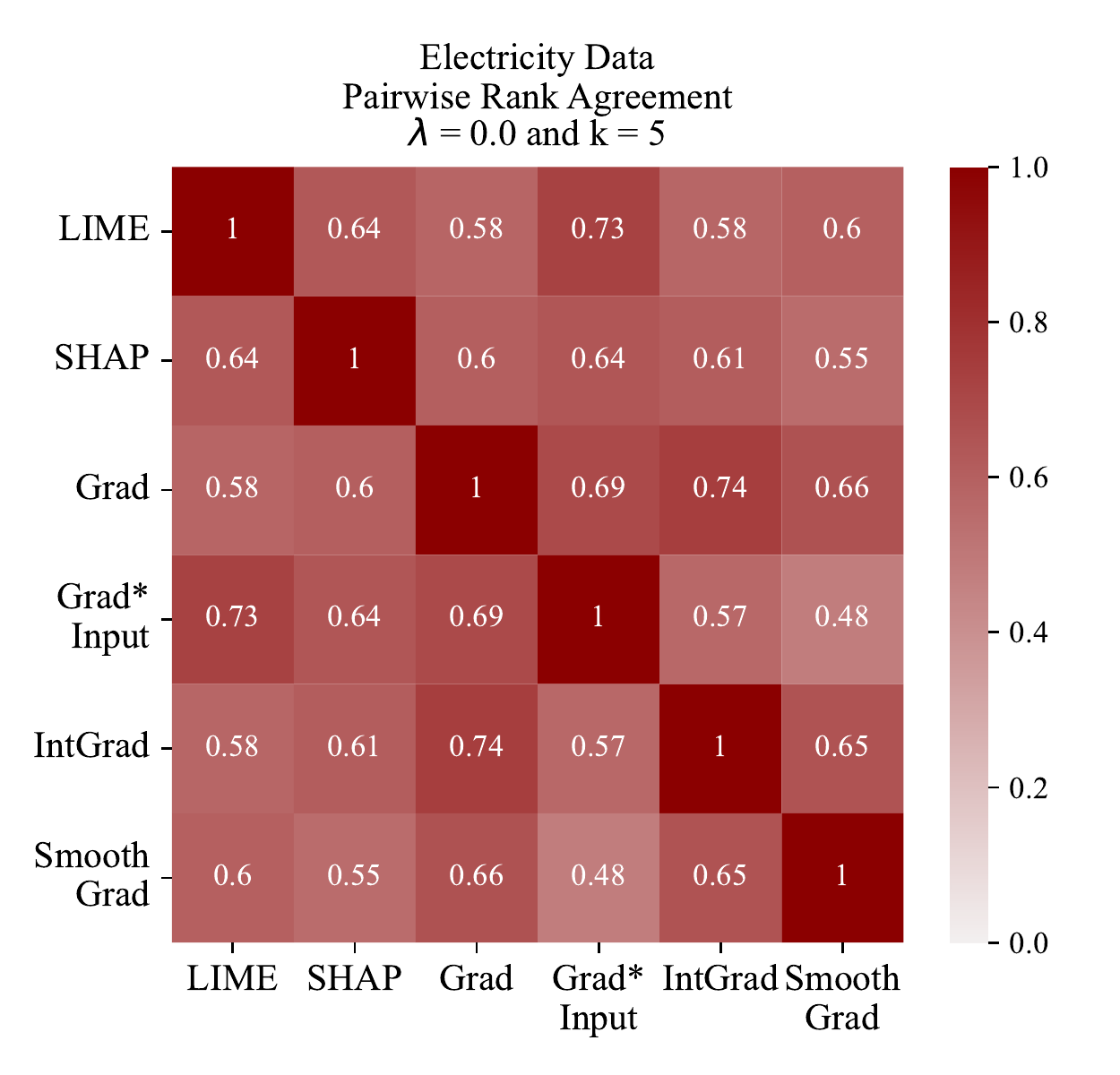}
\includegraphics[width=0.13\textwidth]{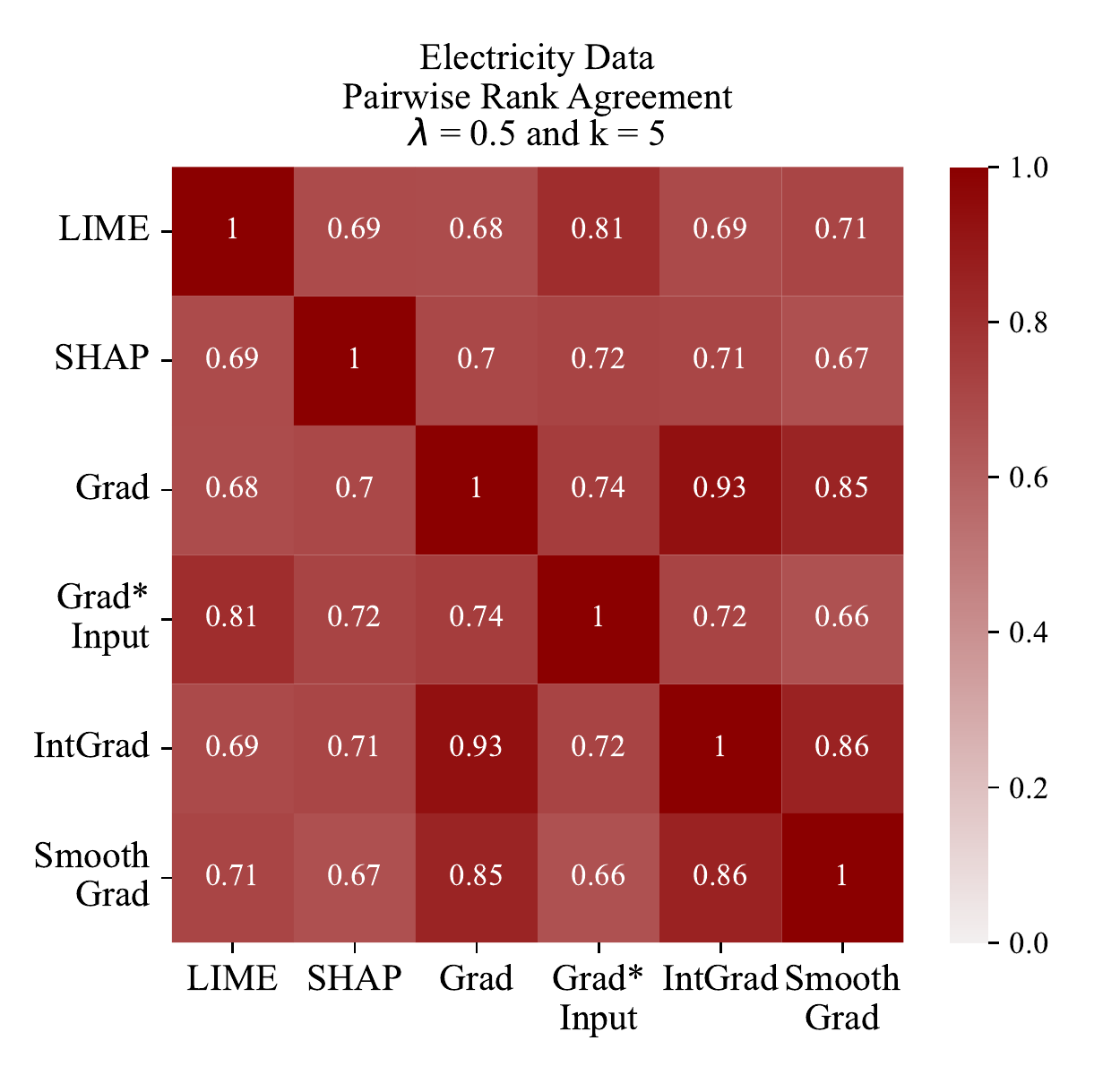}\\
\includegraphics[width=0.13\textwidth]{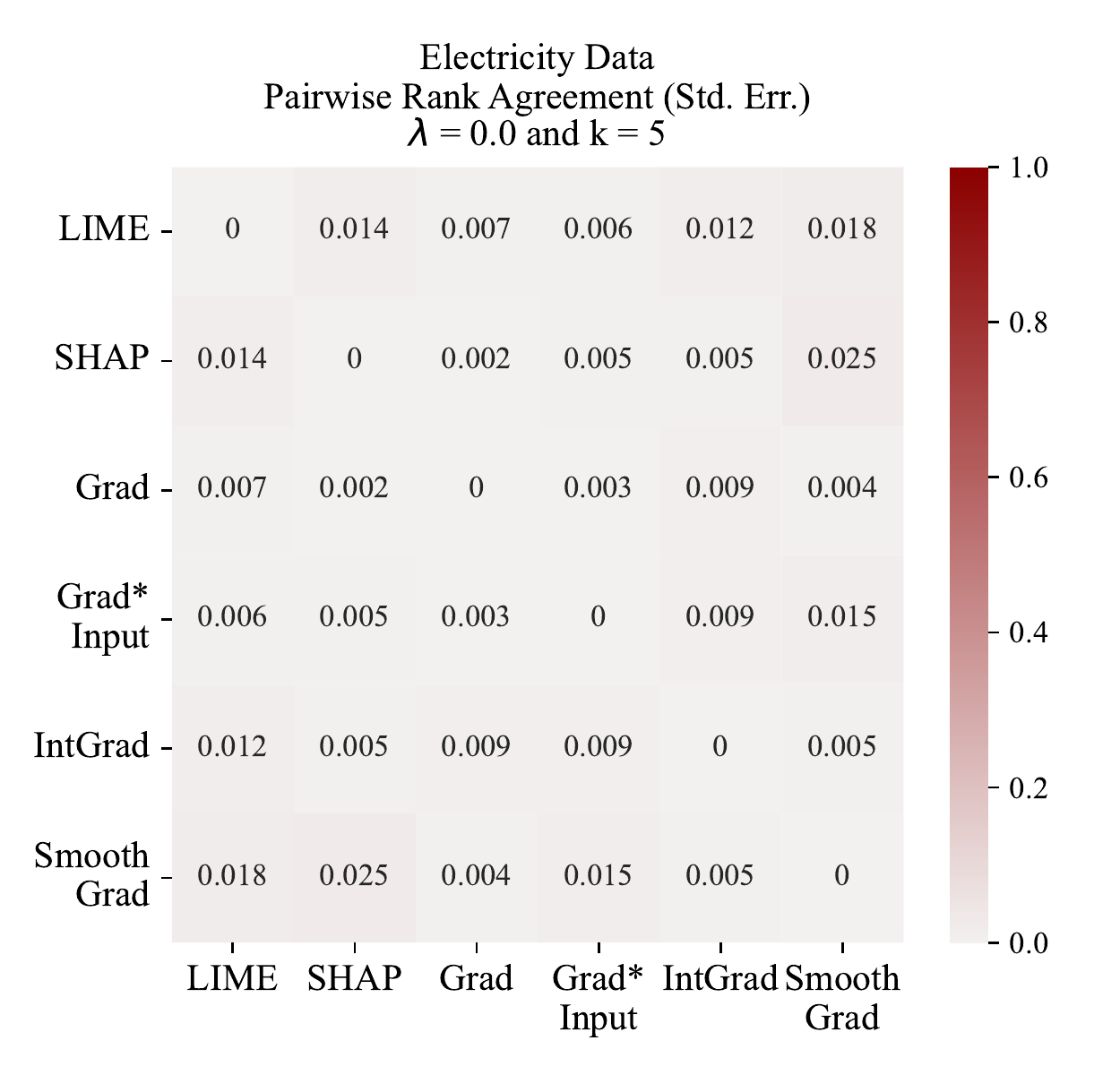}
\includegraphics[width=0.13\textwidth]{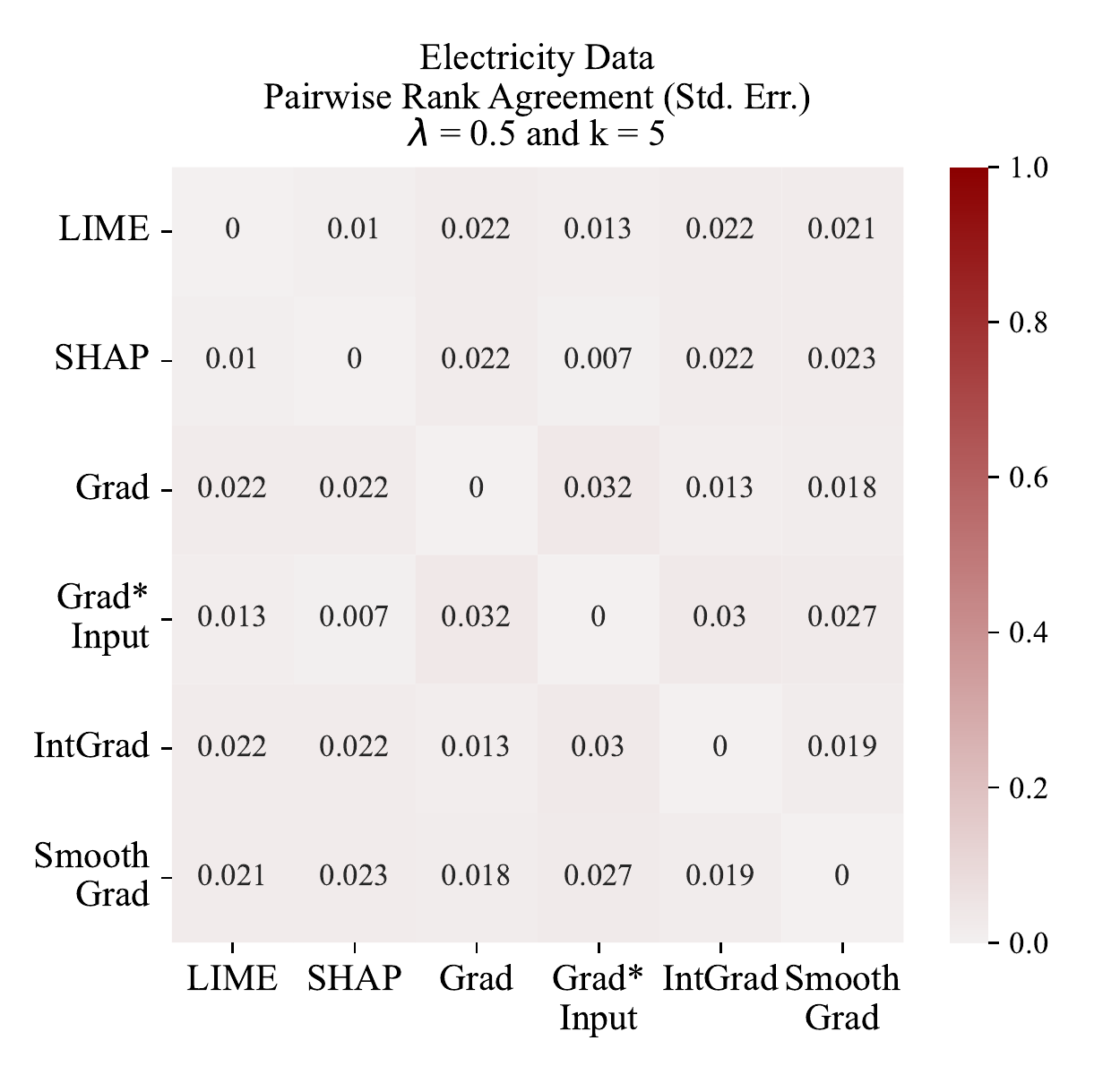}
}}
\fbox{
\parbox[c]{0.28\textwidth}{
\includegraphics[width=0.13\textwidth]{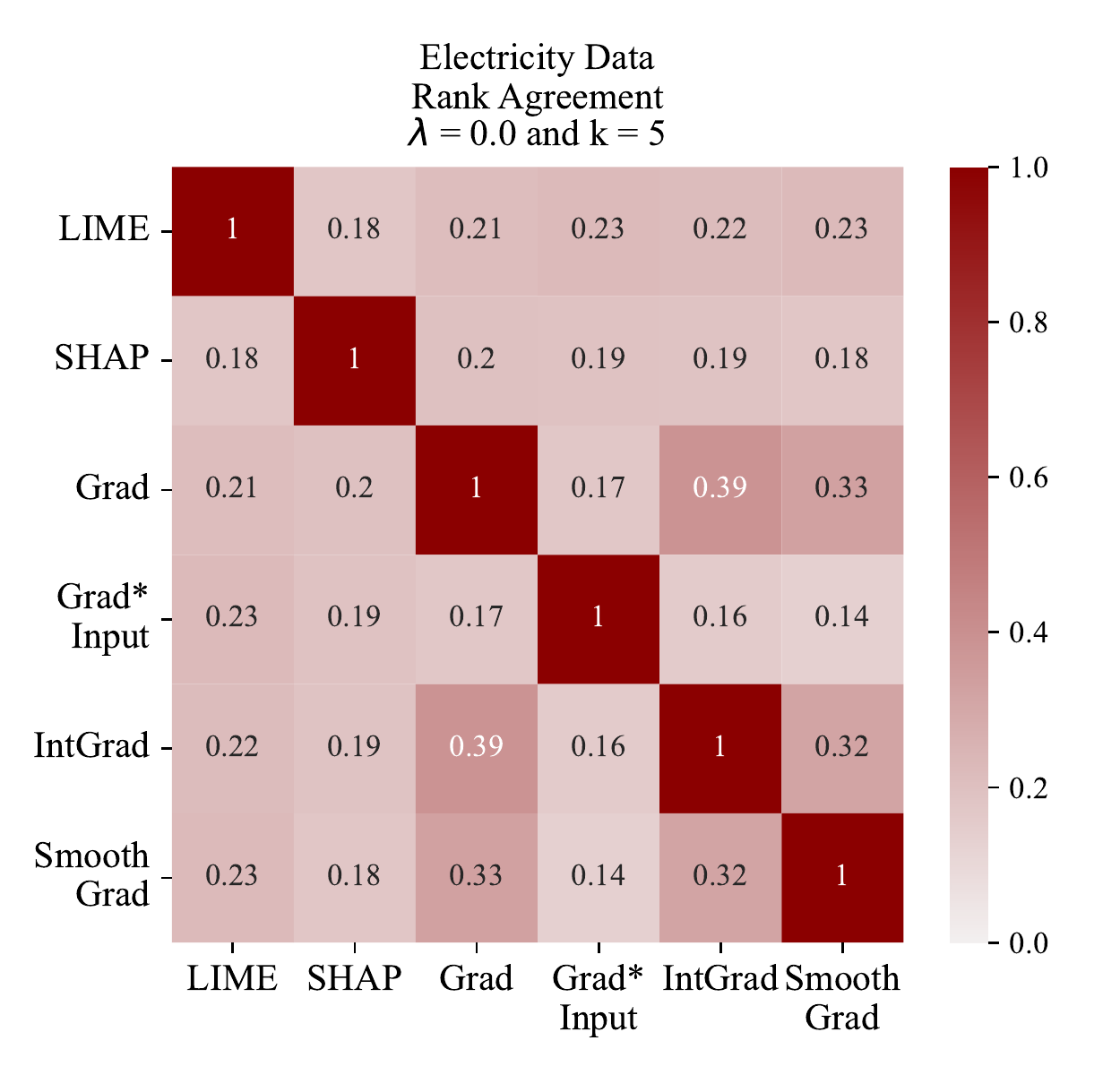}
\includegraphics[width=0.13\textwidth]{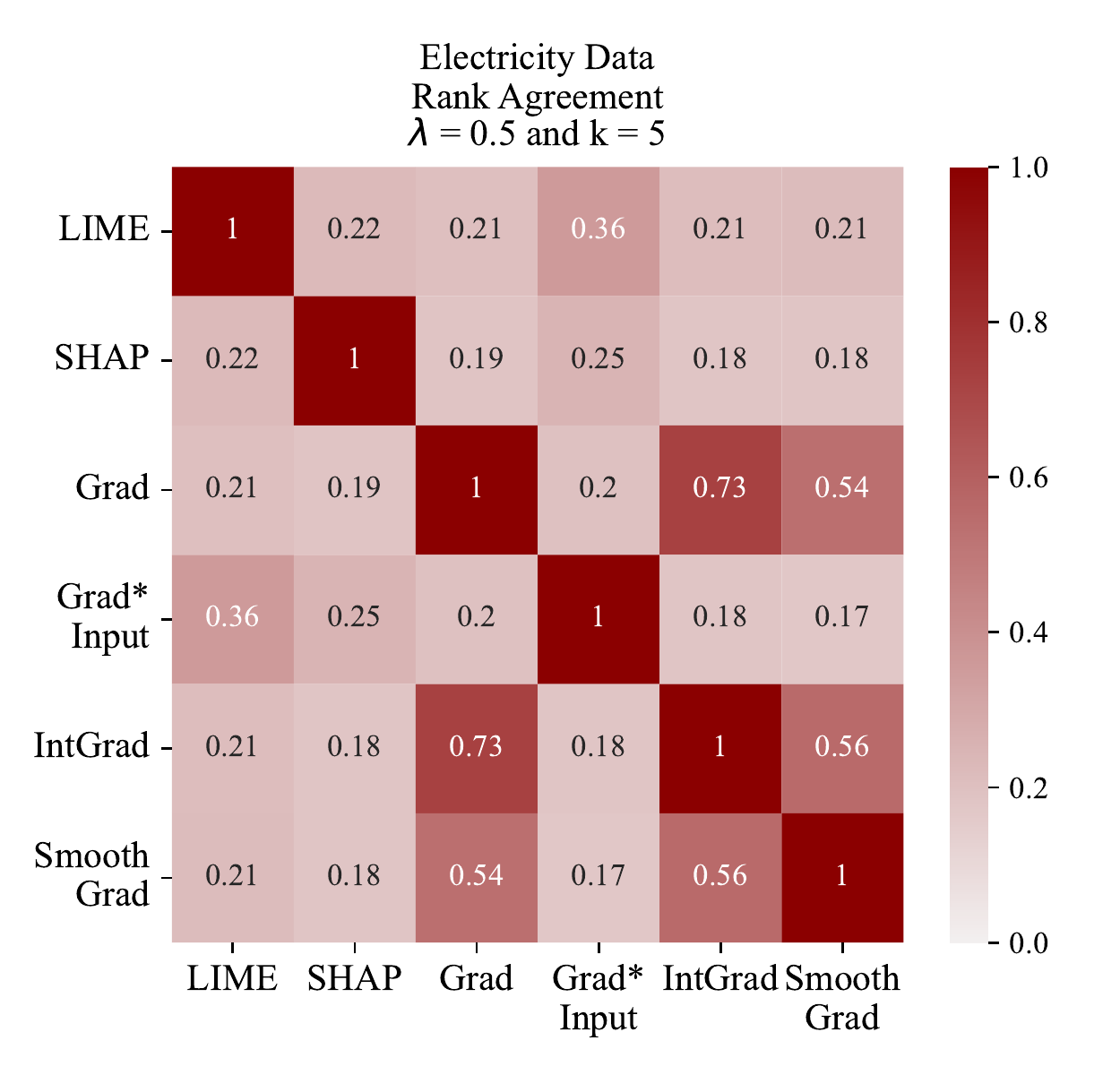}\\
\includegraphics[width=0.13\textwidth]{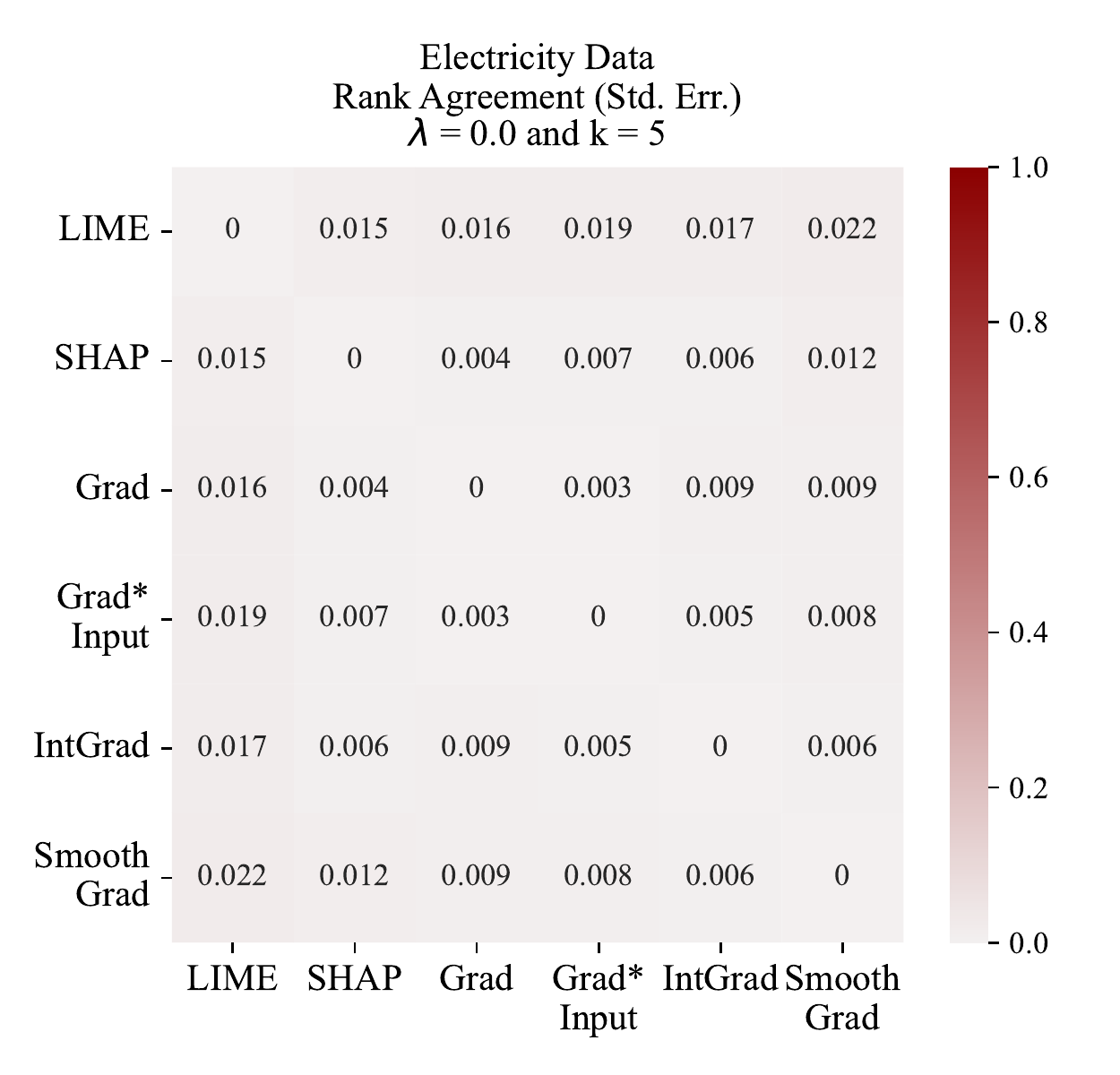}
\includegraphics[width=0.13\textwidth]{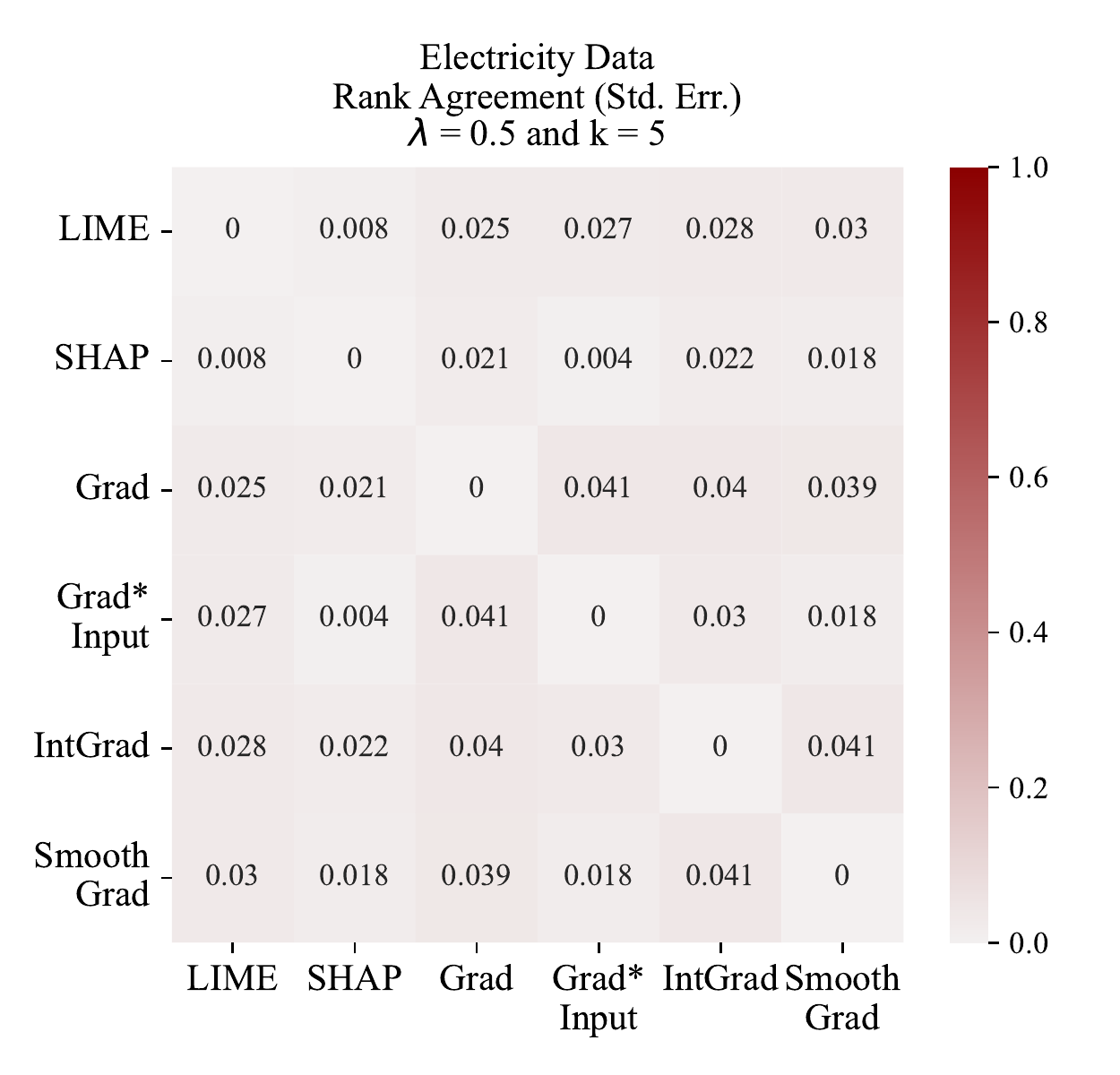}
}}
\\
\fbox{
\parbox[c]{0.28\textwidth}{
\includegraphics[width=0.13\textwidth]{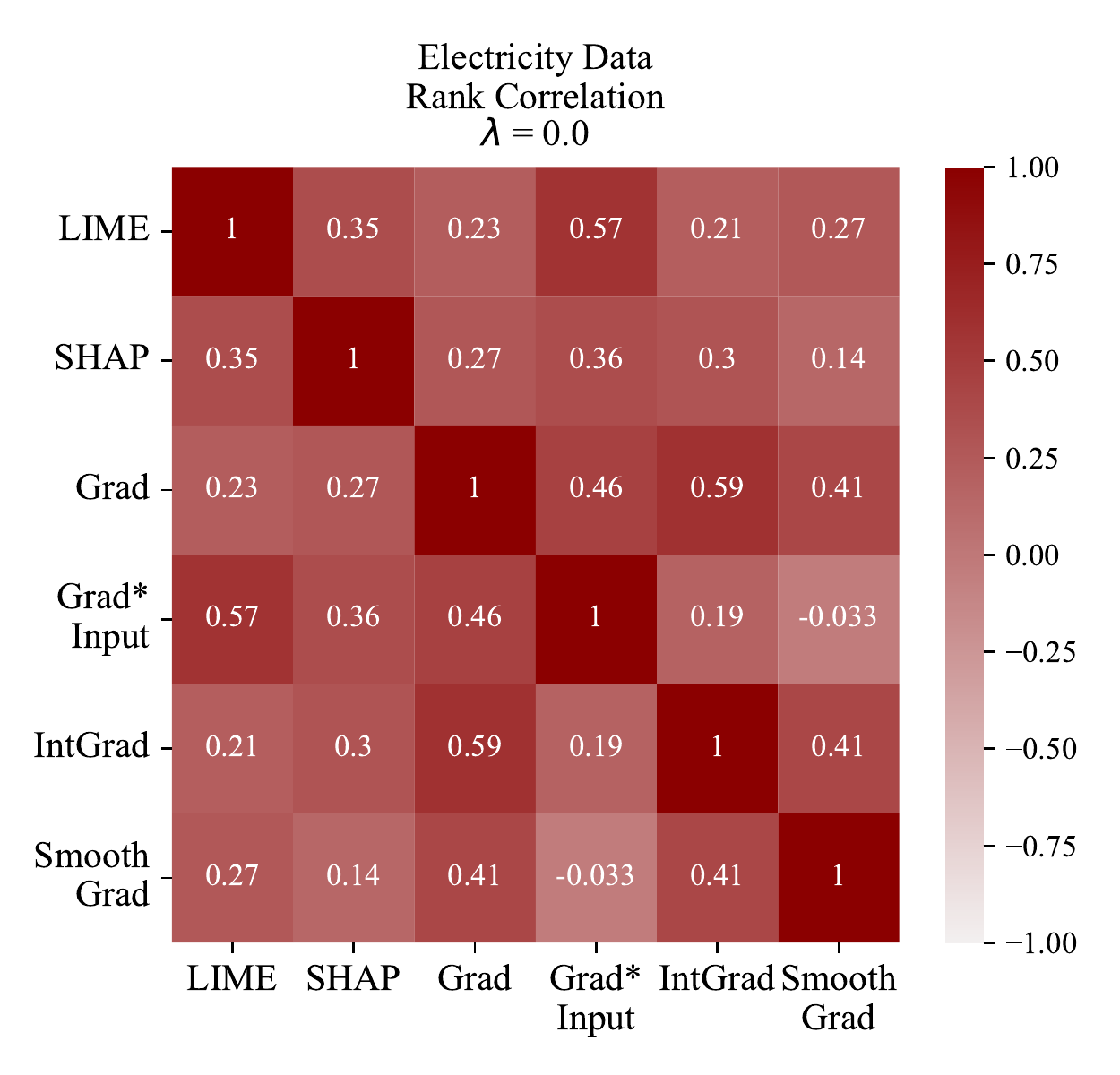}
\includegraphics[width=0.13\textwidth]{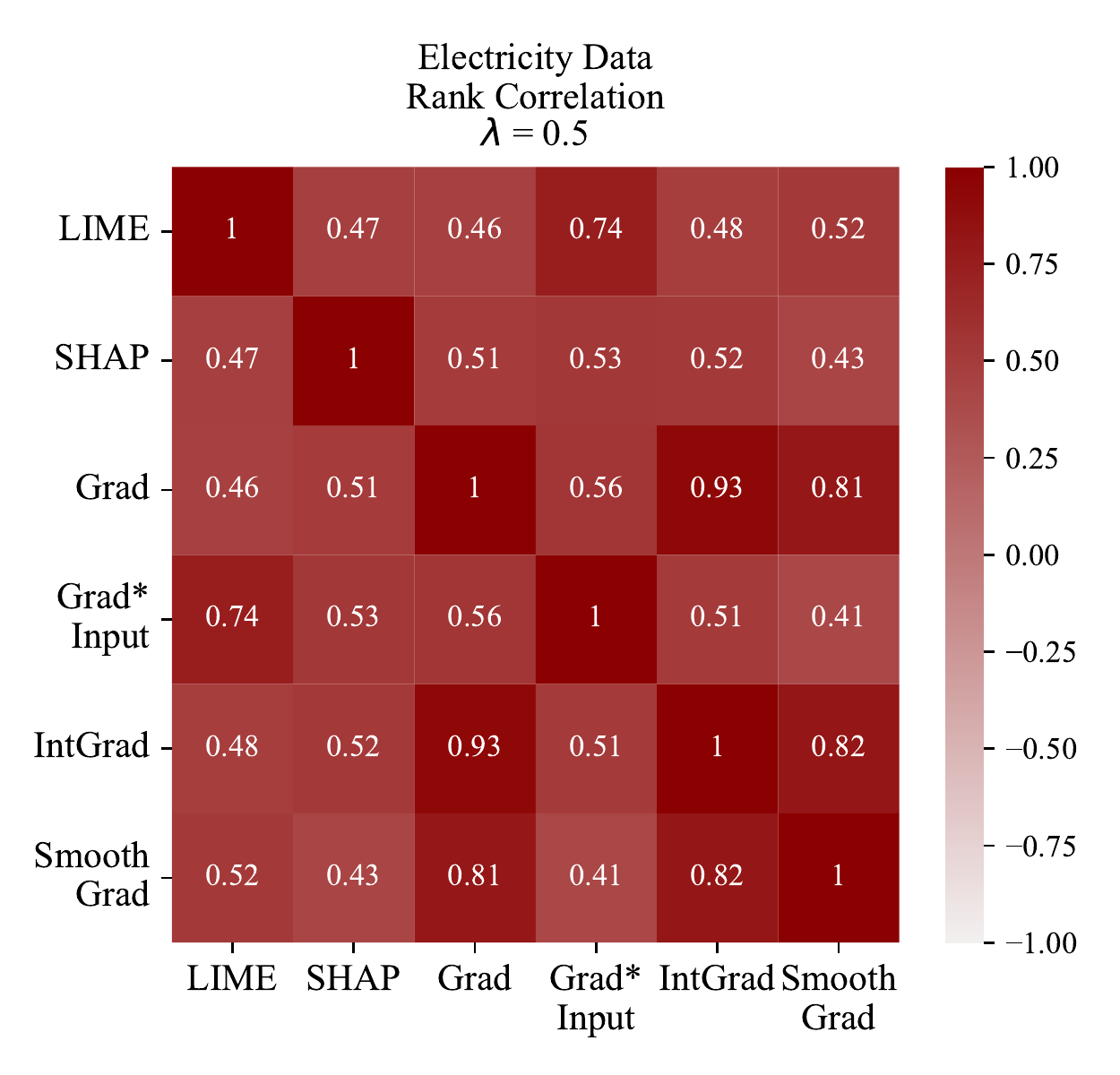}\\
\includegraphics[width=0.13\textwidth]{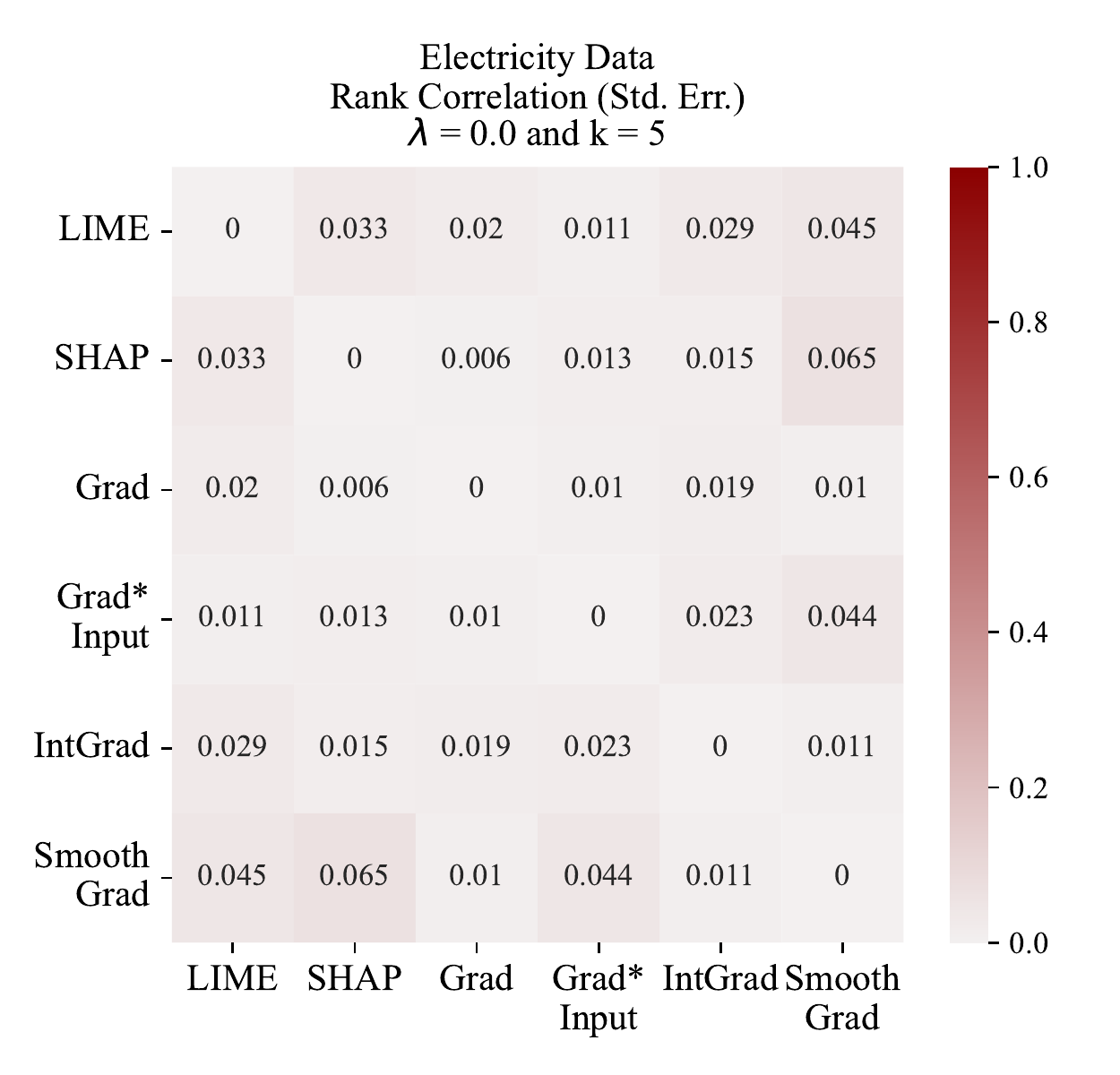}
\includegraphics[width=0.13\textwidth]{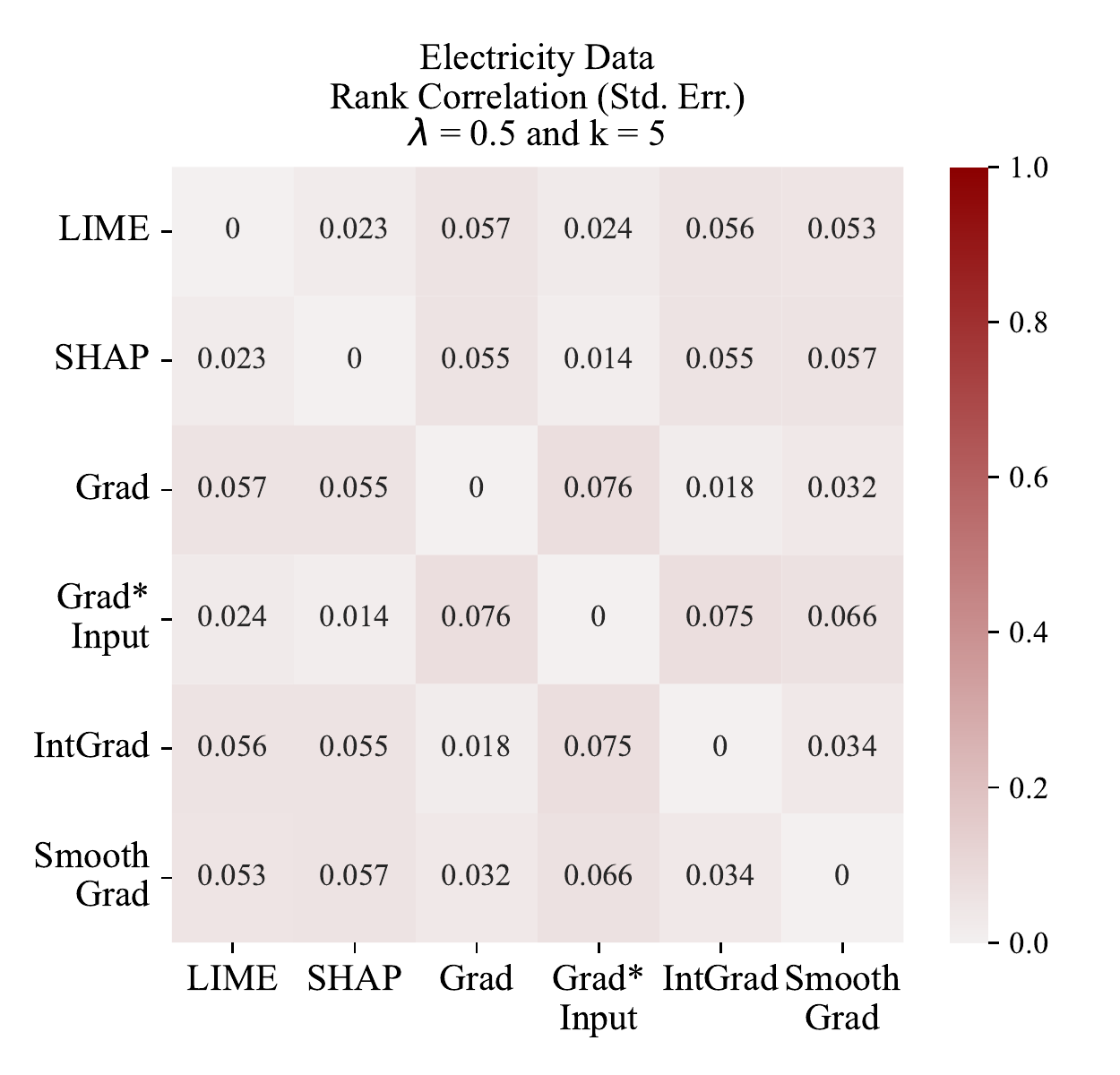}
}}
\fbox{
\parbox[c]{0.28\textwidth}{
\includegraphics[width=0.13\textwidth]{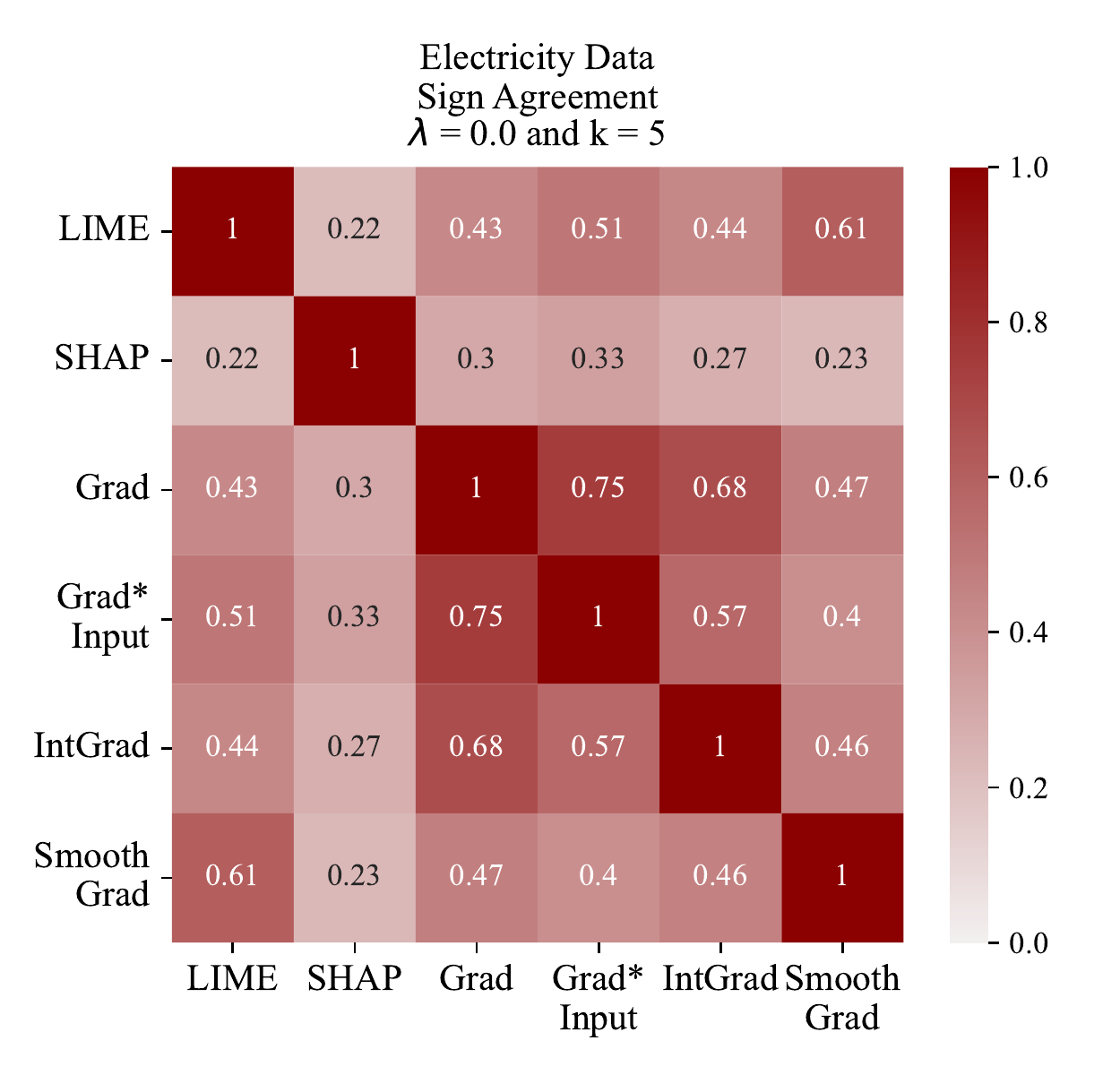}
\includegraphics[width=0.13\textwidth]{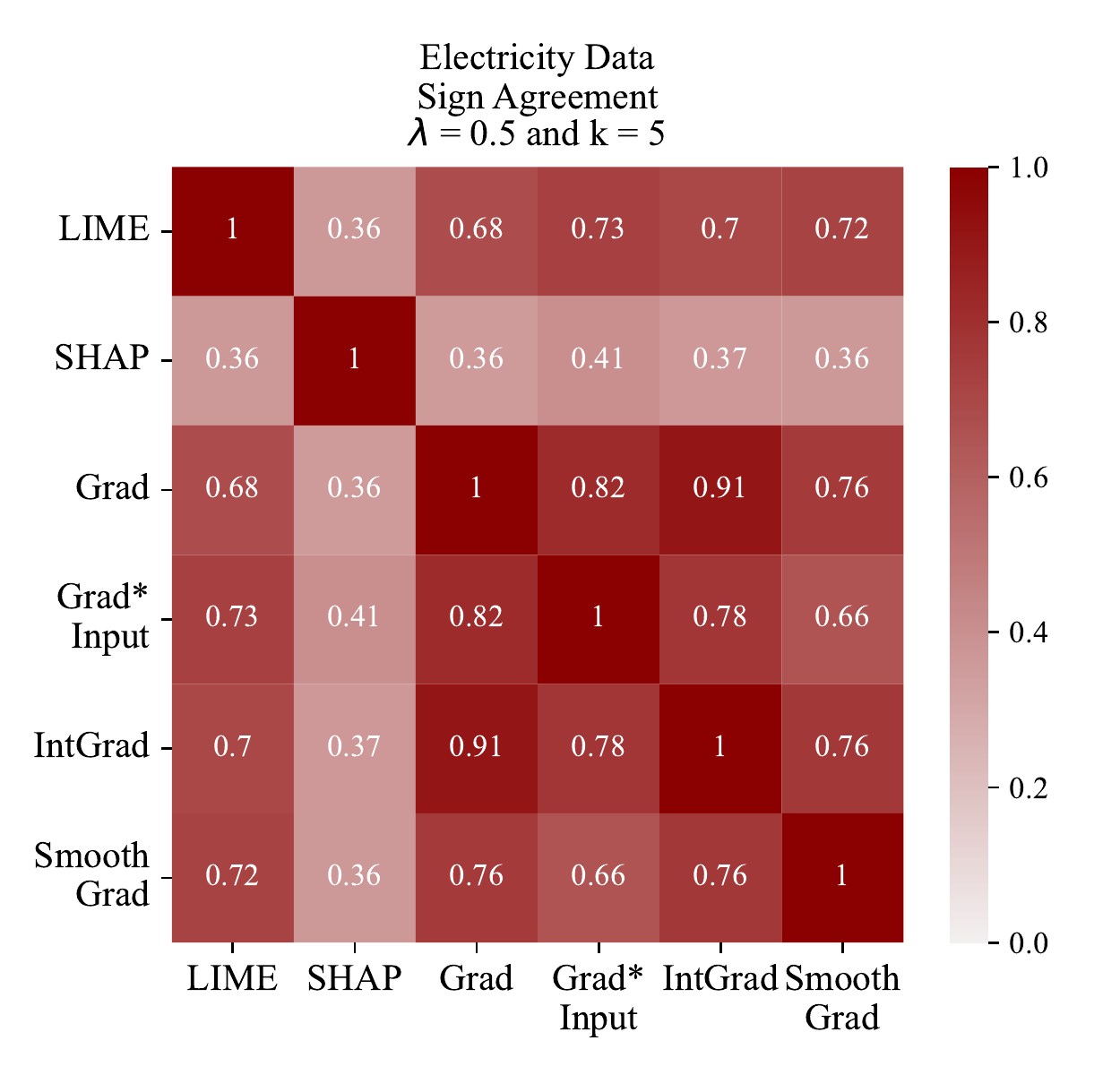}\\
\includegraphics[width=0.13\textwidth]{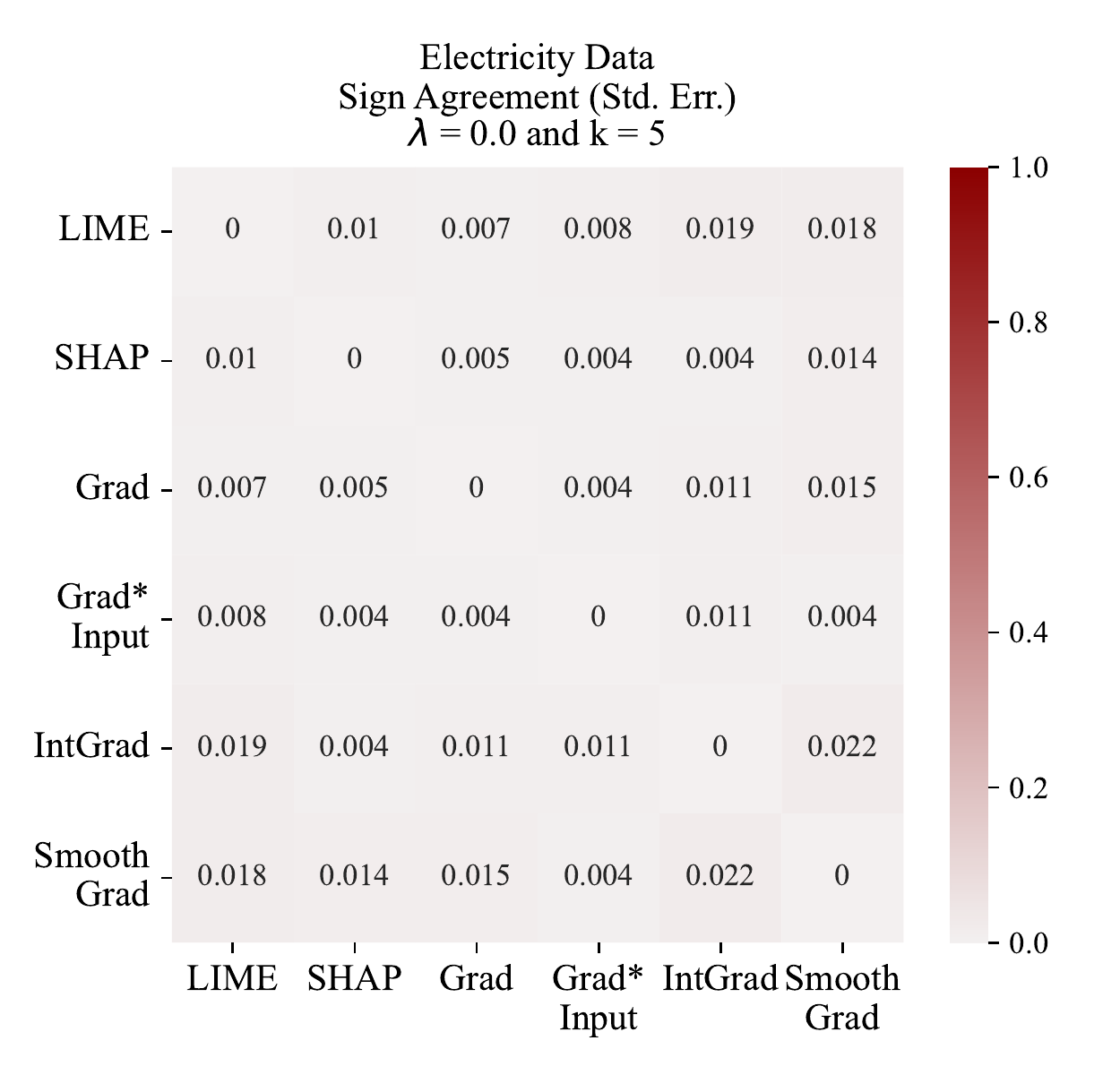}
\includegraphics[width=0.13\textwidth]{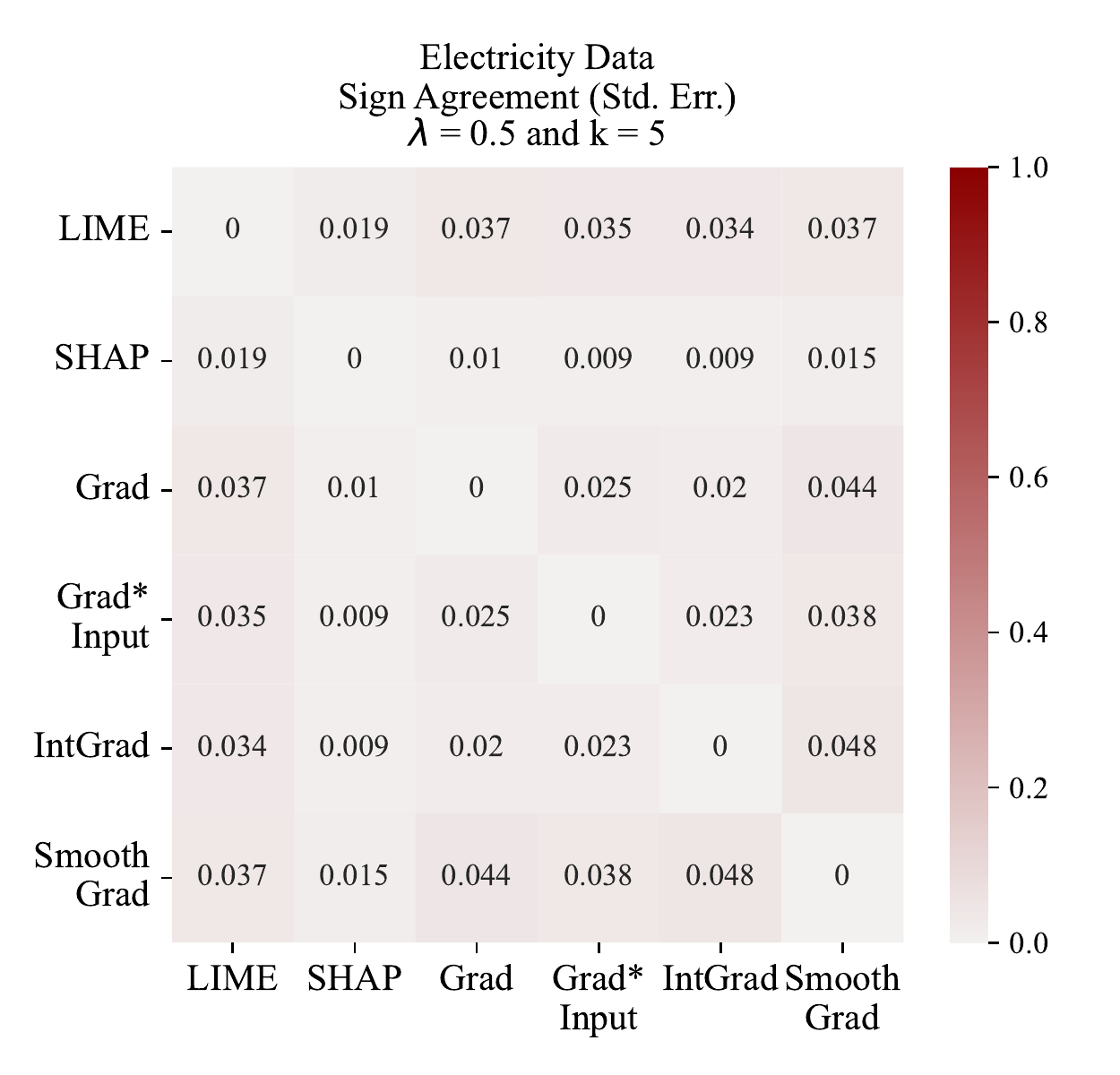}
}}
\fbox{
\parbox[c]{0.28\textwidth}{
\includegraphics[width=0.13\textwidth]{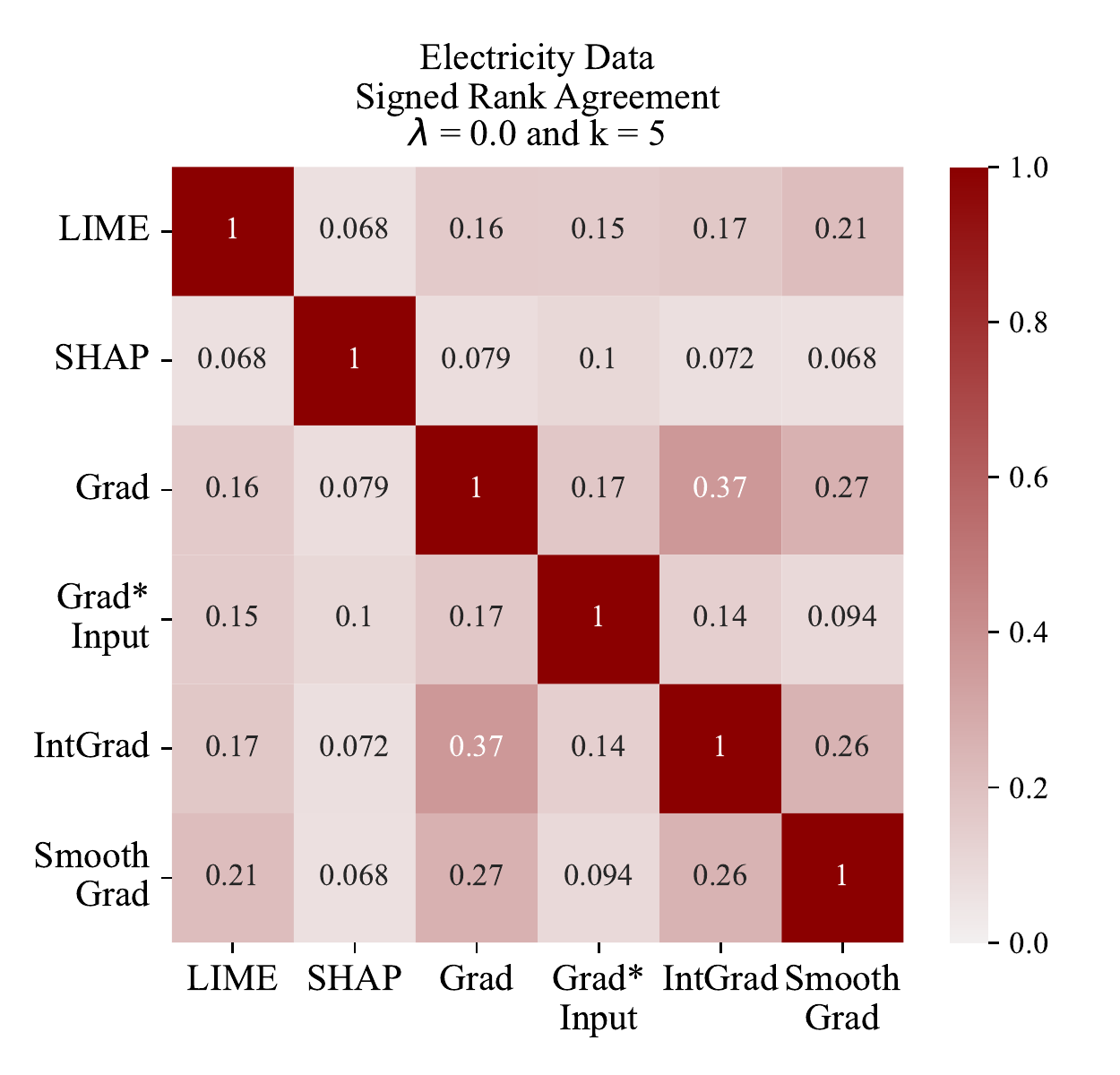}
\includegraphics[width=0.13\textwidth]{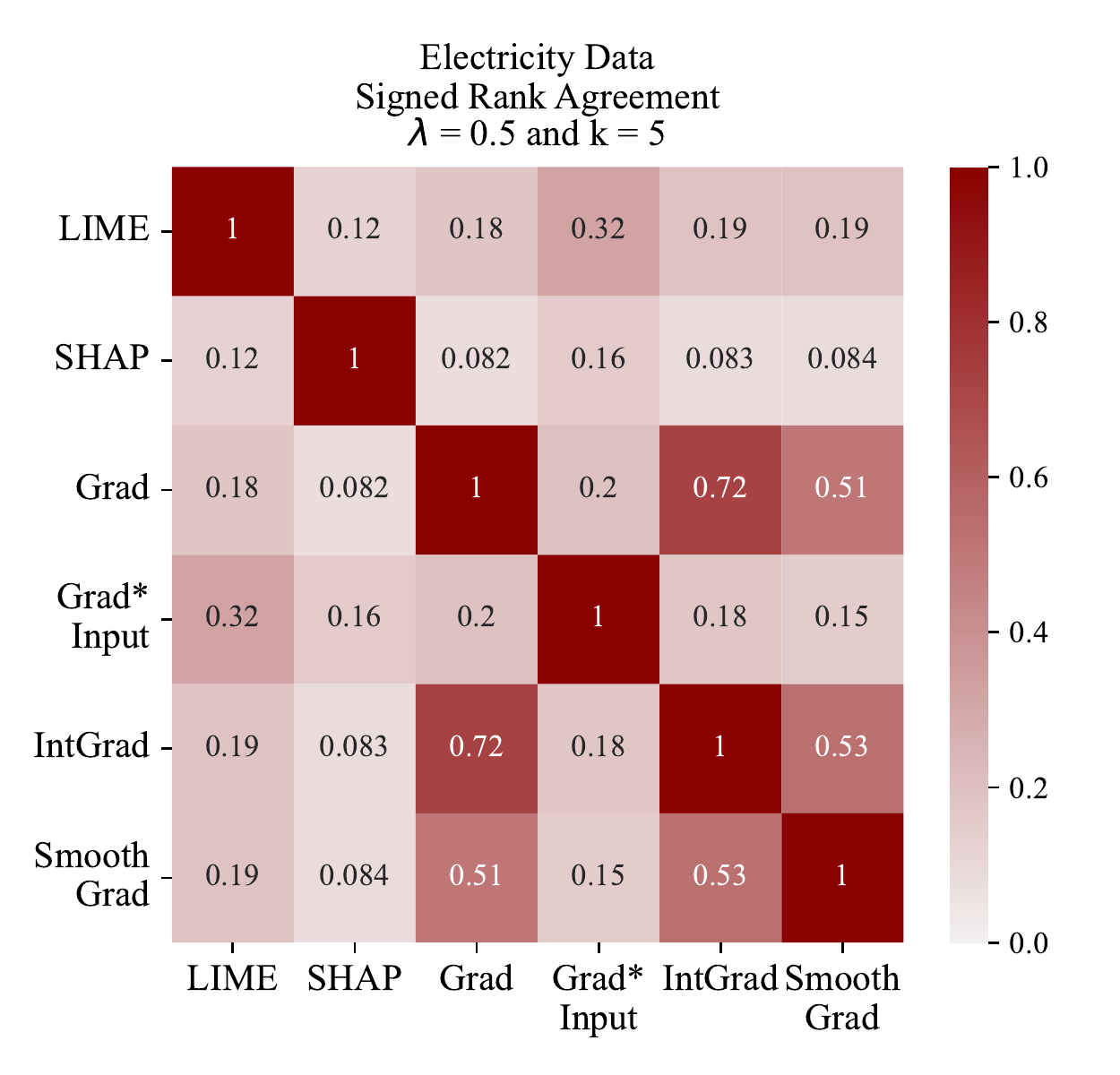}\\
\includegraphics[width=0.13\textwidth]{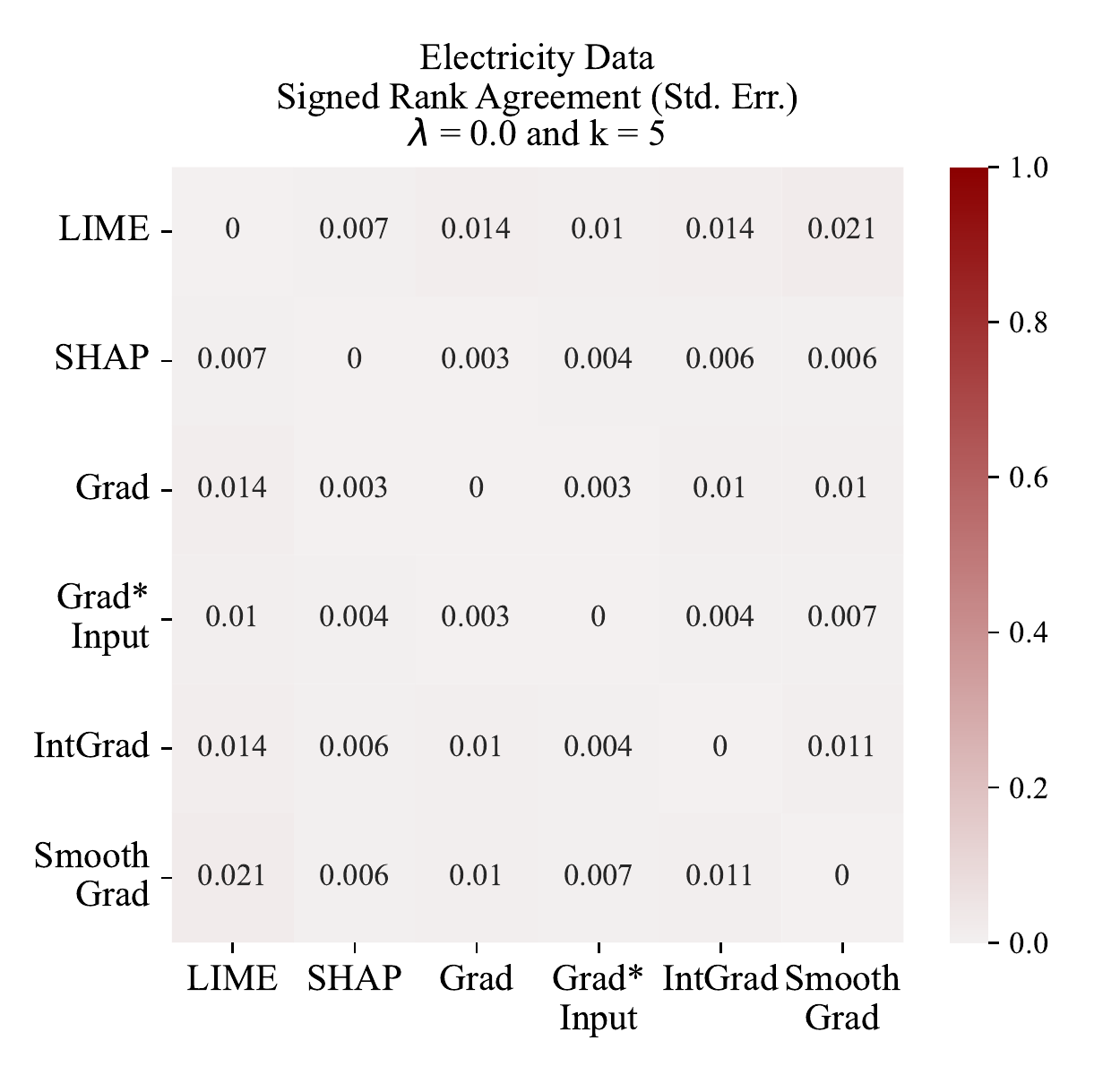}
\includegraphics[width=0.13\textwidth]{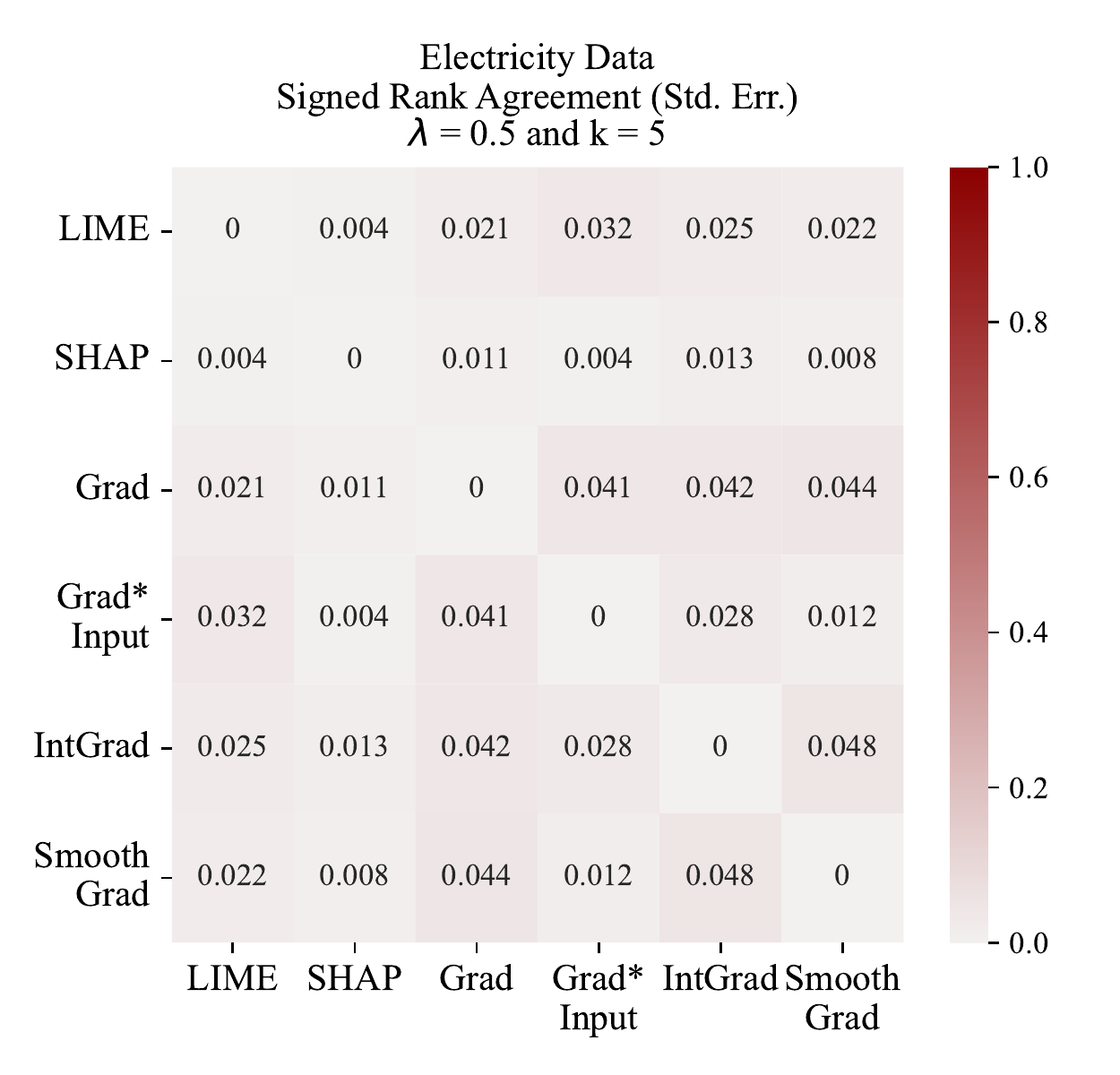}
}}
\caption{Disagreement matrices for all metrics considered in this paper on Electricity data.}
\label{fig:more-redgrids-electricity}
\end{figure*}
\newpage
\subsection{Extended Results}
\label{sec:app-std-error}

\begin{table*}[!ht]
    \centering
    \caption{Average test accuracy for models we trained. This table is organized by dataset, model, the hyperparameters in the loss, and the weight decay coefficient (WD). Averages are over several trials and we report the means $\pm$ one standard error.}
    \label{tab:ext-results}
    \small
    \begin{tabular}{llcccc}
        \toprule
        Dataset & Model & $\lambda$ & $\mu$ & WD & Accuracy \\
        \midrule
        Bank Marketing & Linear & 0.00 & 0.00 & 0.0002 & 74.3516 $\pm$ 0.1313 \\
                      & MLP     & 0.00 & 0.00 & 0.0002 & 79.0653 $\pm$ 0.2133 \\
                      & MLP     & 0.00 & 0.00 & 0.0020 & 78.9666 $\pm$ 0.4625 \\
                      & MLP     & 0.00 & 0.00 & 0.0200 & 79.1430 $\pm$ 0.4260 \\
                      & MLP     & 0.00 & 0.00 & 0.2000 & 79.1934 $\pm$ 0.1383 \\
                      & MLP     & 0.25 & 0.00 & 0.0002 & 79.2565 $\pm$ 0.1241 \\
                      & MLP     & 0.25 & 0.75 & 0.0002 & 79.3321 $\pm$ 0.1265 \\
                      & MLP     & 0.25 & 1.00 & 0.0002 & 79.2691 $\pm$ 0.5393 \\
                      & MLP     & 0.50 & 0.00 & 0.0002 & 79.4707 $\pm$ 0.1363 \\
                      & MLP     & 0.50 & 0.75 & 0.0002 & 79.0086 $\pm$ 0.0882 \\
                      & MLP     & 0.50 & 1.00 & 0.0002 & 79.1934 $\pm$ 0.1241 \\
                      & MLP     & 0.75 & 0.00 & 0.0002 & 78.7902 $\pm$ 0.1865 \\
                      & MLP     & 0.75 & 0.75 & 0.0002 & 77.8618 $\pm$ 0.4173 \\
                      & MLP     & 0.75 & 1.00 & 0.0002 & 77.5299 $\pm$ 0.6848 \\
        \midrule
        California Housing & Linear & 0.00 & 0.00 & 0.0002 & 81.5352 $\pm$ 0.1819 \\
                          & MLP     & 0.00 & 0.00 & 0.0002 & 84.8580 $\pm$ 0.1768 \\
                          & MLP     & 0.00 & 0.00 & 0.0020 & 84.6159 $\pm$ 0.1275 \\
                          & MLP     & 0.00 & 0.00 & 0.0200 & 84.5448 $\pm$ 0.2128 \\
                          & MLP     & 0.00 & 0.00 & 0.2000 & 84.3639 $\pm$ 0.3306 \\
                          & MLP     & 0.25 & 0.00 & 0.0002 & 81.7471 $\pm$ 0.8670 \\
                          & MLP     & 0.25 & 0.75 & 0.0002 & 83.5821 $\pm$ 0.1443 \\
                          & MLP     & 0.25 & 1.00 & 0.0002 & 84.1442 $\pm$ 0.3780 \\
                          & MLP     & 0.50 & 0.00 & 0.0002 & 80.2546 $\pm$ 0.4983 \\
                          & MLP     & 0.50 & 0.75 & 0.0002 & 83.1595 $\pm$ 0.2225 \\
                          & MLP     & 0.50 & 1.00 & 0.0002 & 83.7178 $\pm$ 0.1902 \\
                          & MLP     & 0.75 & 0.00 & 0.0002 & 82.7874 $\pm$ 0.7604 \\
                          & MLP     & 0.75 & 0.75 & 0.0002 & 82.4578 $\pm$ 0.3826 \\
                          & MLP     & 0.75 & 1.00 & 0.0002 & 81.7859 $\pm$ 0.6012 \\
        \midrule
        Electricity & Linear & 0.00 & 0.00 & 0.0002 & 73.3382 $\pm$ 0.1500 \\
                    & MLP    & 0.00 & 0.00 & 0.0002 & 81.2974 $\pm$ 0.1576 \\
                    & MLP    & 0.00 & 0.00 & 0.0020 & 81.1727 $\pm$ 0.2092 \\
                    & MLP    & 0.00 & 0.00 & 0.0200 & 81.5573 $\pm$ 0.1169 \\
                    & MLP    & 0.00 & 0.00 & 0.2000 & 76.9311 $\pm$ 0.5849 \\
                    & MLP    & 0.25 & 0.00 & 0.0002 & 81.5781 $\pm$ 0.1690 \\
                    & MLP    & 0.25 & 0.75 & 0.0002 & 80.5454 $\pm$ 0.1380 \\
                    & MLP    & 0.25 & 1.00 & 0.0002 & 80.9162 $\pm$ 0.5275 \\
                    & MLP    & 0.50 & 0.00 & 0.0002 & 81.4880 $\pm$ 0.1428 \\
                    & MLP    & 0.50 & 0.75 & 0.0002 & 80.0742 $\pm$ 0.1131 \\
                    & MLP    & 0.50 & 1.00 & 0.0002 & 79.6479 $\pm$ 0.4371 \\
                    & MLP    & 0.75 & 0.00 & 0.0002 & 80.6252 $\pm$ 0.1940 \\
                    & MLP    & 0.75 & 0.75 & 0.0002 & 79.0118 $\pm$ 0.4375 \\
                    & MLP    & 0.75 & 1.00 & 0.0002 & 78.6811 $\pm$ 0.6160 \\
    \end{tabular}
\end{table*}

\newpage
\newpage
\subsection{Additional Plots}
\label{sec:app-three-point}

\begin{figure*}[ht!]
    \centering
    \includegraphics[width=0.8\textwidth]{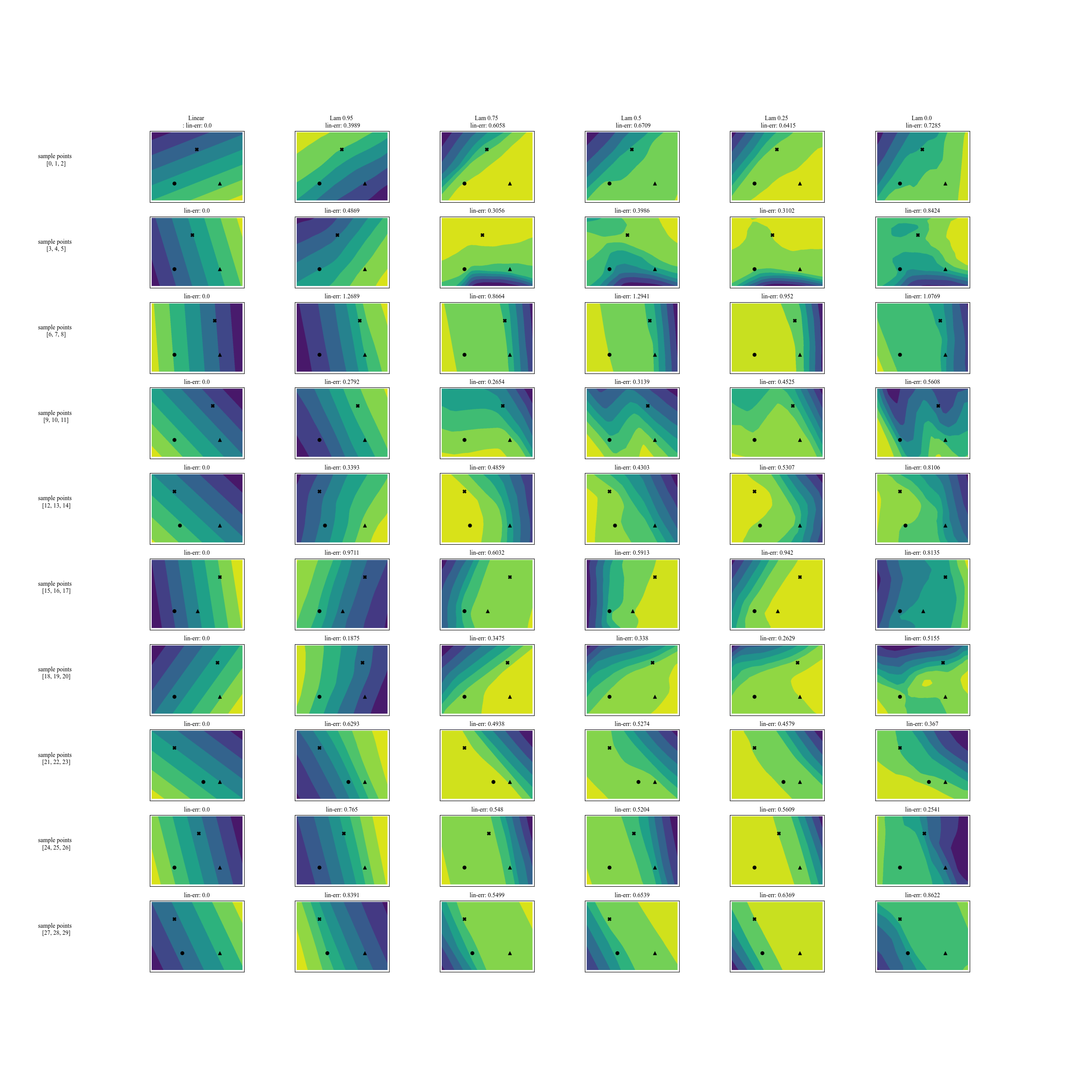}
    \caption{The logit surfaces for MLPs, each trained with a different lambda value, on 10 randomly construcuted three-point planes from the Bank Marketing dataset.}
    \label{fig:bankmarketing-logits-all}
\end{figure*}

\begin{figure*}[ht!]
    \centering
    \includegraphics[width=0.8\textwidth]{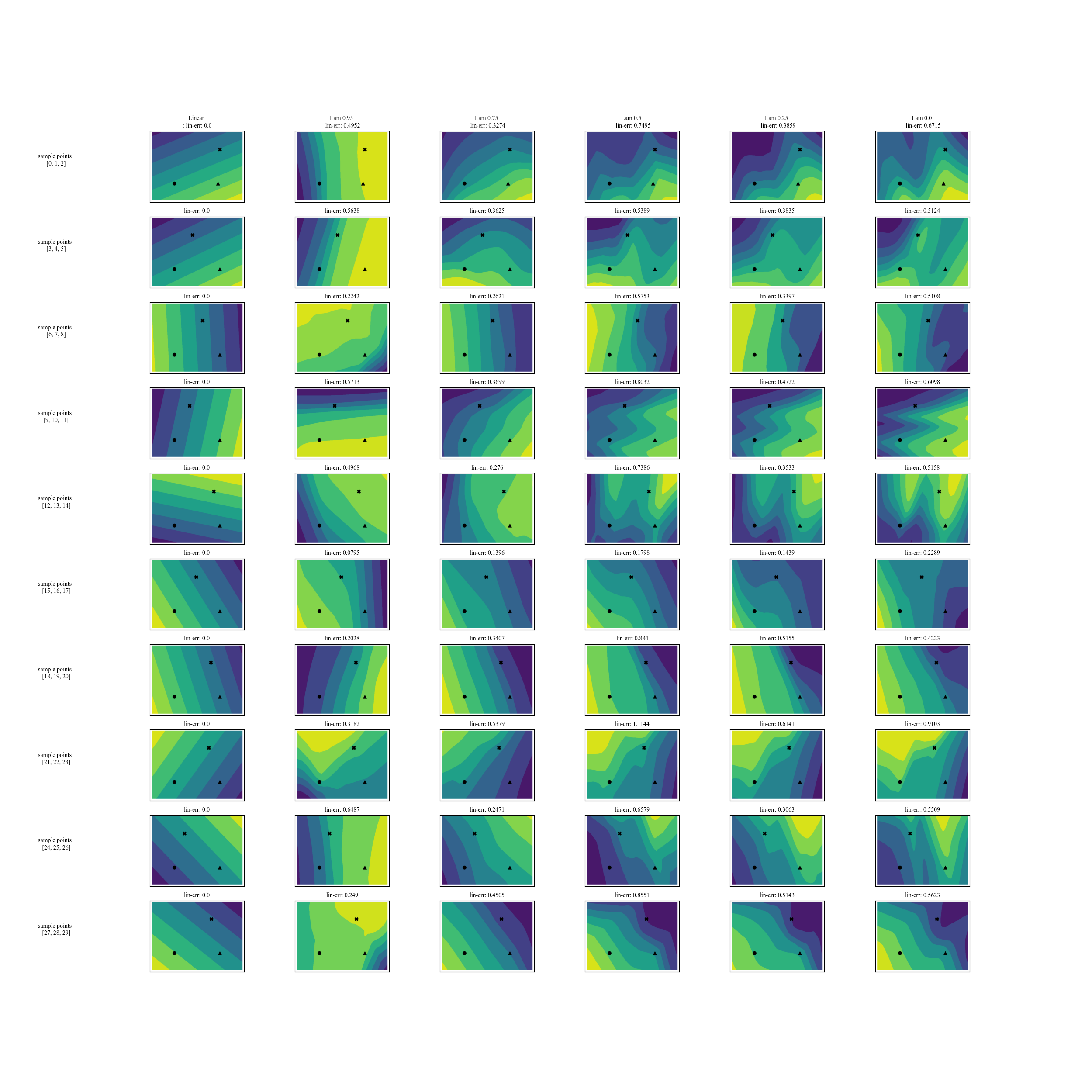}
    \caption{The logit surfaces for MLPs, each trained with a different lambda value, on 10 randomly construcuted three-point planes from the California Housing dataset.}
    \label{fig:cali-logits-all}
\end{figure*}

\begin{figure*}[ht!]
    \centering
    \includegraphics[width=0.8\textwidth]{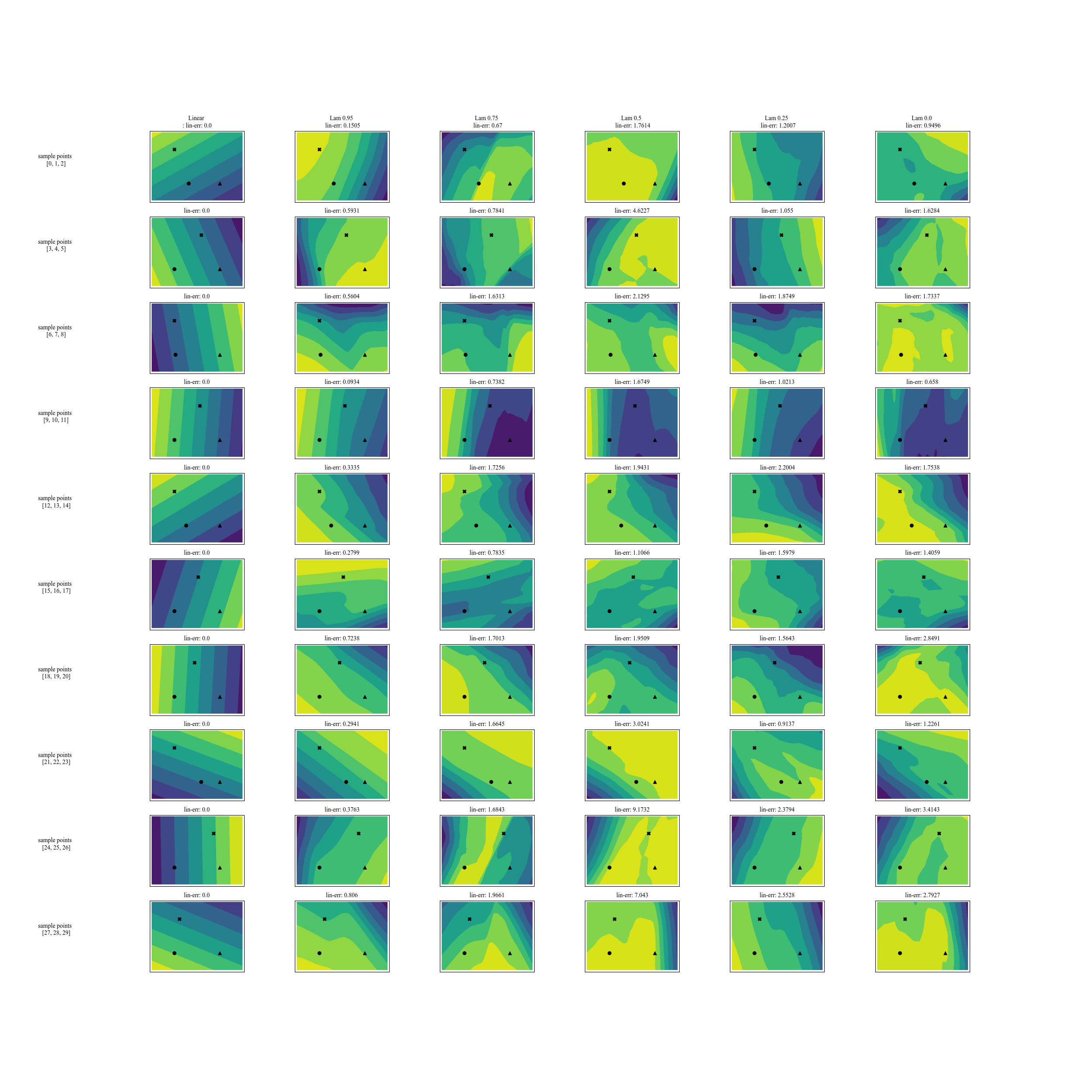}
    \caption{The logit surfaces for MLPs, each trained with a different lambda value, on 10 randomly construcuted three-point planes from the Electricity dataset.}
    \label{fig:elec-logits-all}
\end{figure*}

\begin{figure*}[ht!]
    \centering
    \includegraphics[width=0.275\textwidth]{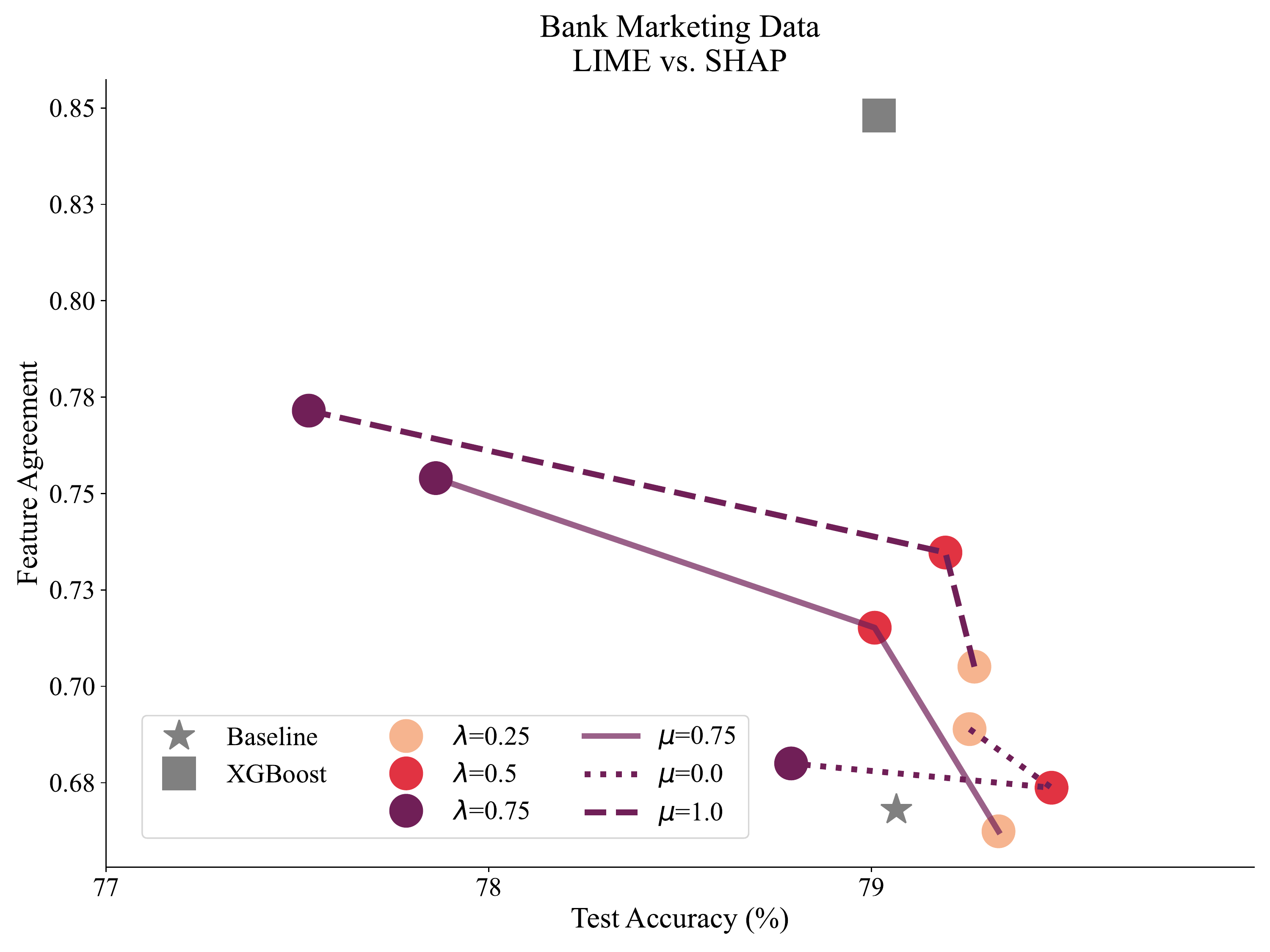}
    \includegraphics[width=0.275\textwidth]{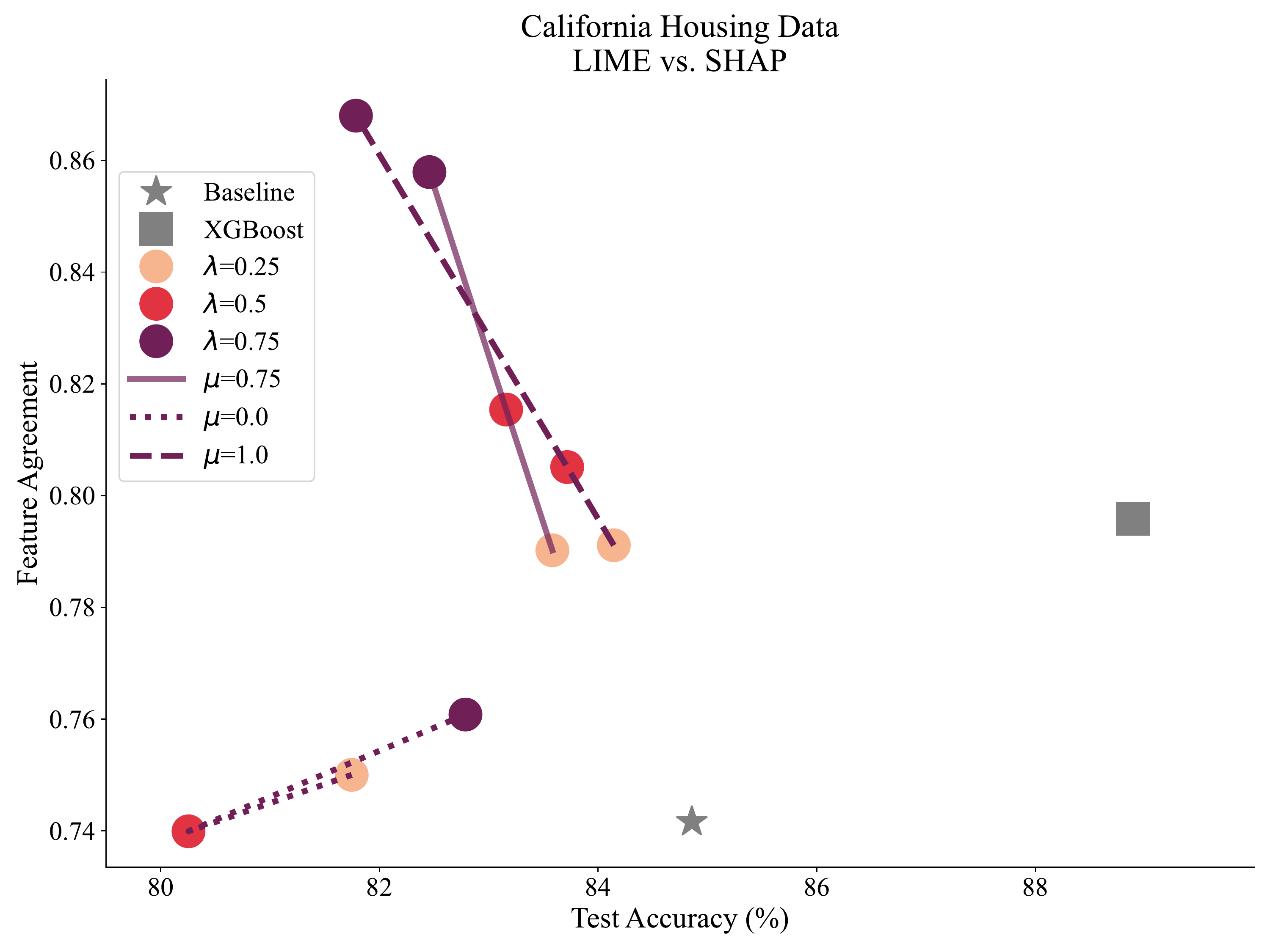}
    \includegraphics[width=0.275\textwidth]{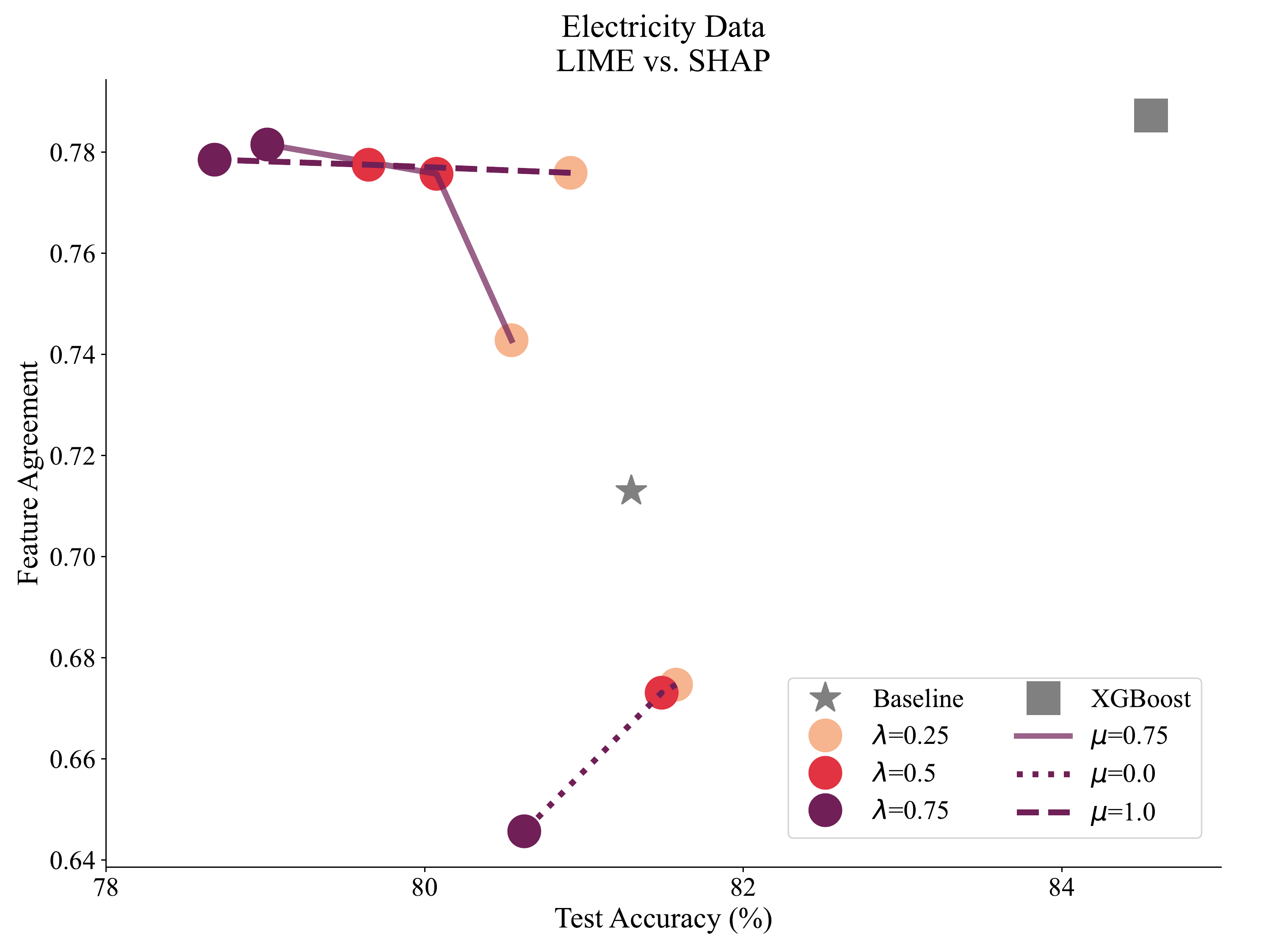}
    \\
    \includegraphics[width=0.275\textwidth]{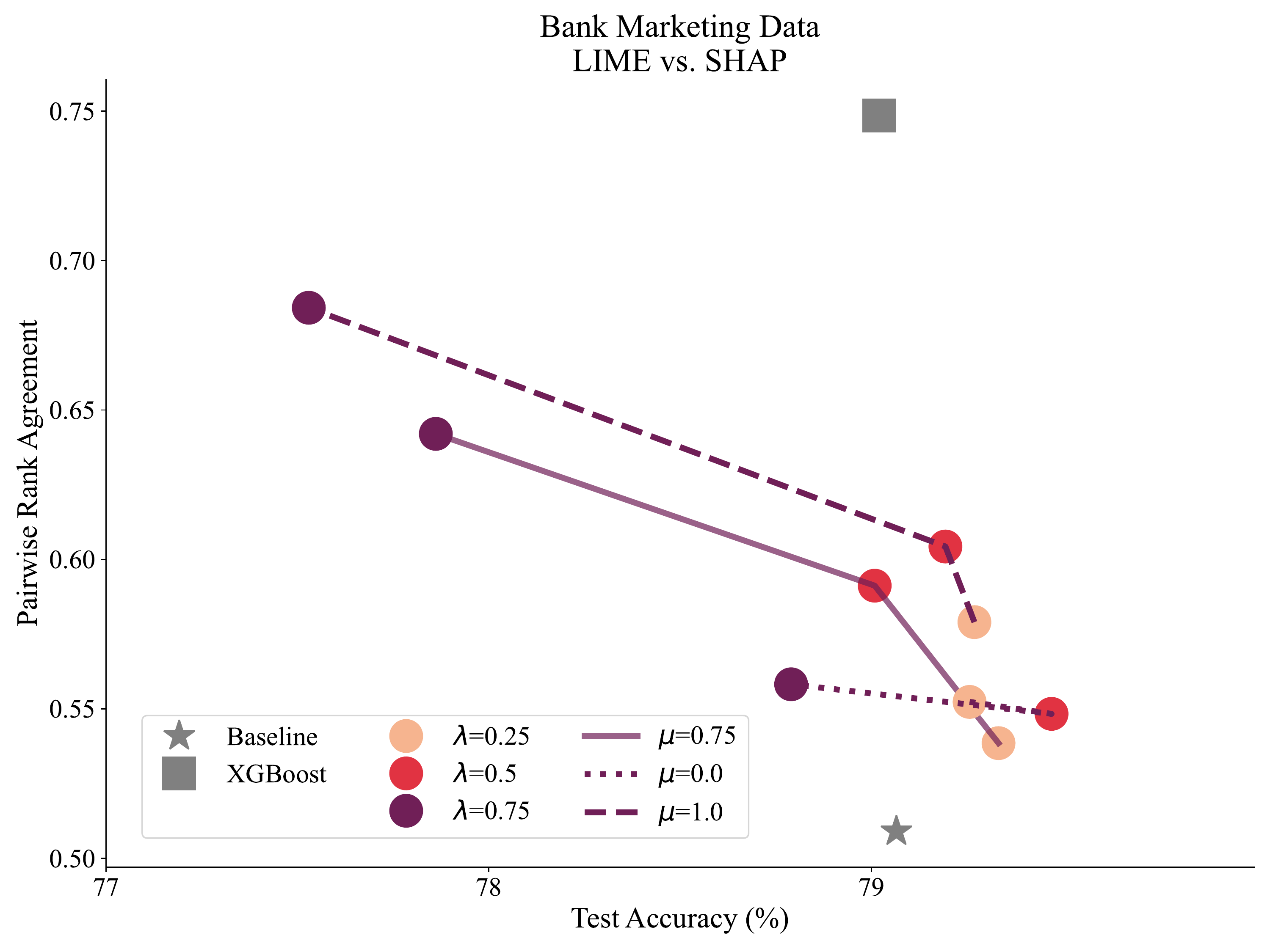}
    \includegraphics[width=0.275\textwidth]{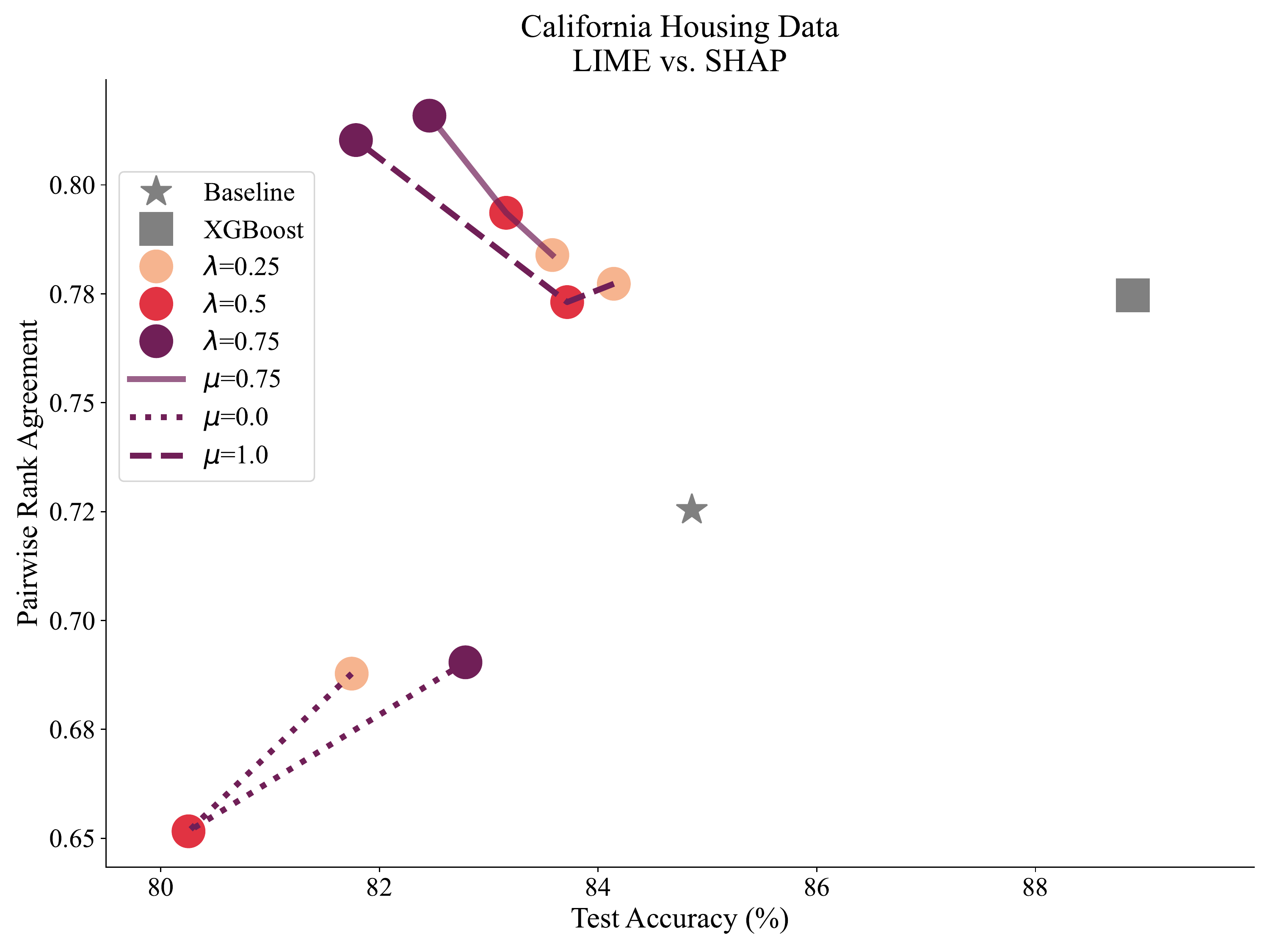}
    \includegraphics[width=0.275\textwidth]{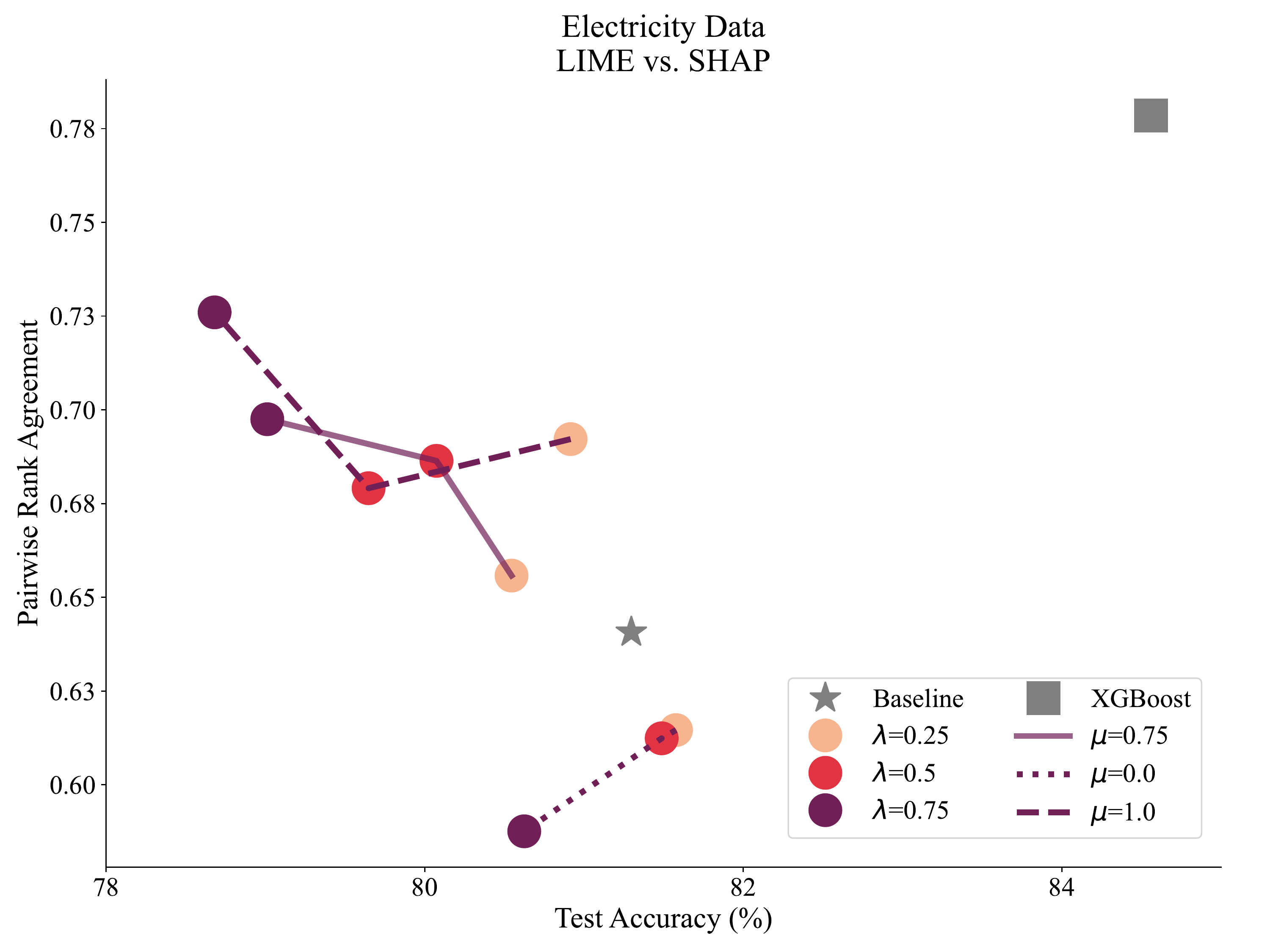}
    \\
    \includegraphics[width=0.275\textwidth]{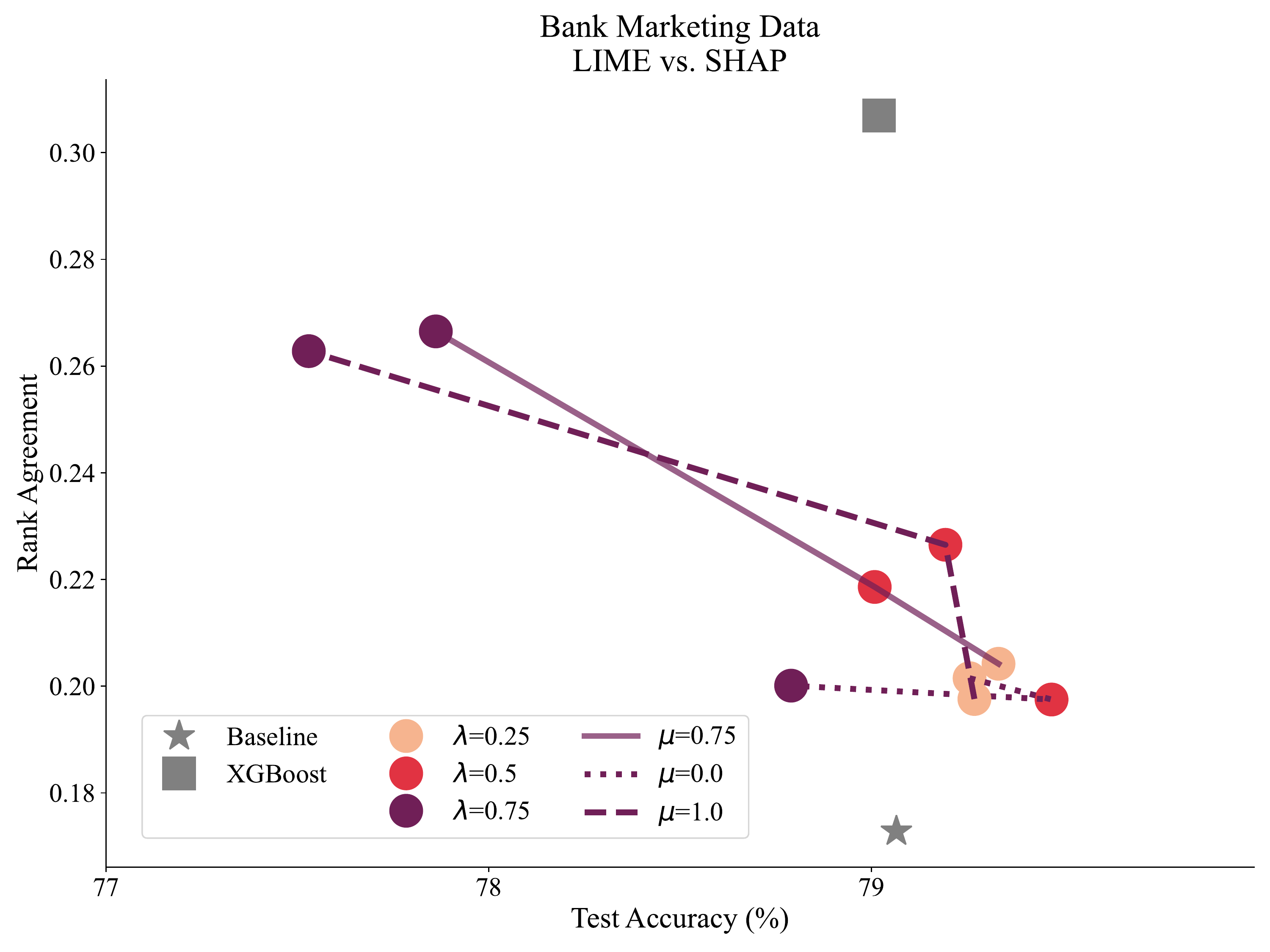}
    \includegraphics[width=0.275\textwidth]{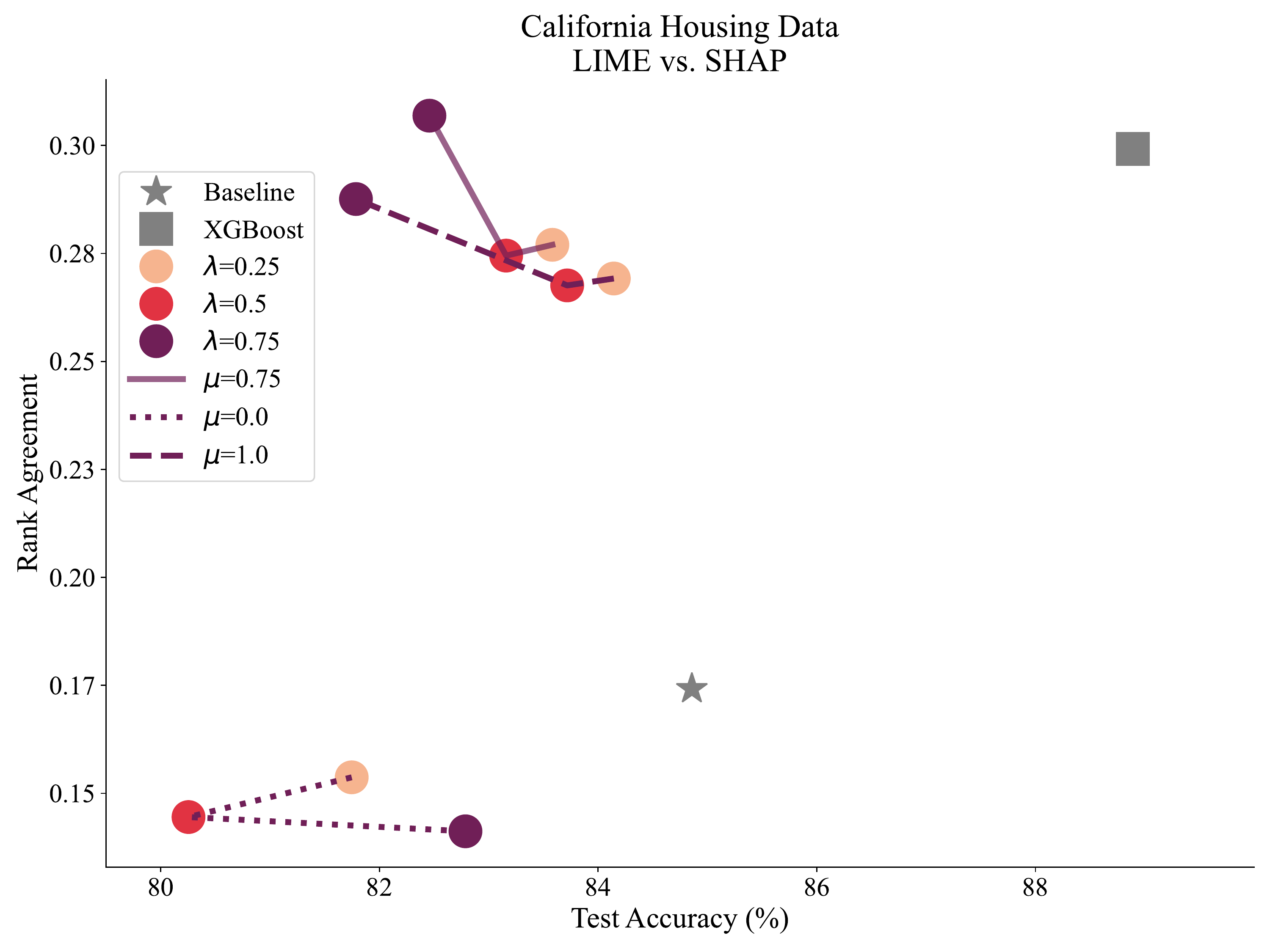}
    \includegraphics[width=0.275\textwidth]{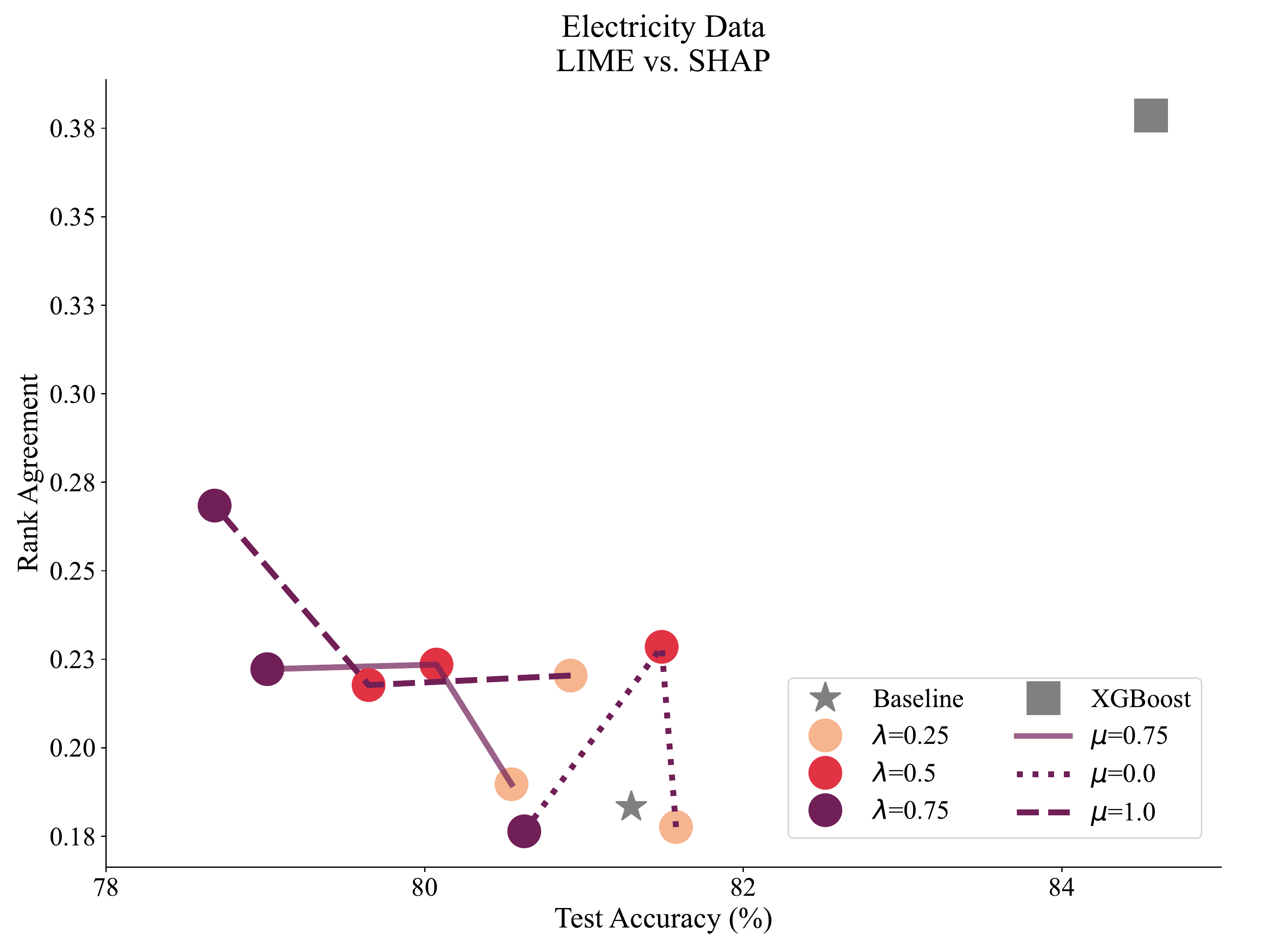}
    \\
    \includegraphics[width=0.275\textwidth]{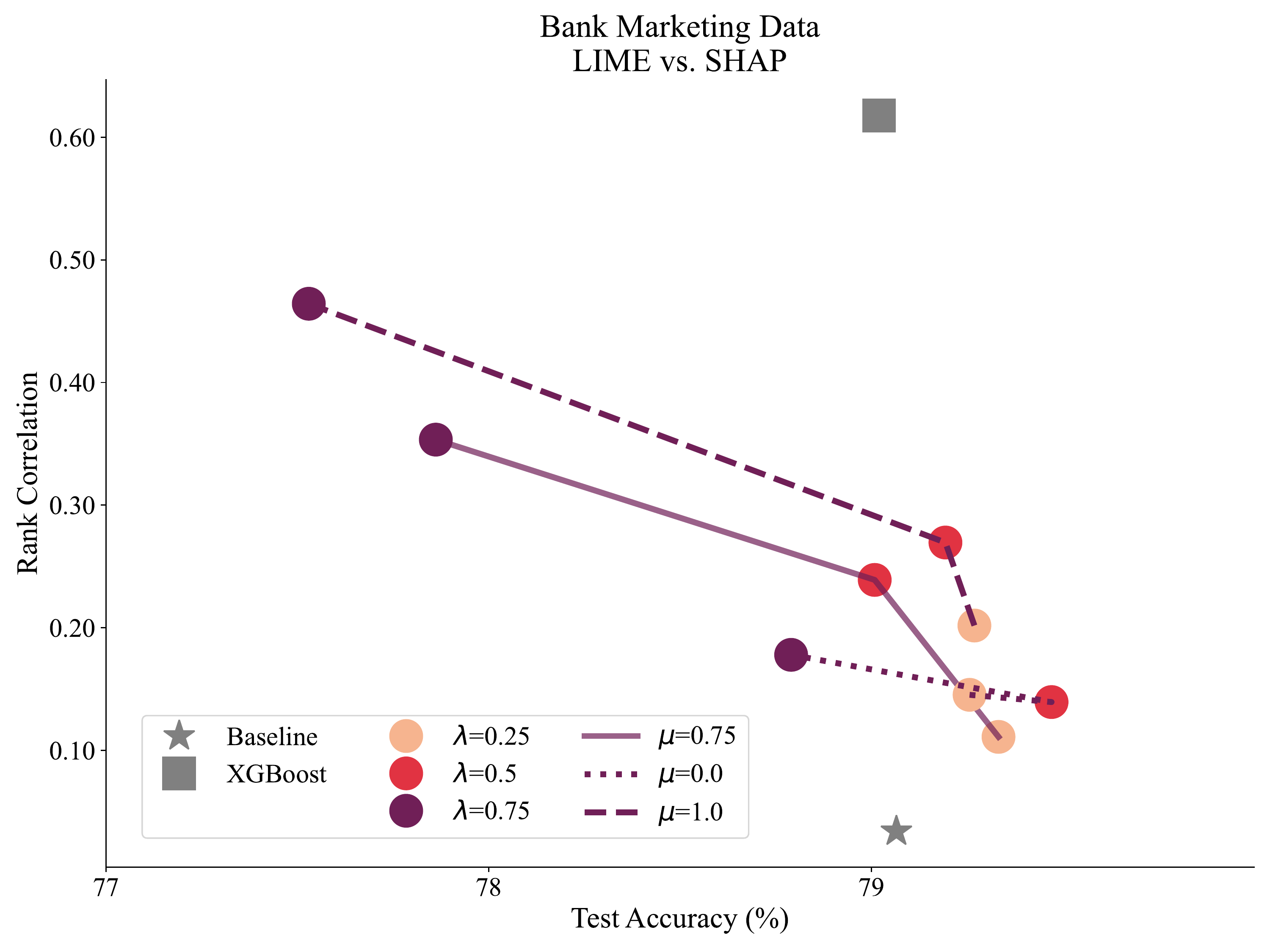}
    \includegraphics[width=0.275\textwidth]{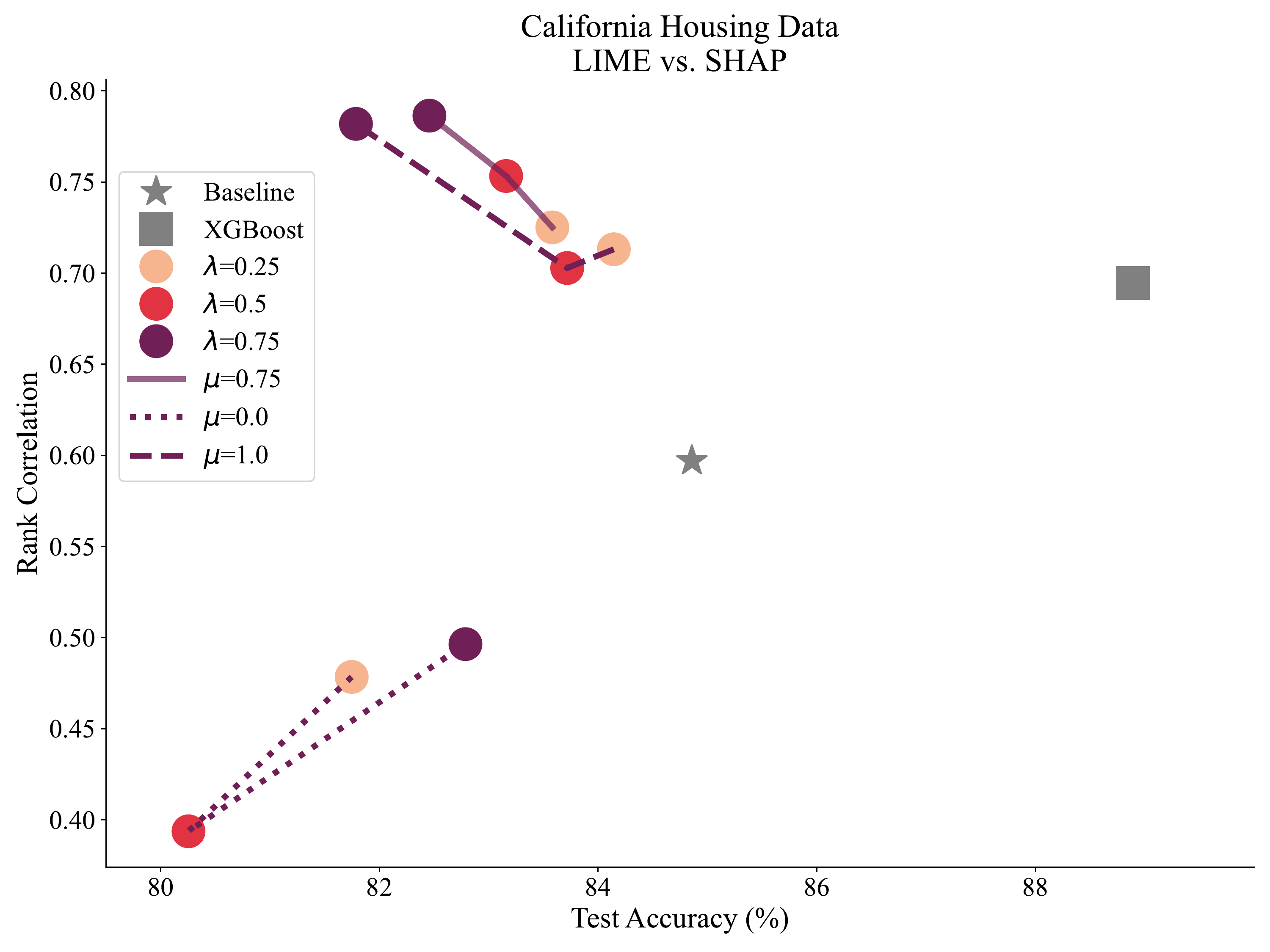}
    \includegraphics[width=0.275\textwidth]{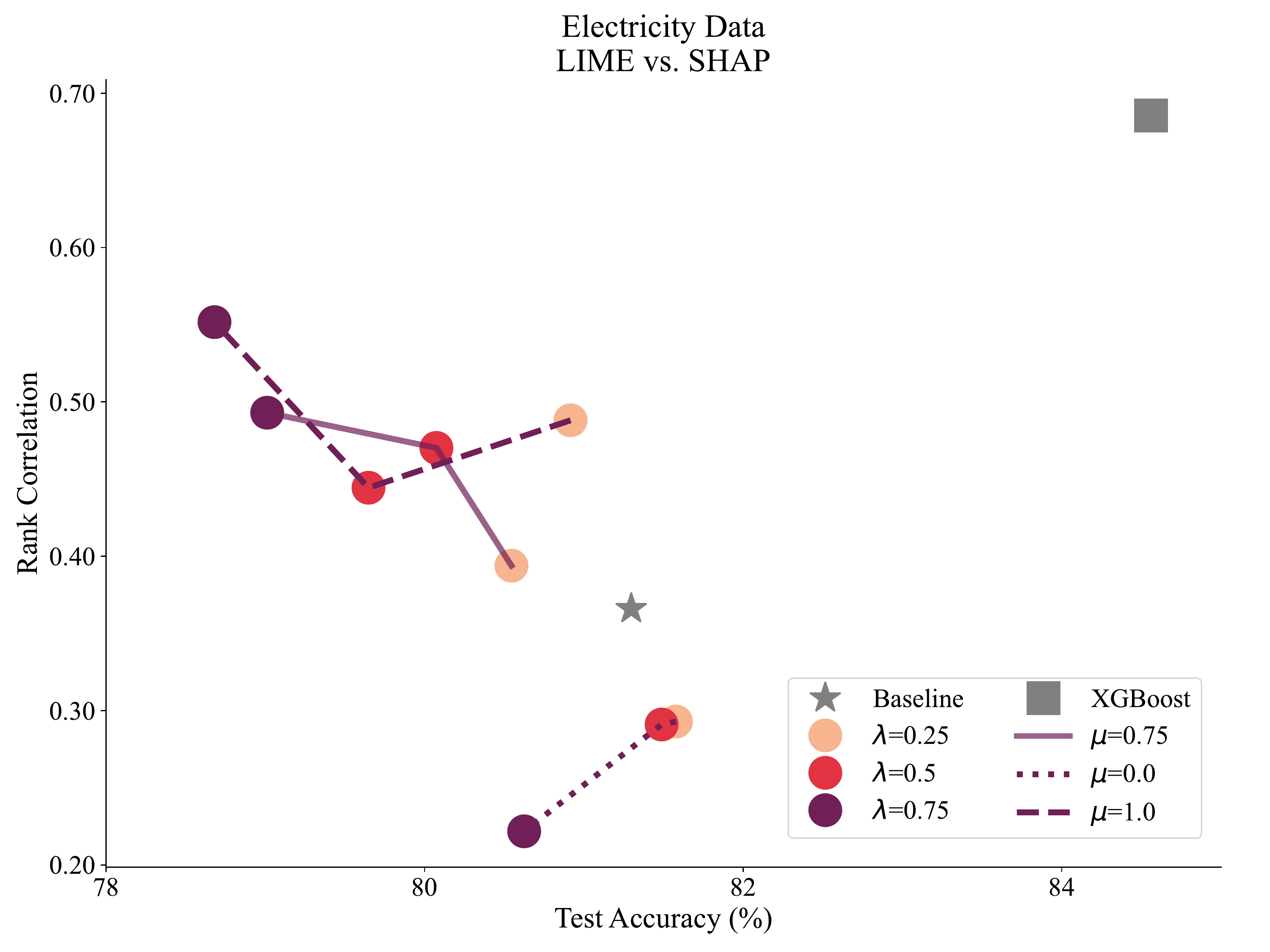}
    \\
    \includegraphics[width=0.275\textwidth]{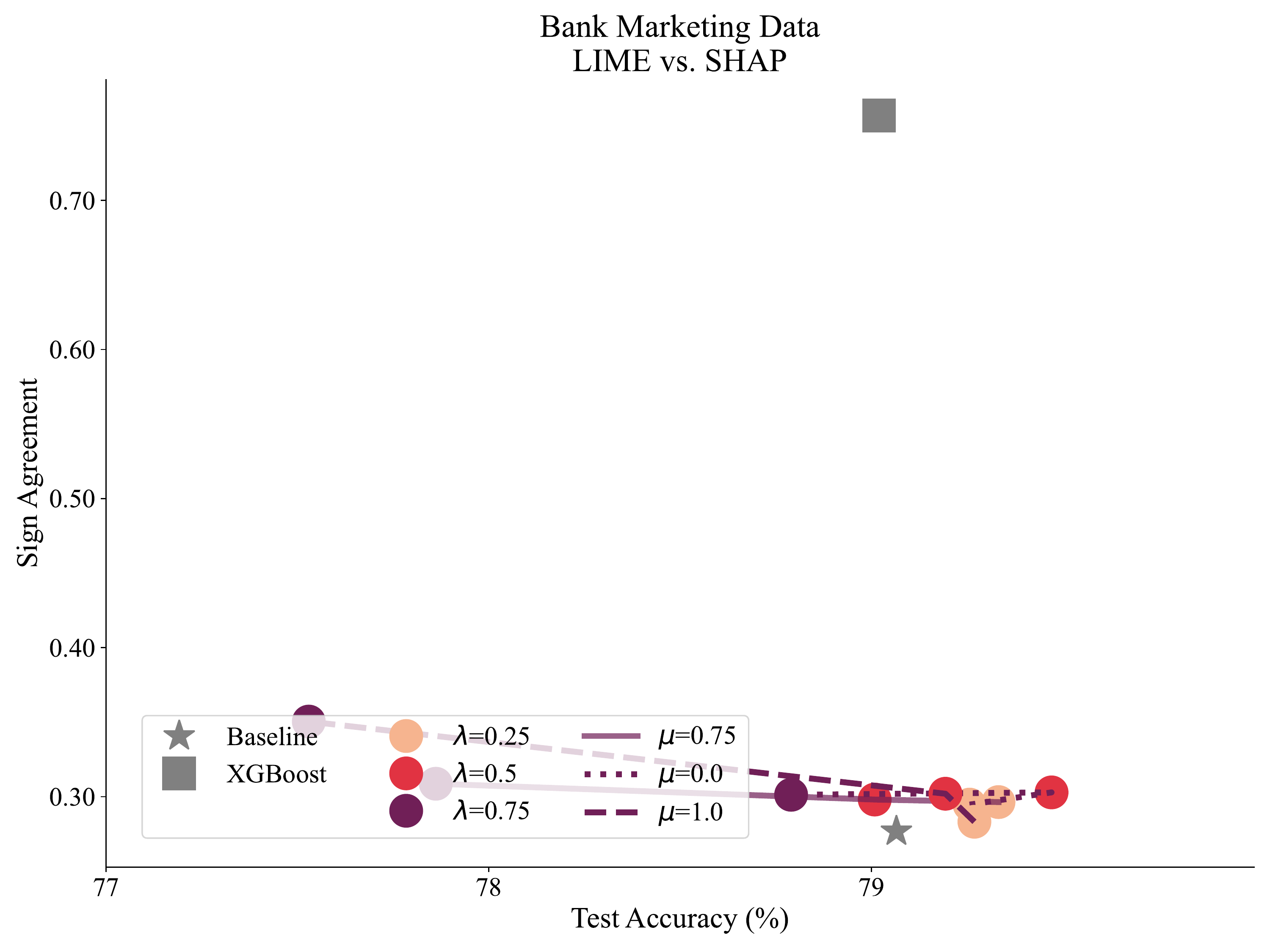}
    \includegraphics[width=0.275\textwidth]{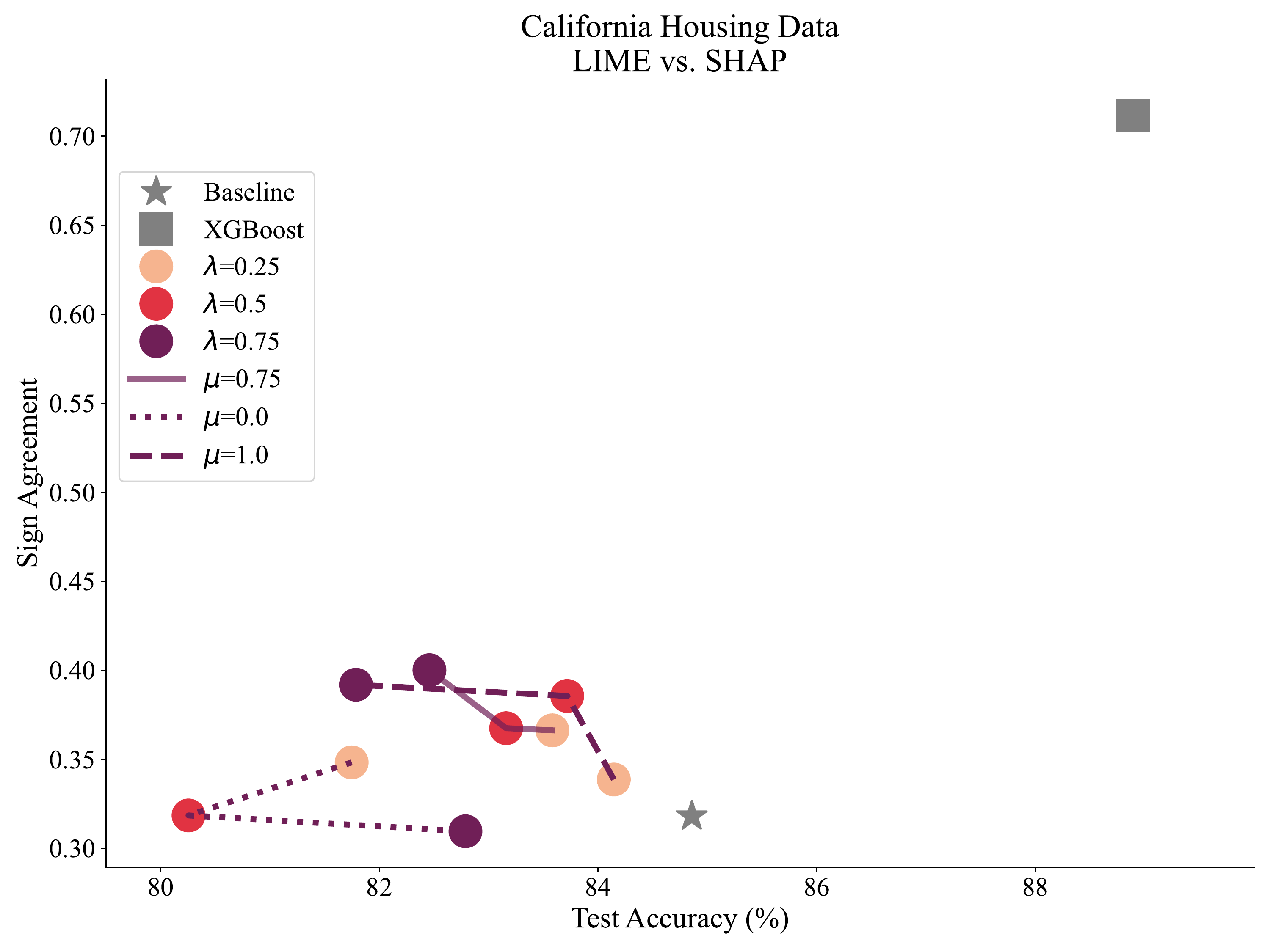}
    \includegraphics[width=0.275\textwidth]{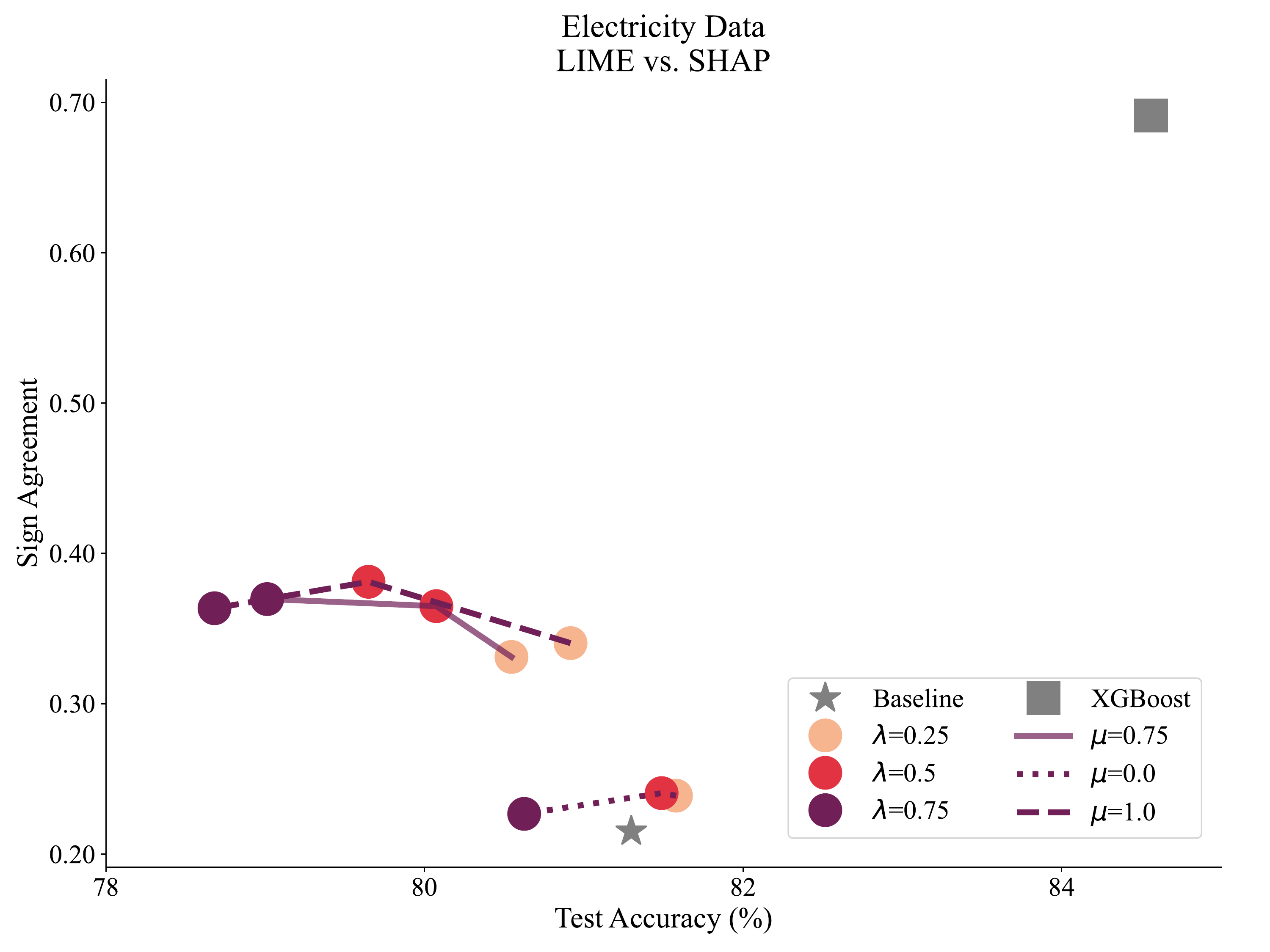}
    \\
    \includegraphics[width=0.275\textwidth]{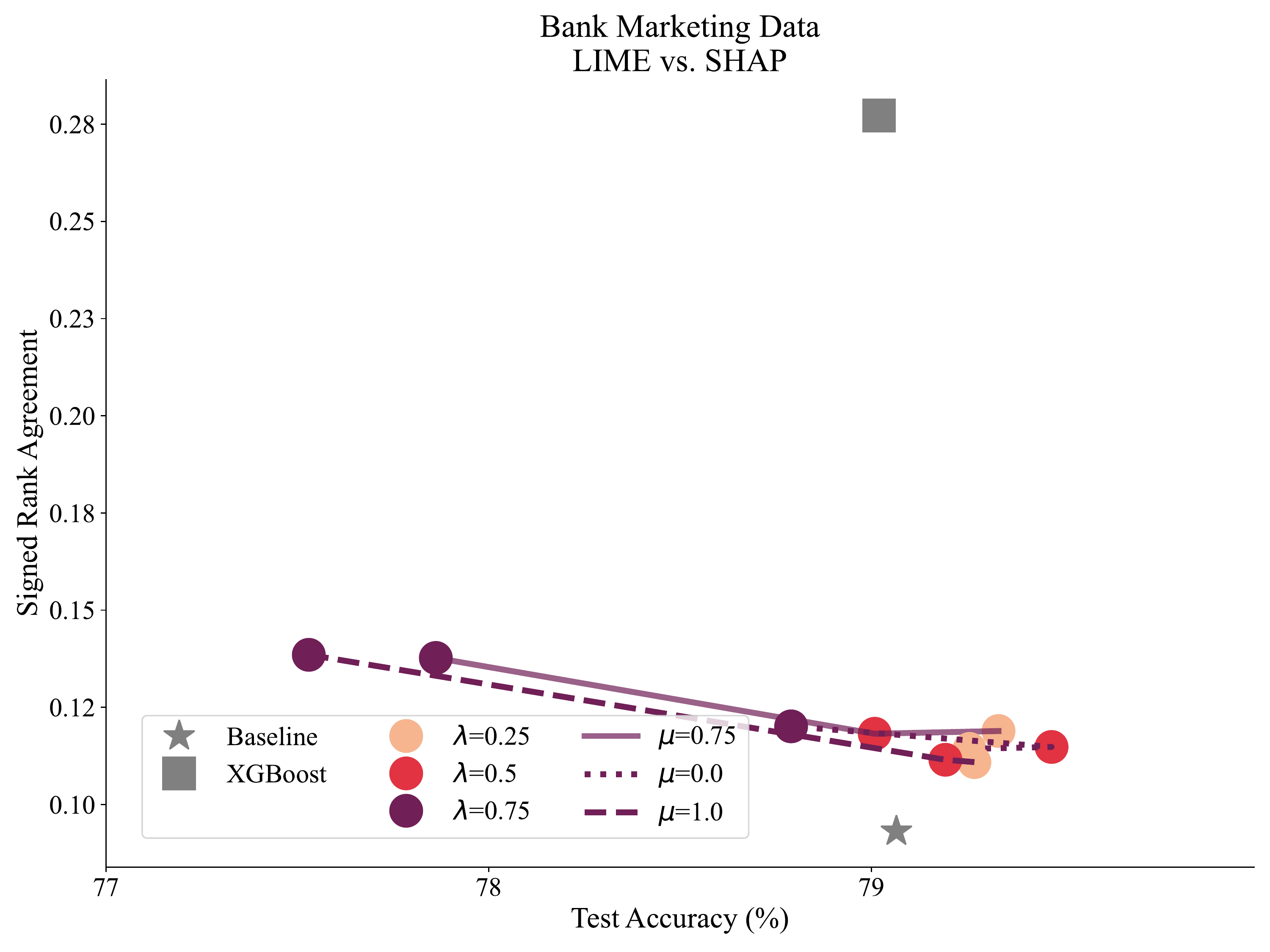}
    \includegraphics[width=0.275\textwidth]{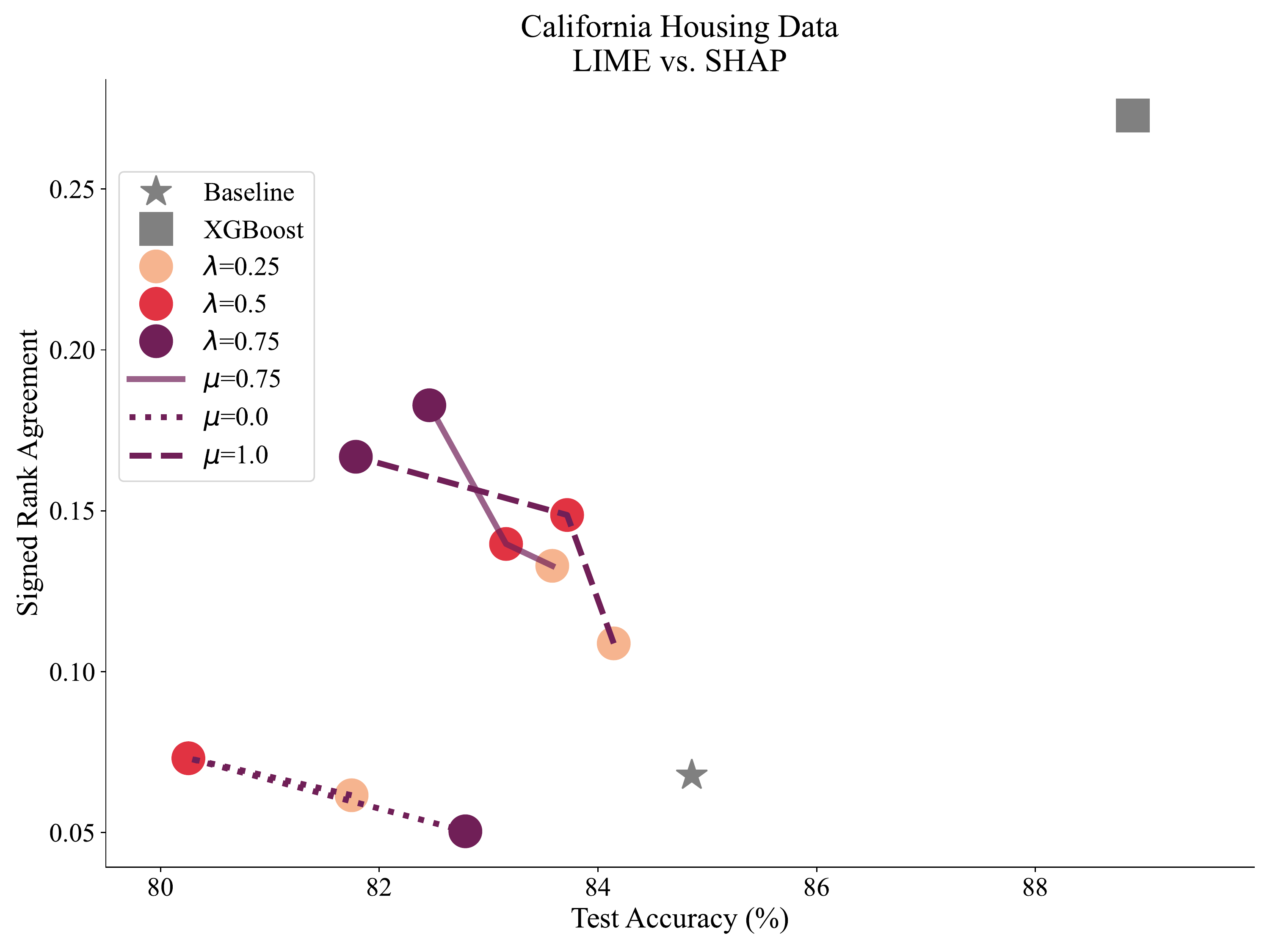}
    \includegraphics[width=0.275\textwidth]{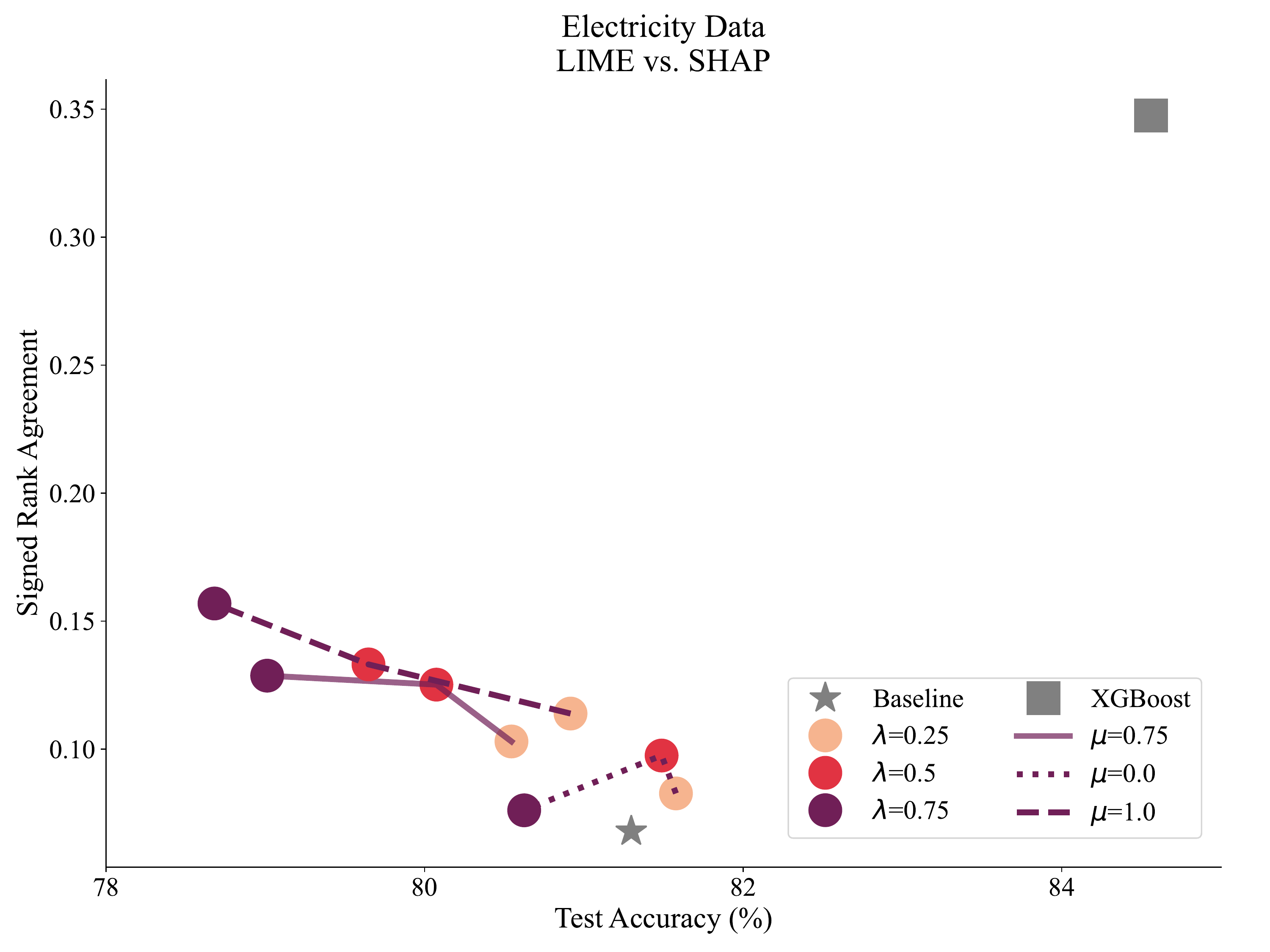}
    \caption{Additional trade-off curve plots for all datasets and metrics.}
    \label{fig:more-busy-paretos}
\end{figure*}

\end{document}